\documentclass[dvipsnames,format=sigconf,authorversion=true]{acmart}

\usepackage{xcolor}
\usepackage{adjustbox}
\usepackage{colortbl}
\usepackage{tabularx}
\usepackage{booktabs} 
\usepackage{graphicx}
\usepackage{subfig}
\usepackage[linesnumbered, vlined, ruled]{algorithm2e}
\SetArgSty{textup}
\usepackage{enumerate}
\usepackage{prettyref}
\usepackage{color, colortbl}
\usepackage{braket}
\definecolor{MyHiLiRow}{gray}{0.9}
\usepackage{tcolorbox}
\usepackage{caption}
\usepackage{seqsplit}

\usepackage{xr}
\newcommand{\argmin}{\mathop{\rm argmin}\limits}
\def\vec#1{\mathbf{#1}}
\def\set#1{\mathcal{#1}}

\newcommand{\pref}{\prettyref}

\newrefformat{fig}{Figure~\ref{#1}}
\newrefformat{supfig}{Figure~\ref{#1}}
\newrefformat{tab}{Table~\ref{#1}}
\newrefformat{suptab}{Figure~\ref{#1}}
\newrefformat{sec}{Section~\ref{#1}}
\newrefformat{app}{Appendix~\ref{#1}}
\newrefformat{alg}{Algorithm~\ref{#1}}
\newrefformat{property}{Property~\ref{#1}}
\newrefformat{theorem}{Theorem~\ref{#1}}
\newrefformat{corollary}{Corollary~\ref{#1}}
\newrefformat{proposition}{Proposition~\ref{#1}}
\newrefformat{def}{Definition~\ref{#1}}
\newrefformat{eq}{equation~(\ref{#1})}

\AtBeginDocument{%
  \providecommand\BibTeX{{%
    \normalfont B\kern-0.5em{\scshape i\kern-0.25em b}\kern-0.8em\TeX}}}

\setcopyright{acmcopyright}

\copyrightyear{2024}
\acmYear{2024}
\setcopyright{acmlicensed}\acmConference[GECCO '24]{Genetic and Evolutionary Computation Conference}{July 14--18, 2024}{Melbourne, VIC, Australia}
\acmBooktitle{Genetic and Evolutionary Computation Conference (GECCO '24), July 14--18, 2024, Melbourne, VIC, Australia}
\acmDOI{10.1145/3638529.3654019}
\acmISBN{979-8-4007-0494-9/24/07}

\begin{document}

\title[Benchmarking PCMs in DE for Mixed-Integer Black-Box Optimization]{Benchmarking Parameter Control Methods in Differential Evolution for Mixed-Integer Black-Box Optimization}





\author{Ryoji Tanabe}
\affiliation{%
  \institution{Yokohama National University}
  \city{Yokohama}
  \state{Kanagawa}
  \country{Japan}}
  \email{rt.ryoji.tanabe@gmail.com}







\begin{abstract}


Differential evolution (DE) generally requires parameter control methods (PCMs) for the scale factor and crossover rate.
Although a better understanding of PCMs provides a useful clue to designing an efficient DE, their effectiveness is poorly understood in mixed-integer black-box optimization.
In this context, this paper benchmarks PCMs in DE on the mixed-integer black-box optimization benchmarking function (\texttt{bbob-mixint}) suite in a component-wise manner.
First, we demonstrate that the best PCM significantly depends on the combination of the mutation strategy and repair method.
Although the PCM of SHADE is state-of-the-art for numerical black-box optimization, our results show its poor performance for mixed-integer black-box optimization.
In contrast, our results show that some simple PCMs (e.g., the PCM of CoDE) perform the best in most cases.
Then, we demonstrate that a DE with a suitable PCM performs significantly better than CMA-ES with integer handling for larger budgets of function evaluations.
Finally, we show how the adaptation in the PCM of SHADE fails.









\end{abstract}





\maketitle

\section{Introduction}
\label{sec:introduction}

\noindent \textit{General context.}
As in the single-objective mixed-integer black-box optimization benchmarking function  (\texttt{bbob-mixint}) suite~\cite{TusarBH19}, this paper considers mixed-integer black-box optimization of an objective function $f: \mathbb{Z}^{n^{\mathrm{int}}} \times \mathbb{R}^{n-n^{\mathrm{int}}} \rightarrow \mathbb{R}$, where $n$ is the total number of variables, $n^{\mathrm{int}}$ is the number of integer variables, and $n-n^{\mathrm{int}}$ is the number of numerical variables.
This problem involves finding a solution $\vec{x} \in \mathbb{Z}^{n^{\mathrm{int}}} \times \mathbb{R}^{n-n^{\mathrm{int}}}$ with an objective value $f(\vec{x})$ as small as possible without any explicit knowledge of $f$.
%

Differential evolution (DE)~\cite{StornP97,PriceSL05} is an efficient evolutionary algorithm for numerical black-box optimization.
Here, numerical black-box optimization can be considered as a special case of mixed-integer black-box optimization when $n^{\mathrm{int}}=0$.
Previous studies~\cite{TanabeF14,Tanabe20} empirically demonstrated that DE performs as well as or better than covariance matrix adaptation evolution strategy (CMA-ES)~\cite{HansenMK03,Hansen16a} under some conditions.

Evolutionary algorithms generally require parameter control methods (PCMs)~\cite{EibenHM99,LoboLM07,KarafotiasHE15} that automatically adjust one or more parameters during the search.
This is true for DE.
Previous studies (e.g., \cite{GamperleMK02,BrestGBMZ06,ZielinskiWLK06}) showed that the performance of DE is sensitive to the setting of two parameters: scale factor $s$ and crossover rate $c$.
To address this issue, a number of DE with PCMs have been proposed~\cite{DasS11}. 
According to \cite{EibenHM99}, PCMs can be classified into three groups: deterministic PCMs, adaptive PCMs, and self-adaptive PCMs.
However, most PCMs in DE are deterministic or adaptive~\cite{TanabeF20}.



A DE is a complex of many components.
For example, as described in \cite{TanabeF20}, ``L-SHADE''~\cite{TanabeF14} mainly consists of the following four components: (i) the current-to-$p$best/1 mutation strategy~\cite{ZhangS09}, (ii) binomial crossover, (iii) the PCM of SHADE~\cite{TanabeF13} for adaptively adjusting the scale factor $s$ and crossover rate $c$, and (iv) linear population size reduction strategy.
This complex property makes an analysis of DE algorithms difficult.
To address this issue, some previous studies (e.g., \cite{ZielinskiWL08,DrozdikAAT15,Tanabe20,TanabeF20,VermettenCKB23}) employed component-wise analysis.
For example, the previous study~\cite{TanabeF20} analyzed only (iii) the PCM of 24 DE algorithms by fixing the other components.


Some previous studies proposed extensions of DE for mixed-integer black-box optimization.
As demonstrated in \cite{LampinenZ99Part1,Liao10}, any DE can handle integer variables by simply using the rounding operator $\mathbb{R} \rightarrow \mathbb{Z}$.
%
%
Some previous studies (e.g., \cite{LiuWHJ22,MolinaPerezMPVC24}) proposed efficient methods for handling integer variables in DE. 
A previous study~\cite{LinDS17} proposed a hybrid method of L-SHADE and ACO$_{\mathrm{MV}}$~\cite{LiaoSOSD14}, called L-SHADE$_{\mathrm{ACO}}$.

%

\vspace{0.2em}
\noindent \textit{Motivation.}
Although some DE algorithms for mixed-integer black-box optimization have been proposed, their analysis has received little attention in the DE community.
In particular, the performance of PCMs in DE is poorly understood in the context of mixed-integer black-box optimization.
On the one hand, the importance of PCMs for numerical black-box optimization has been widely accepted in the DE community.
On the other hand, most previous studies on DE for mixed-integer black-box optimization (e.g., \cite{LampinenZ99Part1,Liao10,LinLCZ18,LiuWHJ22}) did not use any PCM and fixed the two parameters to pre-defined values, e.g., $s=0.5$ and $c=0.9$.

Of course, some DE algorithms for mixed-integer black-box optimization use PCMs.
For example, DE-CaR+S~\cite{MolinaPerezMPVC24} uses a deterministic PCM that randomly generates the scale factor $s$ and crossover rate $s$.
L-SHADE$_{\mathrm{ACO}}$~\cite{LinDS17} uses the PCM of SHADE for adaptation of $s$ and $c$.
However, the effectiveness of the PCMs of DE-CaR+S and SHADE is unclear.
For example, the previous study~\cite{LinDS17} investigated the performance of L-SHADE$_{\mathrm{ACO}}$ but did not investigate the performance of (iii) the PCM of SHADE.
Here, the conclusion ``L-SHADE$_{\mathrm{ACO}}$ performs well'' does not mean ``the PCM of SHADE performs well'' due to the existence of other components.



Some previous studies (e.g., \cite{Hansen11,HamanoSNS22}) proposed extensions of CMA-ES for mixed-integer black-box optimization so that they can handle integer variables.
Three previous studies~\cite{TusarBH19,HamanoSNS22a,MartySAHH23} investigated the performance of CMA-ES with integer handling on the \texttt{bbob-mixint} suite~\cite{TusarBH19}.
Their results showed that the CMA-ES variants perform significantly better than the SciPy implementation of DE.
However, the SciPy implementation of DE is the most classical version of DE~\cite{StornP97} and does not use any PCM.
Thus, it is unclear whether CMA-ES can outperform a DE with an advanced PCM or not.


\vspace{0.2em}
\noindent \textit{Contributions.}
%
Motivated by the above discussion, this paper investigates the performance of nine PCMs in DE for mixed-integer black-box optimization.
Note that we are interested only in a PCM in DE rather than an adaptive DE algorithm. 
%
This paper addresses the following three research questions:

\begin{enumerate}[RQ1:]
\item Are PCMs effective in DE for mixed-integer black-box optimization? If so, which PCMs are useful in which situations?
\item Can a DE algorithm with a suitable PCM outperform CMA-ES with integer handling?
\item How does a state-of-the-art PCM behave? 
  
\end{enumerate}

\vspace{0.2em}
\noindent \textit{Outline.}
Section \ref{sec:preliminaries} gives some preliminaries.
Section \ref{sec:setting} describes our experimental setup.
Section \ref{sec:results} shows the analysis results to answer the three research questions.
Section \ref{sec:conclusion} concludes this paper.

\vspace{0.2em}
\noindent \textit{Supplementary file.}
%
Figure S.$*$, Table S.$*$, and Algorithm S.$*$ indicate a figure, table, and algorithm in the supplement, respectively.

\vspace{0.2em}
\noindent \textit{Code availability.}
%
%
The Python implementation of DE is available at \url{https://github.com/ryojitanabe/de_bbobmixint}.

\section{Preliminaries}
\label{sec:preliminaries}



First, \pref{sec:de} describes the basic DE.
Then, \pref{sec:repair} describes two repair methods: the Baldwinian and Lamarckian repair methods.
Finally, \pref{sec:pcm} describes nine PCMs in DE investigated.

\subsection{Differential evolution}
\label{sec:de}


\pref{alg:de} shows the procedure of DE.
Let $\set{P}=\{\vec{x}_i\}^{\mu}_{i=1}$ be the population of size $\mu$ at iteration $t$.
Each individual $\vec{x} = (x_1, \ldots, x_{n})^{\top}$ in $\set{P}$ consists of the $n$-dimensional numerical vector in $\mathbb{R}^n$.
Since $\vec{x}$ violates the integer constraint, $\vec{x}$ must be repaired so that $\vec{x}$ is a feasible solution for mixed-integer black-box optimization.

At the beginning of the search $t=1$, the population $\set{P}$ of size $\mu$ is initialized randomly (line 1).
The optional external archive $\set{A}$ is also initialized, where $\set{A}$ maintains inferior individuals.
$\set{A}$ is used only when using the current-to-$p$best/1~\cite{ZhangS09} and rand-to-$p$best/1~\cite{ZhangS09books} mutation strategies described later.


After the initialization of $\set{P}$, the following steps (lines 2--14) are repeatedly performed until the termination conditions are satisfied.
For each $i \in \{1, \ldots, \mu\}$, a parameter pair $\langle s_i, c_i \rangle$ is generated by a PCM (line 4), where $\langle  \rangle$ means a tuple.
The scale factor $s > 0$ determines the magnitude of differential mutation.
The crossover rate $c \in [0,1]$ determines the number of elements inherited from each individual $\vec{x}$ to a child $\vec{u}$.
When $\langle s_i, c_i \rangle$ is fixed for each $i \in \{1, \ldots, \mu\}$ at any $t$, Algorithm \ref{alg:de} becomes the DE with no PCM.
Here, \pref{alg:de_no} shows the DE with no PCM.

\def\HiLi{\leavevmode\rlap{\hbox to \hsize{\color{black!7}\leaders\hrule height .8\baselineskip depth .5ex\hfill}}}

\begin{algorithm}[t]
\small
\SetSideCommentRight
$t \leftarrow 1$, initialize $\set{P} =\left\{ \vec{x}_1, \ldots, \vec{x}_{\mu}\right\}$ randomly, $\set{A} \leftarrow \emptyset$ \;
\HiLi Initialize internal parameters for $s$ and $c$\;
\While{The termination criteria are not met}{
\HiLi Generate a  pair of $s$ and $c$ for each individual in $\set{P}$\;
  \For{$i \in \{1, ..., \mu\}$}{
    $\vec{v}_{i} \leftarrow$ Apply mutation with $s_i$ to individuals in $\set{P}$\;
    $\vec{u}_{i} \leftarrow$ Apply crossover with $c_i$ to $\vec{x}_{i}$ and $\vec{v}_{i}$\;
  }
  \For{$i \in \{1, \ldots, \mu\}$}{
    \If{$f(\vec{u}_{i}) \leq f(\vec{x}_{i})$} {
      $\set{A} \leftarrow \set{A} \cup \{\vec{x}_{i}\}$\;
      $\vec{x}_{i} \leftarrow \vec{u}_{i}$\;
    }
  }
  \lIf{$|\set{A}| > a$} {
    Delete randomly selected $|\set{A}| - a$ individuals in $\set{A}$
  }
  \HiLi Update internal parameters for the adaptation of $s$ and $c$\;
  $t \leftarrow t+1$\;
}
\caption{The basic DE algorithm with a PCM}
\label{alg:de}
\end{algorithm}

\begin{table}
\renewcommand{\arraystretch}{0.7}
\begin{center}
  \caption{Eight representative mutation strategies for DE.
}
  \label{tab:de_mutation}
{\small
%
\begin{tabular}{ll}
\midrule
Strategies & Definitions\\
\toprule
rand/1 & ${\begin{aligned}[t] 
\vec{v}_i = \vec{x}_{r_1} + s_i \; (\vec{x}_{r_2} - \vec{x}_{r_3})
\end{aligned}}$ \\\midrule
rand/2 & ${\begin{aligned}[t] 
\vec{v}_i = \vec{x}_{r_1} + s_i \; (\vec{x}_{r_2} - \vec{x}_{r_3}) + s_i \; (\vec{x}_{r_4} - \vec{x}_{r_5})
\end{aligned}}$ \\\midrule
best/1 & ${\begin{aligned}[t] 
\vec{v}_i = \vec{x}_{\mathrm{best}} + s_i \, (\vec{x}_{r_1} - \vec{x}_{r_2})
\end{aligned}}$ \\\midrule
best/2 & ${\begin{aligned}[t]
    \vec{v}_i = \vec{x}_{\mathrm{best}} + s_i \, (\vec{x}_{r_1} - \vec{x}_{r_2}) + s_i \, (\vec{x}_{r_3} - \vec{x}_{r_4})   
\end{aligned}}$ \\\midrule
current-to-rand/1 & ${\begin{aligned}[t] 
\vec{v}_i = \vec{x}_i + s_i \, (\vec{x}_{r_1} - \vec{x}_i) + s_i \, (\vec{x}_{r_2} - \vec{x}_{r_3})
\end{aligned}}$ \\\midrule
current-to-best/1 & ${\begin{aligned}[t] 
\vec{v}_i = \vec{x}_i + s_i \, (\vec{x}_{\mathrm{best}} - \vec{x}_i) + s_i \, (\vec{x}_{r_1} - \vec{x}_{r_2})
\end{aligned}}$ \\\midrule
\shortstack{current-to-$p$best/1} & ${\begin{aligned}[t] 
\vec{v}_i = \vec{x}_i  +  s_i \, (\vec{x}_{p{\rm best}} - \vec{x}_i) +  s_i \, ( \vec{x}_{r_1} - \tilde{\vec{x}}_{r_2})
\end{aligned}}$ \\\midrule
\shortstack{rand-to-$p$best/1} & ${\begin{aligned}[t] 
\vec{v}_i = \vec{x}_{r_1}  +  s_i \, (\vec{x}_{p\mathrm{best}} - \vec{x}_{r_1}) +  s_i \, ( \vec{x}_{r_2} - \tilde{\vec{x}}_{r_3})
\end{aligned}}$ \\\midrule
\end{tabular}
}
\end{center}
\end{table}

For each $i \in \{1, \ldots, \mu\}$, a mutant vector $\vec{v}_{i}$ is generated by applying differential mutation to randomly selected individuals (line 6).
\pref{tab:de_mutation} shows eight representative DE mutation strategies.
If an element of $\vec{v}_{i}$ is outside the bounds, we applied the bound handling method described in \cite{ZhangS09} to it.
In \pref{tab:de_mutation}, the indices $r_1$, $r_2$, $r_3, r_4, $ and $r_5$ are randomly selected from $\{1, ..., \mu\} \backslash \{i\}$ such that they differ from each other.
In \pref{tab:de_mutation}, $\vec{x}_{\mathrm{best}}$ is the best individual with the lowest objective value in  $\set{P}$.
For each $i \in \{1, \ldots, \mu\}$, $\vec{x}_{p\mathrm{best}}$ is randomly selected from the top $\max(\lfloor p \times \mu \rfloor , 2)$ individuals in $\set{P}$, where $p \in [0, 1]$ controls the greediness of the current-to-$p$best/1 and rand-to-$p$best/1 strategies.
A better individual is likely to be selected as $\vec{x}_{p\mathrm{best}}$ when using a smaller $p$ value.
For the current-to-$p$best/1 and rand-to-$p$best/1 strategies, $\tilde{\vec{x}}_{r_2}$ and  $\tilde{\vec{x}}_{r_3}$ are randomly selected from the union of $\set{P}$ and the external archive $\set{A}$.
The use of inferior individuals in $\set{A}$ facilitates the diversity of mutant vectors.
The rand/1 strategy is the most basic strategy.
Since the best/1 and current-to-best/1 strategies are likely to generate mutant vectors near the best individual, they are exploitative.
As in the rand/2 strategy, the use of two difference vectors makes the search explorative.
The current-to-$p$best/1 strategy is used in state-of-the-art DE algorithms (e.g., \cite{BrestMB17,TanabeF14,ZhangS09}).

For each $i \in \{1, \ldots, \mu\}$, after the mutant vector $\vec{v}_i$ has been generated, a child $\vec{u}_i$ is generated by applying crossover to $\vec{x}_i$ and $\vec{v}_i$ (line 7). 
The binomial crossover~\cite{StornP97} is the most representative crossover method in DE, which it can be implemented as follows: for each $j \in \{1, \ldots, n\}$,
$\text{if} \: \mathrm{randu}[0,1] \leq c_i \:\: {\rm or} \:\: j = j_{\mathrm{rand}}$, $u_{i,j} = v_{i,j}$. Otherwise, $u_{i,j} = x_{i,j}$.
%
%
Here, $\mathrm{randu}[a,b]$ returns a random value generated from a uniform distribution in the range $[a,b]$. 
An index $j_{\mathrm{rand}}$ is also randomly selected from $\{1, \ldots, n\}$ and ensures that at least one element is inherited from $\vec{v}_i$ even when $c_i = 0$.


DE performs environmental selection in a pair-wise manner (lines 8--11).
For each $i \in \{1, \ldots, \mu\}$, if $f(\vec{u}_{i}) \leq f(\vec{x}_{i})$, $\vec{x}_{i}$ is replaced with $\vec{u}_{i}$ (line 11).
Thus, the comparison is performed only among the parent $\vec{x}_{i}$ and its child $\vec{u}_{i}$.
Environmental selection in DE allows the replacement of the parent with its child even when they have the same objective value, i.e., $f(\vec{u}_{i}) = f(\vec{x}_{i})$.
As discussed in \cite[Section 4.2.3, pp. 192]{PriceSL05}, this property is helpful for DE to escape a plateau, which generally appears in mixed-integer black-box optimization~\cite{TusarBH19,VolzNKT19}.

If the parent $\vec{x}_{i}$ is replaced with its child $\vec{u}_{i}$, $\vec{x}_{i}$ is added to $\set{A}$ (line 10).
When the archive size $|\set{A}|$ exceeds a pre-defined size $a$, randomly selected individuals in $\set{A}$ are deleted to keep the archive size constant (line 12).
At the end of each iteration, the internal parameters in the PCM are updated (line 13).



\subsection{Lamarckian and Baldwinian repair methods}
\label{sec:repair}


Let $\vec{x} = (x_1, \ldots, x_n)^{\top} \in \mathbb{R}^n$ be an individual in DE.
Since $\vec{x}$ is an infeasible solution for mixed-integer black-box optimization, 
 $\vec{x}$ must be repaired before evaluating $\vec{x}$ by the objective function.
The rounding operator has been generally used to repair $\vec{x}$ in the DE community~\cite{LampinenZ99Part1,Liao10}.
Let $i$ be an index for an integer variable.
In the rounding operator, the $i$-th variable $x_i$ in $\vec{x}$ is rounded to the nearest integer.
For example, if $x_i = 2.024$, $x_i$ is rounded to $2$.

The Lamarckian and Baldwinian repair methods have been considered in the evolutionary computation community~\cite{WhitleyGM94,StreichertUZ04,IshibuchiKN05,Wessing13}, where these terms come from the Lamarckian evolution and Baldwin effect in the field of evolutionary biology, respectively.
Let $\vec{x}^{\mathrm{rep}}$ be a repaired feasible version of an individual $\vec{x}$ in DE by the rounding operator.
In both the Lamarckian and Baldwinian repair methods, $f(\vec{x}^{\mathrm{rep}})$ is used as $f(\vec{x})$.

On the one hand, the Lamarckian repair method replaces $\vec{x}$ with $\vec{x}^{\mathrm{rep}}$.
Thus, in the Lamarckian repair method, the result of the repair is reflected to the original $\vec{x}$.
All individuals in the population are feasible when using the Lamarckian repair method.

On the other hand, the Baldwinian repair method does not make any modifications to $\vec{x}$.
Thus, in the Baldwinian repair method, $\vec{x}$ is infeasible even after the repair.
The repaired feasible solution $\vec{x}^{\mathrm{rep}}$ is used only to compute the objective function $f$.

Except for \cite{LampinenZ99Part1}, most previous studies on DE for mixed-integer black-box optimization did not clearly describe which repair method was used.
As pointed out in \cite{Salcedo-Sanz09}, there is also no clear winner between the Lamarckian and Baldwinian repair methods in evolutionary algorithms.
Thus, it is unclear which repair method is suitable for DE for mixed-integer black-box optimization.





\subsection{Nine PCMs in DE}
\label{sec:pcm}

This section briefly describes the following nine PCMs in DE:
the PCM of CoDE (P-Co)~\cite{WangCZ11},
the PCM of SinDE (P-Sin)~\cite{DraaBB15},
the PCM of DE-CaR+S (P-CaRS)~\cite{MolinaPerezMPVC24},
the PCM of jDE (P-j)~\cite{BrestGBMZ06},
the PCM of JADE (P-JA)~\cite{ZhangS09},
the PCM of SHADE (P-SHA)~\cite{TanabeF13},
the PCM of EPSDE (P-EPS)~\cite{MallipeddiSPT11}, 
the PCM of CoBiDE (P-CoBi)~\cite{WangLHL14}, and
the PCM of cDE (P-c)~\cite{Tvrdik06}.
Here, our descriptions of PCMs are based on \cite{TanabeF20}.
We re-emphasized that we focus on PCMs in DE (e.g., P-SHA) rather than complex DE algorithms (e.g., SHADE and L-SHADE).
While P-Co, P-Sin, and P-CaRS are deterministic PCMs with no feedback, the others are adaptive PCMs.
Except for P-CaRS, we selected these PCMs based on the results in \cite{TanabeF20}.
Since DE-CaR+S is one of the latest DE algorithms for mixed-integer black-box optimization, we investigate the performance of P-CaRS.
The nine PCMs can be incorporated into \pref{alg:de} in a plug-in manner.
Although this section briefly describes the nine PCMs due to the paper length limitation, their details can be found in Algorithms \ref{alg:de_pco}--\ref{alg:de_pc}.

Below, for each $i \in \{1, \ldots, \mu\}$, the pair of $s_i$ and $c_i$ is said to be \emph{successful} if $f(\vec{u}_{i}) \leq f(\vec{x}_{i})$ in \pref{alg:de} (line 9).
Otherwise, the pair of $s_i$ and $c_i$ is said to be \emph{failed}.
Since the use of successful parameters leads to the improvement of individuals, it is expected that successful parameters are more suitable for a given problem than failed parameters.







\vspace{0.2em}
\noindent \textit{P-Co~\cite{WangCZ11}.}
P-Co is the simplest of the nine PCMs.
For each iteration $t$, for each $i \in \{1, \ldots, \mu\}$, a pair of $s_{i}$ and $c_{i}$ is randomly selected from three pre-defined pairs of $s$ and $c$: $\langle 1, 0.1\rangle$, $\langle 1, 0.9\rangle$, and $\langle 0.8, 0.2\rangle$.


\vspace{0.2em}
\noindent \textit{P-Sin~\cite{DraaBB15}.}
All individuals use the same $s$ and $c$ for each iteration $t$.
As its name suggests, P-Sin uses the sinusoidal function to generate the $s$ and $c$ values for each $t$ as follows: $s = \frac{1}{2} \left(\frac{t}{t^{\rm max}} ({\sin}(2 \pi \omega t)) + 1 \right)$ and $c = \frac{1}{2} \left(\frac{t}{t^{\rm max}} ({\sin}(2 \pi \omega t + \pi)) + 1 \right)$.
Here, $\omega$ is the angular frequency, and $t^{\rm max}$ is the maximum number of iterations.
In \cite{DraaBB15}, $\omega = 0.25$ was recommended.


\vspace{0.2em}
\noindent \textit{P-CaRS~\cite{MolinaPerezMPVC24}.}
In the nine PCMs, only P-CaRS was designed for mixed-integer black-box optimization.
For each iteration $t$, for each individual, $s_i$ is a random value in the range $[0.5, 0.55]$.
In contrast, the same $c$ value is assigned to all individuals.
For each iteration, $c$ is randomly selected from $\{0.5, 0.6, 0.7, 0.8, 0.9\}$.

\vspace{0.2em}
\noindent \textit{P-j~\cite{BrestGBMZ06}.}
A pair of $s_{i}$ and $c_{i}$ is assigned to each individual, where $s_{i}=0.5$ and $c_{i}=0.9$ at $t=1$.
For each iteration $t$, each individual generates a child by using $s^{\mathrm{trial}}_{i}$ and $c^{\mathrm{trial}}_{i}$ instead of $s_{i}$ and $c_{i}$.
With pre-defined probabilities $\tau_s$ and $\tau_{c}$, $s^{\mathrm{trial}}_{i}$ and $c^{\mathrm{trial}}_{i}$ are set to random values as follows: $s^{\mathrm{trial}}_{i} = \mathrm{randu}[0.1,1]$ and $c^{\mathrm{trial}}_{i} = \mathrm{randu}[0,1]$.
Otherwise, $s^{\mathrm{trial}}_{i}=s_{i}$ and $c^{\mathrm{trial}}_{i} = c_{i}$.
In \cite{BrestGBMZ06}, $\tau_s=0.1$ and $\tau_{c}=0.1$ were recommended.
If $s^{\mathrm{trial}}_{i}$ and $c^{\mathrm{trial}}_{i}$ are successful, $s_i = s^{\mathrm{trial}}_{i}$ and $c_i = c^{\mathrm{trial}}_{i}$ for the next iteration.


\vspace{0.2em}
\noindent \textit{P-JA~\cite{ZhangS09}.}
%
P-JA adaptively adjusts $s$ and $c$ by using two meta-parameters $m_{s}$ and $m_{c}$, respectively.
Both $m_{s}$ and $m_{c}$ are initialized to $0.5$ for $t=1$.
For each iteration, for each individual, $s_{i}$ and $c_{i}$ are set to values randomly selected from a Cauchy distribution $C(m_s, 0.1)$ and a Normal distribution $N(m_c, 0.1)$, respectively.
At the end of each iteration, $m_{s}$ and $m_{c}$ are updated based on sets $\Theta_s$ and $\Theta_{c}$ of successful $s$ and $c$ values: $m_{s} = (1 - \alpha) m_{s} + \alpha  \, \mathrm{Lmean}(\Theta_{s})$ and $m_{c} = (1 - \alpha) m_{c} + \alpha \, \mathrm{mean}(\Theta_{c})$.
Here, $\alpha \in [0,1]$ is a learning rate, and $\alpha=0.1$ was recommended in \cite{ZhangS09}.
$\mathrm{Lmean}(\Theta)$ and $\mathrm{mean}(\Theta)$ return the Lehmer mean and mean of the input set $\Theta$, respectively.
If $\Theta_s=\emptyset$ and $\Theta_{c}=\emptyset$ at that $t$, P-JA does not update $m_{s}$ and $m_{c}$.



\vspace{0.2em}
\noindent \textit{P-SHA~\cite{TanabeF13}.}
P-SHA is similar to P-JA.
Instead of $m_s$ and $m_c$, P-SHA adaptively adjusts $s$ and $c$ by using two historical memories $\vec{m}_{s} = (m_{s,1}, ..., m_{s,h})^{\top}$ and $\vec{m}_{c} = (m_{c,1}, ..., m_{c,h})^{\top}$, respectively.
Here, $h$ is a memory size, and $h=10$ was recommended in \cite{TanabeF17}.
For $t=1$,  all $h$ elements in $\vec{m}_{s} $ and $ \vec{m}_{c}$ are initialized to $0.5$.
A memory index $k \in \{1, ..., h\}$ is also initialized to 1.

Although some slightly different versions of P-SHA are available, we consider the simplest one described in \cite{TanabeF20}.
For each iteration, for each $i \in \{1, \ldots, \mu\}$, $s_{i}$ and $c_{i}$ are
set to values randomly selected from $C(m_{s,r}, 0.1)$ and $N(m_{c,r}, 0.1)$, respectively.
Here, $r$ is a random number in $\{1, \ldots, h\}$.
At the end of each iteration, the $k$-th elements $m_{s,k}$ and $m_{c,k}$ are updated  based on sets $\Theta_s$ and $\Theta_{c}$ of successful $s$ and $c$ values: $m_{s,k} = \mathrm{Lmean}(\Theta_{s})$ and $m_{c,k} = \mathrm{Lmean}(\Theta_{c})$.
After the update, $k$ is incremented.
If $k > h$, $k$ is re-initialized to $1$.


\vspace{0.2em}
\noindent \textit{P-EPS~\cite{MallipeddiSPT11}.}
P-EPS uses two parameter sets for the adaptation of $s$ and $c$: $\set{Q}_s = \{0.4, 0.5, ..., 0.9\}$ and $\set{Q}_c = \{0.1, 0.2, ..., 0.9\}$.
For $t=1$, for each $i \in \{1, \ldots, \mu\}$, $s_i$ and $c_i$ are initialized with values randomly selected from $\set{Q}_s$ and $\set{Q}_c$, respectively.
At the end of each iteration, if $s_i$ and $c_i$ are failed, they are re-initialized.

\vspace{0.2em}
\noindent \textit{P-CoBi~\cite{WangLHL14}.}
P-CoBi is similar to P-EPS.
The only difference between the two is how to generate $s_{i}$ and $c_{i}$.
In P-CoBi, $s_{i}$ and $c_{i}$ are set to values randomly selected from a bimodal distribution consisting of two Cauchy distributions as follows: $s_{i} \sim C (0.65, 0.1) $ or $C (1, 0.1)$, and $c_{i} \sim  C (0.1, 0.1)$ or $C (0.95, 0.1)$.



\vspace{0.2em}
\noindent \textit{P-c~\cite{Tvrdik06}.}
For each iteration, for each $i \in \{1, \ldots, \mu\}$, P-c randomly selects a pair of $s$ and $c$ from nine combinations of values taken from $\{0.5, 0.8, 1\}$ and $\{0, 0.5, 1\}$, i.e., $\vec{q}_1 = \langle 0.5, 0 \rangle, \vec{q}_2 = \langle 0.5, 0.5 \rangle, ..., \vec{q}_9 = \langle 1, 1 \rangle$.
Here, for each $k \in \{1, ..., 9\}$, the probability $\tau_{k} \in [0,1]$ of selecting $\vec{q}_k$ is given as follows:
$\tau_{k} = (o_{k} + \epsilon) / (\sum^9_{l=1} (o_{l} + \epsilon))$, 
where $\epsilon$ is a parameter to avoid $\tau_{k}=0$.
In addition, $o_{k}$ represents the number of successful trials of $\vec{q}_{k}$ from the last initialization.
When any $\tau_{k}$ is below the threshold $\delta$, $o_1, \ldots, o_9$ are reinitialized to $0$.
The recommended settings of $\epsilon$ and $\delta$ are $2$ and $1/45$, respectively.

\section{Experimental setup}
\label{sec:setting}

This section describes the experimental setup.
We conducted all experiments using the COCO platform~\cite{HansenARMTB21}.
We used a workstation with an Intel(R) 48-Core Xeon Platinum 8260 (24-Core$\times 2$) 2.4GHz and 384GB RAM using Ubuntu 22.04.
The \texttt{bbob-mixint} suite~\cite{TusarBH19} used in this work consists of the 24 mixed-integer functions $f_1, \ldots, f_{24}$, which are mixed-integer versions of the 24 noiseless BBOB functions~\cite{hansen2012fun}.
For each $n$-dimensional problem, $4n/5$ and $1n/5$ variables are integer and continuous, respectively.
The feasible solution space $\mathbb{X}$ consists of $\mathbb{X}=\{0, 1\}^{\frac{n}{5}} \times \{0, 1, 2, 3\}^{\frac{n}{5}} \times \{0, 1, \ldots, 7\}^{\frac{n}{5}} \times \{0, 1, \ldots, 15\}^{\frac{n}{5}} \times [-5, 5]^{\frac{n}{5}}$.
Details of the 24 functions can be found in \url{https://numbbo.github.io/gforge/preliminary-bbob-mixint-documentation/bbob-mixint-doc.pdf}.
We set $n$ to $5, 10, 20, 40,$ $80,$ and $160$. 
According to the COCO platform, we set the number of instances to 15 for each function.
In other words, we perform 15 independent runs for each function.

We implemented DE algorithms with the nine PCMs in Python.
We used the default settings of the hyper-parameters for the nine PCMs.
In addition to them, we evaluate the performance of DE with no PCM as a baseline.
We denote this version of DE as ``NOPCM''.
Here, as in most previous studies \cite{BrestGBMZ06,ZhangS09}, we set $s=0.5$ and $c=0.9$ for NOPCM.
We set $\mu$ to $100$.
We set $p=0.05$ and the archive size $a=\mu$ in the current-to-$p$best/1 and rand-to-$p$best/1 strategies.
We set the maximum number of function evaluations to $10^4 \times n$.

\definecolor{c1}{RGB}{150,150,150}
\definecolor{c2}{RGB}{220,220,220}

\section{Results}
\label{sec:results}

This section describes our analysis results.
Through experiments, Sections \ref{sec:vs_pcms}--\ref{sec:analysis_pcm_shade} aim to address the three research questions (\textbf{RQ1}--\textbf{RQ3}) described in \pref{sec:introduction}, respectively.
For the sake of simplicity, we refer to ``a DE with a PCM'' as ``a PCM''.
For example, we refer to a DE with P-j as P-j.

\subsection{Comparisons of PCMs}
\label{sec:vs_pcms}

\begin{figure*}[t]
\newcommand{\width}{0.316}
\centering
\subfloat[rand/1 ($n=10$)]{
\includegraphics[width=\width\textwidth]{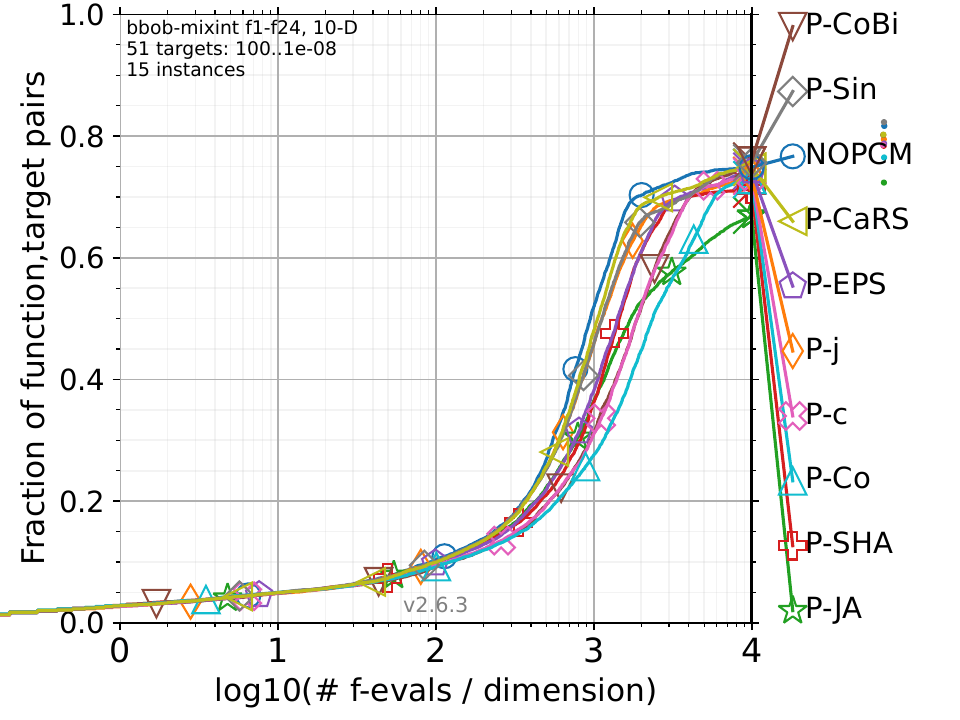}
}
\subfloat[rand/1 ($n=80$)]{
\includegraphics[width=\width\textwidth]{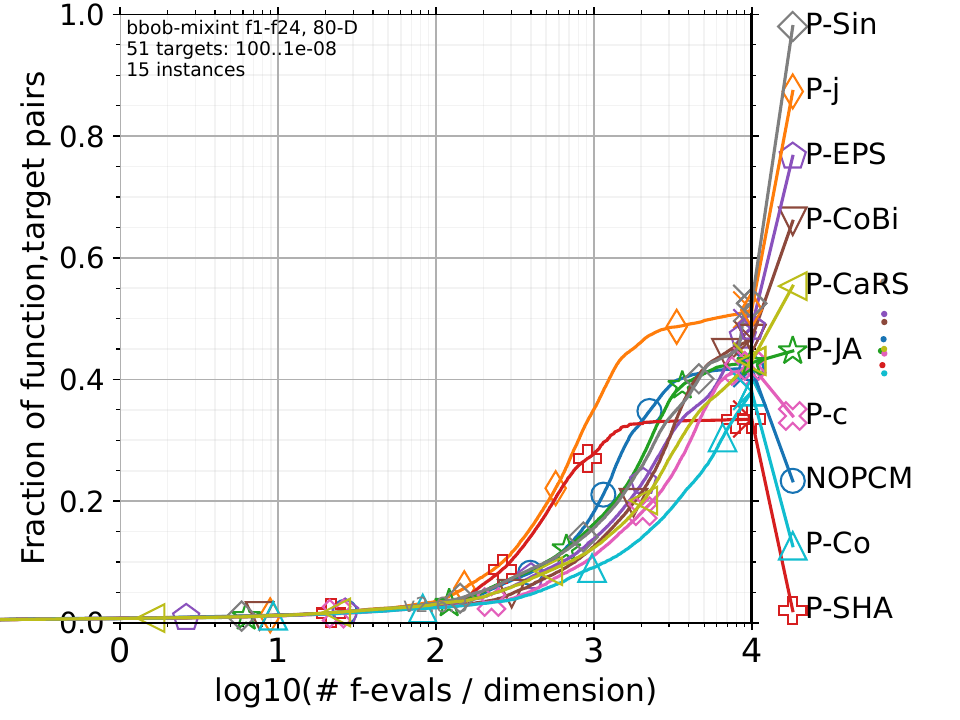}
}
\subfloat[rand/1 ($n=160$)]{
\includegraphics[width=\width\textwidth]{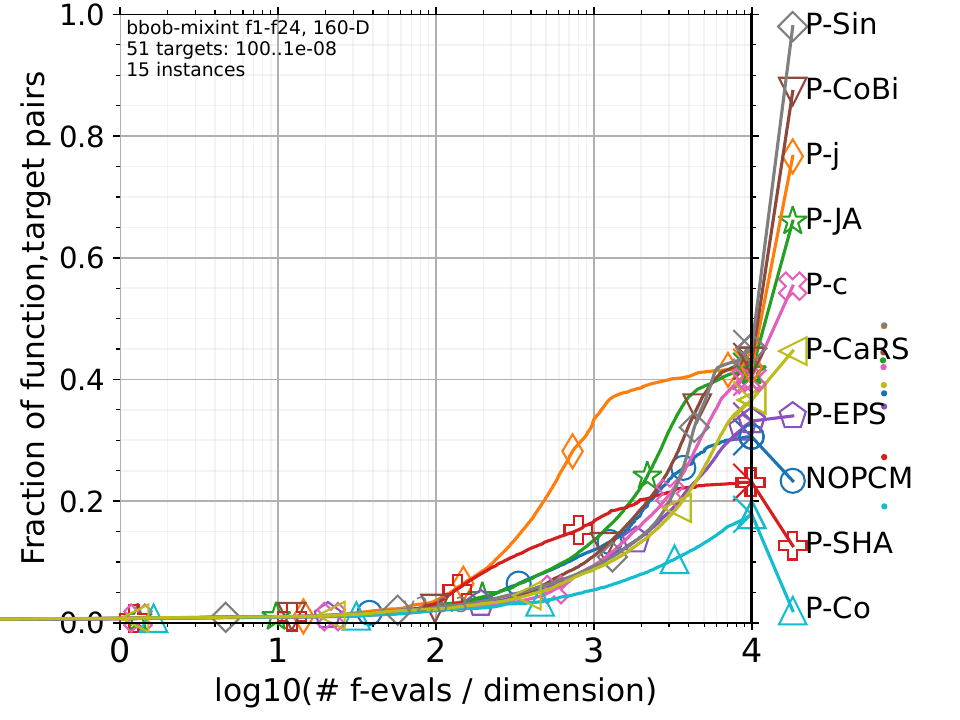}
}
\caption{Comparison of the nine PCMs and NOPCM using the rand/1 strategy and Baldwinian repair method on the 24 \texttt{bbob-mixint} functions with $n \in \{10, 80, 160\}$.}
\label{fig:vs_de_Baldwin_r1}
\subfloat[rand-to-$p$best/1 ($n=10$)]{
\includegraphics[width=\width\textwidth]{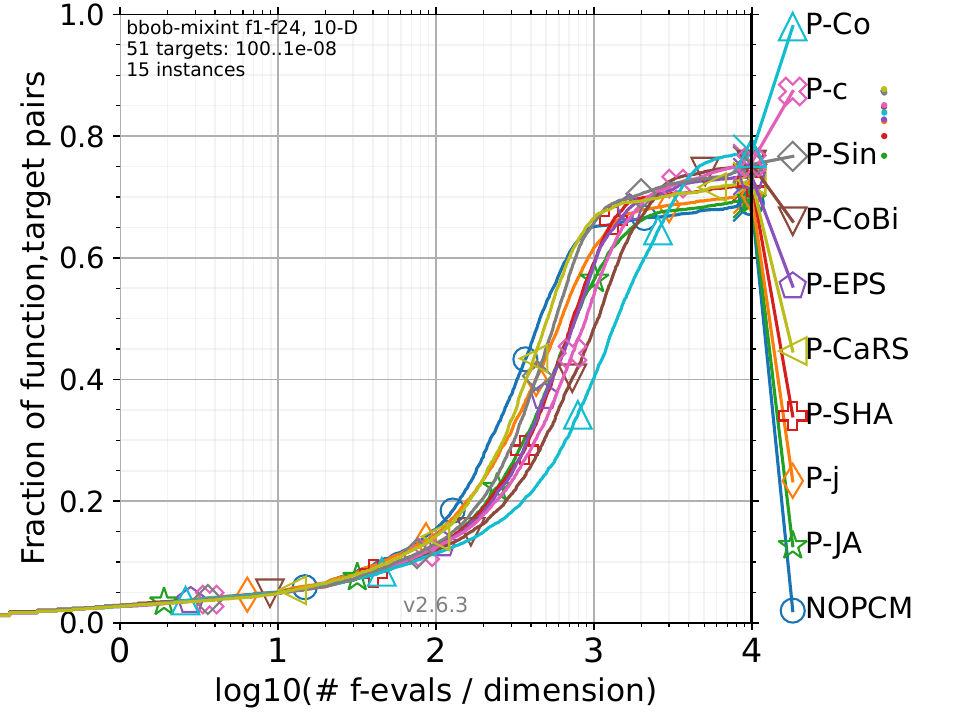}
}
\subfloat[rand-to-$p$best/1 ($n=80$)]{
\includegraphics[width=\width\textwidth]{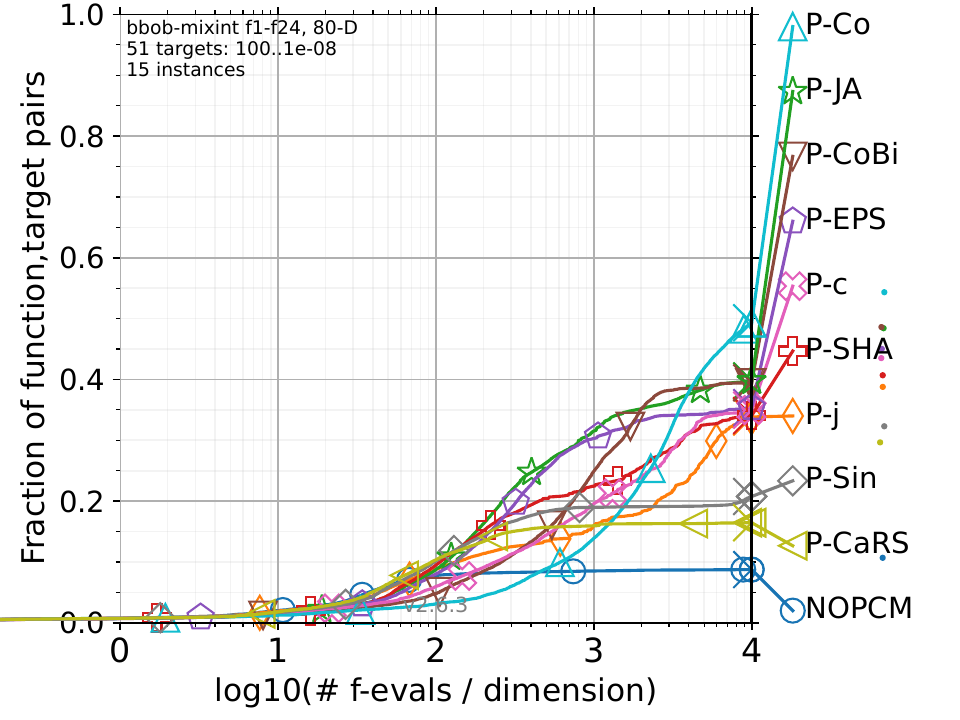}
}
\subfloat[rand-to-$p$best/1 ($n=160$)]{
\includegraphics[width=\width\textwidth]{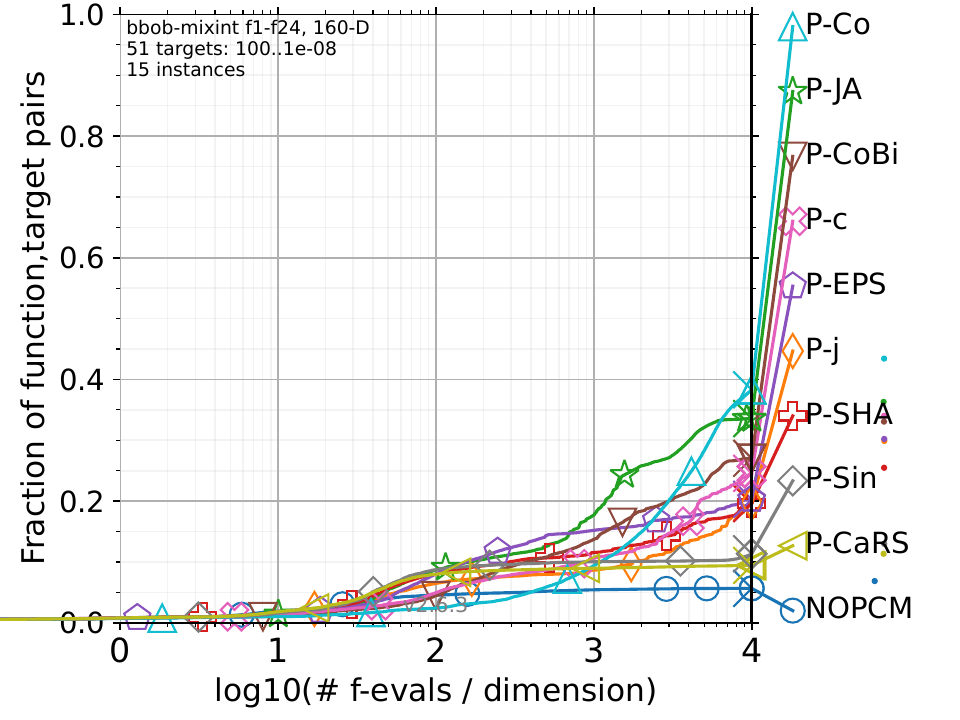}
}
\caption{Comparison of the nine PCMs and NOPCM using the rand-to-$p$best/1 strategy and Baldwinian repair method on the 24 \texttt{bbob-mixint} functions with $n \in \{10, 80, 160\}$.}
\label{fig:vs_de_Baldwin_rtp1}
%
\subfloat[rand/1 ($n=10$)]{
\includegraphics[width=\width\textwidth]{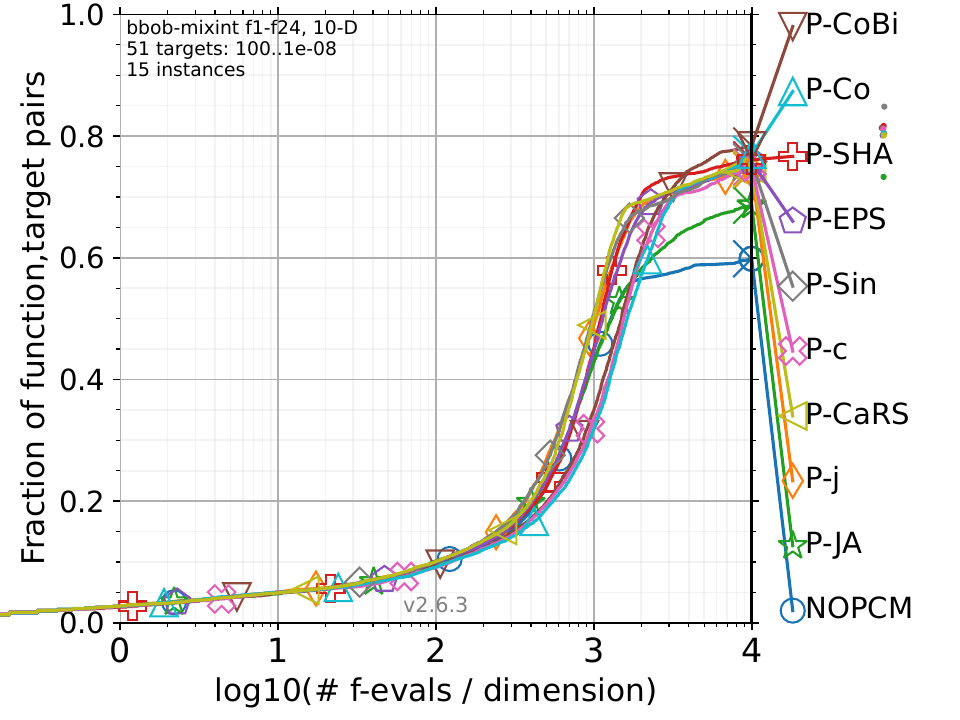}
}
\subfloat[rand/1 ($n=80$)]{
\includegraphics[width=\width\textwidth]{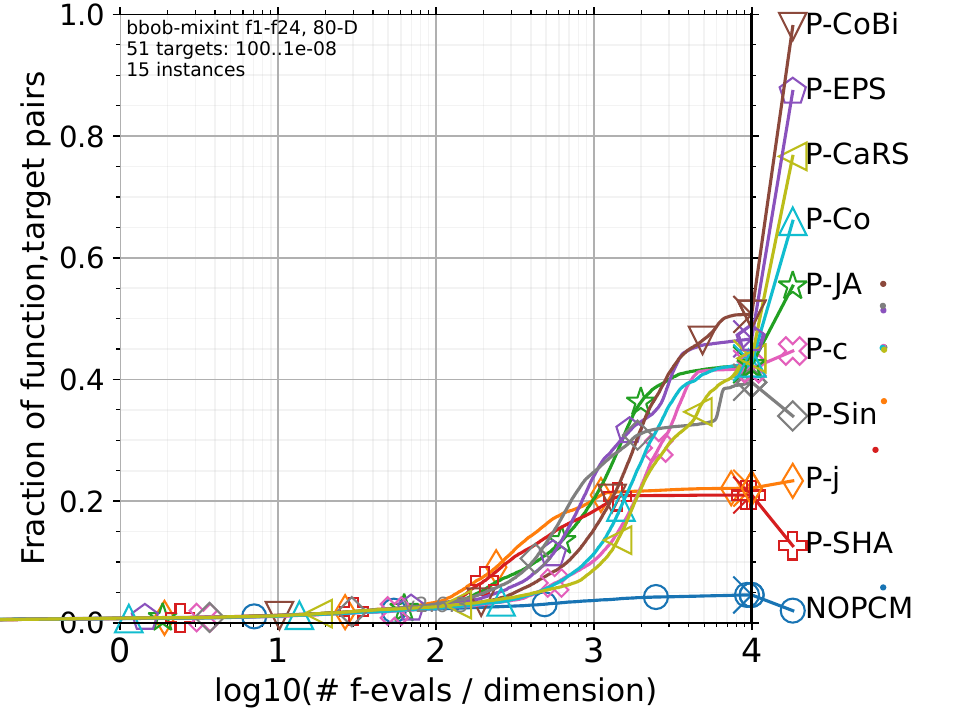}
}
\subfloat[rand/1 ($n=160$)]{
\includegraphics[width=\width\textwidth]{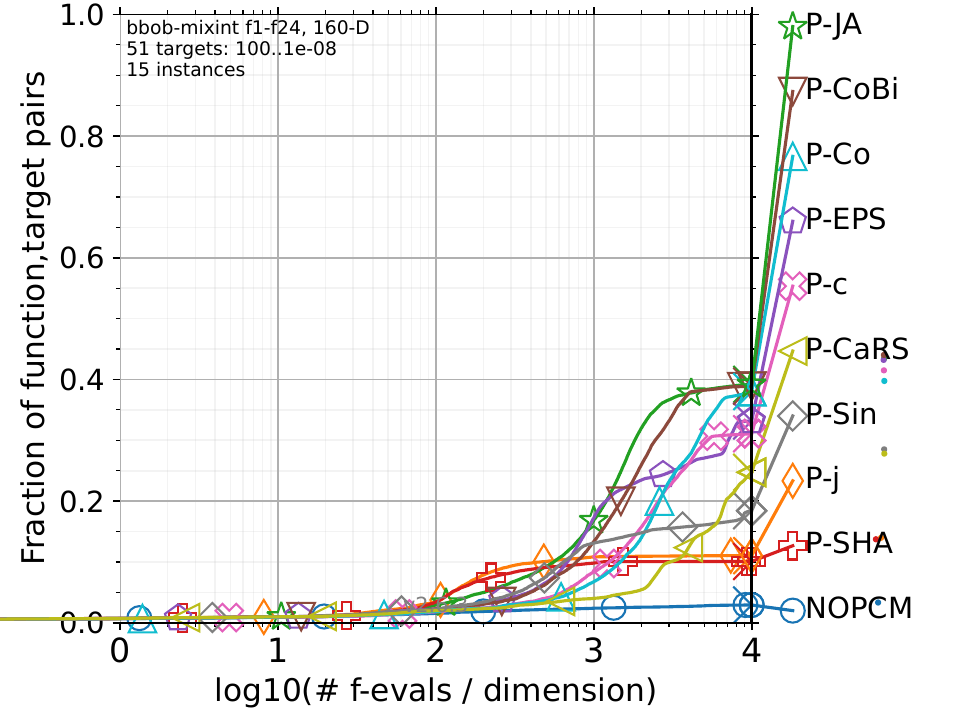}
}
\caption{Comparison of the nine PCMs and NOPCM using the rand/1 strategy and Lamarckian repair method on the 24 \texttt{bbob-mixint} functions with $n \in \{10, 80, 160\}$.}
   \label{fig:vs_de_Lamarckian}
\end{figure*}

Figures \ref{fig:vs_de_Baldwin_r1}--\ref{fig:vs_de_Lamarckian} show comparison of the 10 DE algorithms with the nine PCMs (\pref{sec:pcm}) and NOPCM (\pref{sec:setting}) on the 24 \texttt{bbob-mixint} functions for $n \in \{10, 80, 160\}$. 
Figures \ref{fig:vs_de_Baldwin_r1} and \ref{fig:vs_de_Baldwin_rtp1} show the results when using the rand/1 and rand-to-$p$best/1 strategies, respectively.
Here, the Baldwinian repair method is used in Figures \ref{fig:vs_de_Baldwin_r1} and \ref{fig:vs_de_Baldwin_rtp1}.
In contrast, \pref{fig:vs_de_Lamarckian} shows the results when using the rand/1 strategy and Lamarckian repair method.
Figures \ref{supfig:vs_de_rand_1_Baldwin}--\ref{supfig:vs_de_rand_to_pbest_1_Lamarckian} show all results of DE algorithms using the eight mutation strategies for $n \in \{5, 10, 20, 40, 80, 160\}$.

Figures \ref{fig:vs_de_Baldwin_r1}--\ref{fig:vs_de_Lamarckian} show the bootstrapped empirical cumulative distribution (ECDF) \cite{HansenARMTB21} based on the results on the 24 \texttt{bbob-mixint} functions with each $n$.
We used the COCO software~\cite{HansenARMTB21} to generate all ECDF figures in this paper.
Let $f_{\mathrm{target}} = f(\vec{x}^{*}) + f_{\Delta}$ be a target value to reach, where $\vec{x}^{*}$ is the optimal solution, and $f_{\Delta}$ is any one of 51 evenly log-spaced $f_{\Delta}$ values $\{10^{2}, 10^{1.8}, \ldots, 10^{-7.8}, 10^{-8}\}$.
Thus, 51 $f_{\mathrm{target}}$ values are available for each function instance, and $18\,360$ $f_{\mathrm{target}}$ values ($=51 \times 15 \times 24$) are available for all 15 instances of the 24 \texttt{bbob-mixint} functions.
In the ECDF figure, the vertical axis represents the proportion of $f_{\mathrm{target}}$ values reached by an optimizer within specified function evaluations.
The horizontal axis represents the number of function evaluations.
For example, Figure~\ref{fig:vs_de_Baldwin_r1}(b) shows that P-j solved about $35\%$ of the $18\,360$ $f_{\mathrm{target}}$ values within $10^3 \times n$ function evaluations for $n=80$.
Figure~\ref{fig:vs_de_Baldwin_r1}(b) shows that P-j is about ten times faster than P-Co to reach the same precision.


\textbf{Statistical significance} is tested with the rank-sum test for $f_{\Delta} \in \{10^{1}, 10^{0}, 10^{-1}, 10^{-2}, 10^{-3}, 10^{-5}, 10^{-7}\}$ by using COCO.
The statistical test results can be found in \url{https://zenodo.org/doi/10.5281/zenodo.10608500}.
Due to the paper length limitation, we do not describe the statistical test results, but they are consistent with the ECDF figures in most cases.

\pref{tab:best_pcm} shows the best PCMs on the 24 \texttt{bbob-mixint} functions with each $n$ in terms of the ECDF value at $10^4n$ evaluations when using each mutation strategy.
\pref{tab:best_pcm}(a) and (b) show the results when using the Baldwinian and Lamarckian repair methods, respectively.

\vspace{0.2em}
\noindent \textit{4.1.1 \hspace{1em} Comparison when using rand/1 and the Baldwinian repair method.}
As shown in \pref{fig:vs_de_Baldwin_r1}(a), when using the rand/1 strategy for $n=10$, NOPCM performs the best almost until $10^4n$ function evaluations.
P-CoBi performs slightly better than other PCMs exactly at $10^4n$ function evaluations.
These results suggest that DE with the rand/1 mutation strategy does not require any PCM for low dimension.
In fact, as shown in \pref{tab:best_pcm}(a), NOPCM is the best performer for $n=5$.
However, NOPCM performs poorly for $n\geq 20$.

As shown in Figures \ref{fig:vs_de_Baldwin_r1}(b) and (c), P-Sin performs the best for $n=80$ and $160$ at $10^4n$ function evaluations, followed by P-j.
Although P-j is outperformed by P-Sin at the end of the run, P-j shows the better anytime performance than other PCMs, including P-Sin.

\vspace{0.2em}
\noindent \textit{4.1.2 \hspace{1em} Comparison when using rand-to-$p$best/1 and the Baldwinian repair method.}
%
As shown in \pref{fig:vs_de_Baldwin_rtp1}, the rankings of the PCMs for the rand/1 and rand-to-$p$best/1 mutation strategies are totally different.
NOPCM performs the worst for any $n$.
Although P-Sin is the best for $n=160$ when using rand/1, P-Sin is the third worst when using rand-to-$p$best/1.
P-Co shows the second worst performance for Figures \ref{fig:vs_de_Baldwin_r1}(b)--(c), but P-Co performs the best for Figures \ref{fig:vs_de_Baldwin_rtp1}(b)--(c) at $10^4n$ function evaluations.


\begin{table}[t]
\renewcommand{\arraystretch}{0.8}  
\centering
  \caption{Best PCMs on the \texttt{bbob-mixint} suite based on the results when using each mutation strategy at $10^4 n$ function evaluations. In the table, ctb/1, ctr/1, ct$p$/1, and rt$p$/1 represent the current-to-best/1, current-to-rand/1, current-to-$p$best/1, and rand-to-$p$best/1 mutation strategies, respectively.}
\label{tab:best_pcm}
{\small
\scalebox{0.97}[1]{
\subfloat[Baldwinian repair method]{
\begin{tabular}{lccccccccccc}
\toprule
Strategy & $n=5$ & $n=10$ & $n=20$ & $n=40$ & $n=80$ & $n=160$\\
\midrule
rand/1 & NOPCM & P-CoBi & P-c & P-Sin & P-Sin & P-Sin\\
rand/2 & P-Sin & P-Sin & P-Sin & P-Sin & P-j & P-j\\
best/1 & P-Co & P-Co & P-Co & P-Co & P-Co & P-JA\\
best/2 & P-CoBi & P-CoBi & P-EPS & P-CoBi & P-c & P-c\\
ctb/1 & P-CoBi & P-Co & P-Co & P-CoBi & P-Co & P-JA\\
ctr/1 & P-CoBi & P-CoBi & P-CoBi & P-CoBi & P-CoBi & P-CoBi\\
ct$p$/1 & P-Co & P-CoBi & P-Co & P-Co & P-Co & P-Co\\
rt$p$/1 & P-Co & P-Co & P-Co & P-Co & P-Co & P-Co\\
\bottomrule
\end{tabular}
}
}
\\
\scalebox{0.98}[1]{
\subfloat[Lamarckian repair method]{
\begin{tabular}{lccccccccccc}
\toprule
Strategy & $n=5$ & $n=10$ & $n=20$ & $n=40$ & $n=80$ & $n=160$\\
\midrule
rand/1 & P-CoBi & P-CoBi & P-CoBi & P-CaRS & P-CoBi & P-JA\\
rand/2 & P-Sin & P-Sin & P-CoBi & P-Sin & P-CoBi & P-CoBi\\
best/1 & P-Co & P-Co & P-Co & P-Co & P-Co & P-Co\\
best/2 & P-CoBi & P-CoBi & P-Co & P-CaRS & P-Co & P-Co\\
ctb/1 & P-Co & P-CoBi & P-Co & P-Co & P-Co & P-Co\\
ctr/1 & P-CoBi & P-CoBi & P-Co & P-CoBi & P-CoBi & P-CoBi\\
ct$p$/1 & P-CoBi & P-CoBi & P-CoBi & P-Co & P-Co & P-Co\\
rt$p$/1 & P-CoBi & P-SHA & P-Co & P-Co & P-Co & P-Co\\
\bottomrule
\end{tabular}
}
}
}
%
\centering
  \caption{Top three configurations on the \texttt{bbob-mixint} suite for each $n$.
  ``B'' and ``L'' indicate the Baldwinian and Lamarckian repair methods, respectively. 
    }
\label{tab:best_config}
{\footnotesize
\scalebox{1}[1]{
\begin{tabular}{llllllll}
\toprule
$n$ & 1st & 2nd & 3rd\\
\midrule
$n=5$ & $\langle$P-CoBi, rand/1, L$\rangle$ & $\langle$P-CoBi, ct$p$/1, L$\rangle$ & $\langle$P-CoBi, ctr/1, L$\rangle$\\
$n=10$ & $\langle$P-CoBi, ctb/1, L$\rangle$ & $\langle$P-CoBi, rand/1, L$\rangle$ & $\langle$P-CoBi, ct$p$/1, L$\rangle$\\
$n=20$ & $\langle$P-Co, rt$p$/1, B$\rangle$ & $\langle$P-CoBi, rand/1, L$\rangle$ & $\langle$P-Co, ctr/1, L$\rangle$\\
$n=40$ & $\langle$P-CoBi, ctr/1, L$\rangle$ & $\langle$P-Sin, rand/2, B$\rangle$ & $\langle$P-CaRS, rand/1, L$\rangle$\\
$n=80$ & $\langle$P-Sin, rand/1, B$\rangle$ & $\langle$P-j, rand/2, B$\rangle$ & $\langle$P-CoBi, ctr/1, L$\rangle$\\
$n=160$ & $\langle$P-j, rand/2, B$\rangle$ & $\langle$P-Sin, rand/1, B$\rangle$ & $\langle$P-Co, rt$p$/1, L$\rangle$\\
\bottomrule
\end{tabular}
}
}
\end{table}

\vspace{0.2em}
\noindent \textit{4.1.3 \hspace{1em} Comparison when using the Lamarckian repair method.}
Interestingly, as shown in Figure \ref{fig:vs_de_Lamarckian}, the use of the Lamarckian repair method significantly deteriorates or improves the performance of some PCMs.
For example, as shown in Figures \ref{fig:vs_de_Baldwin_r1}(b)--(c) and \ref{fig:vs_de_Lamarckian}(b)--(c), the use of the Lamarckian repair method significantly deteriorates the performance of P-j with the rand/1 strategy for $n \in \{80, 160\}$.
In contrast, as seen from Figures \ref{fig:vs_de_Baldwin_r1} and \ref{fig:vs_de_Lamarckian}, the performance of P-Co is significantly improved by using the Lamarckian one.

\begin{figure*}[t]
\newcommand{\width}{0.315}
\centering
\subfloat[$n=10$]{
\includegraphics[width=\width\textwidth]{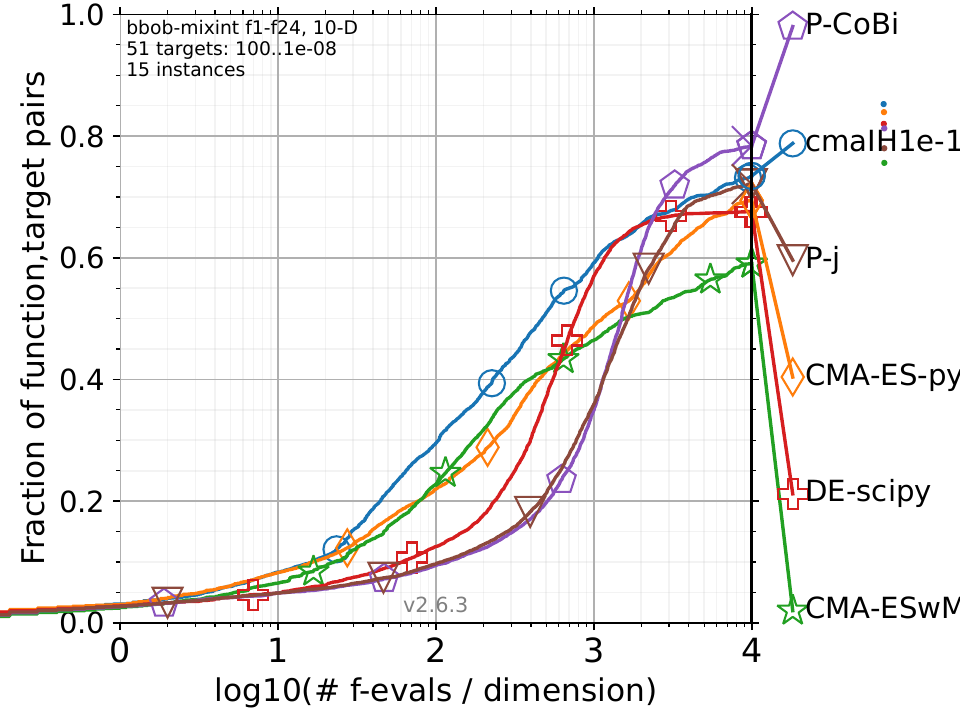}
}
\subfloat[$n=80$]{
\includegraphics[width=\width\textwidth]{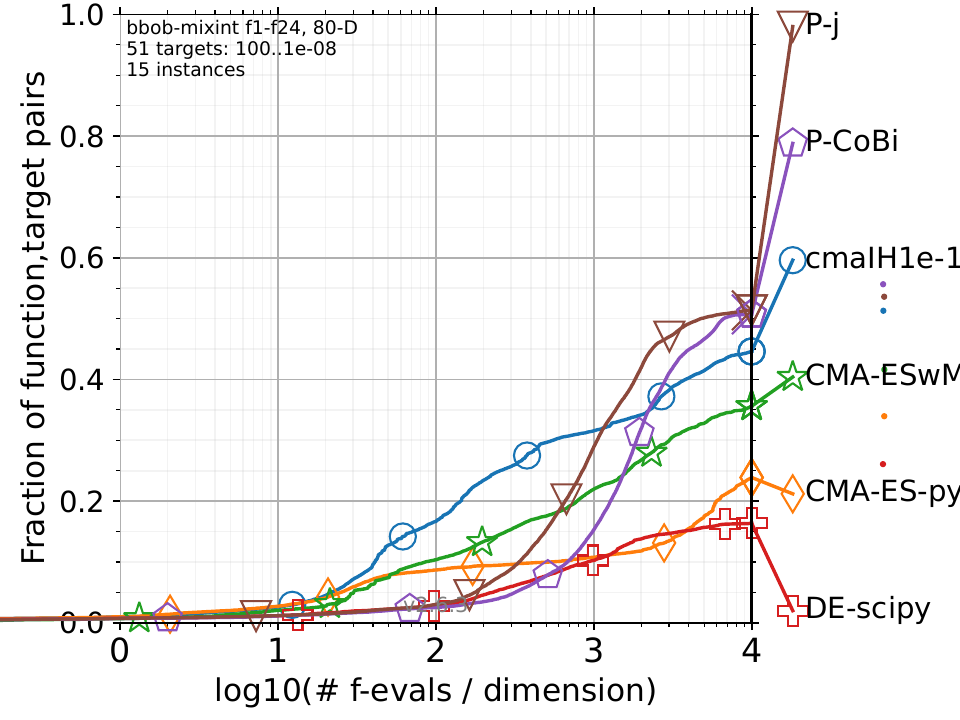}
}
\subfloat[$n=160$]{
\includegraphics[width=\width\textwidth]{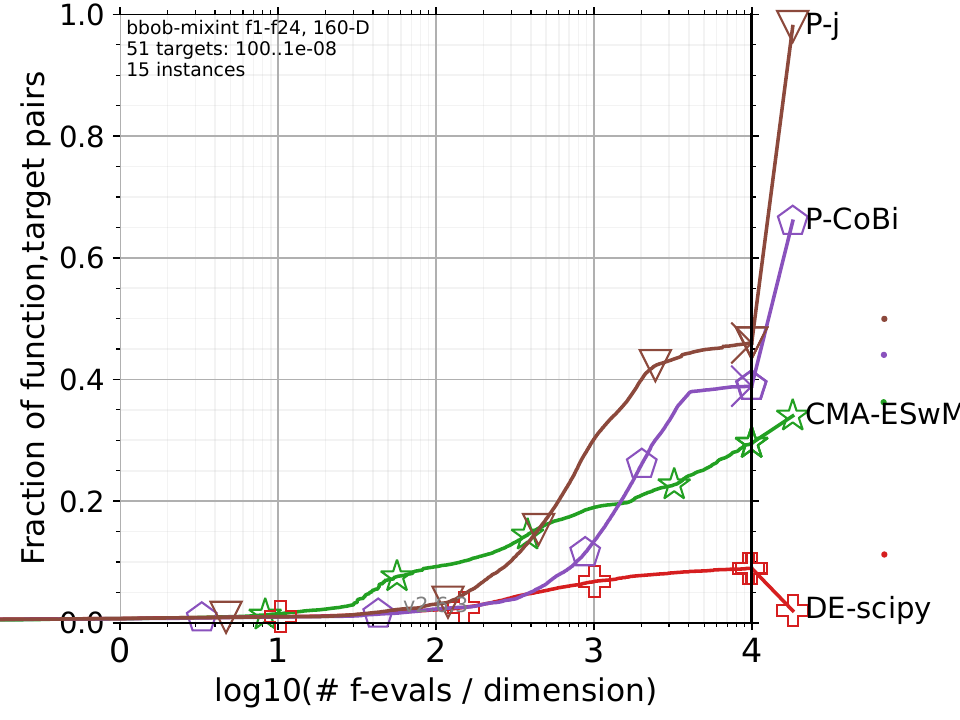}
}
\caption{Comparison of P-j and P-CoBi with the three CMA-ES variants on the 24 \texttt{bbob-mixint} functions with $n \in \{10, 80, 160\}$.}
\label{fig:vs_cma}
\end{figure*}

\vspace{0.2em}
\noindent \textit{4.1.4 \hspace{1em} Summary.}
As seen from \pref{tab:best_pcm}, with one exception, any one of the nine PCMs performs the best for each case.
Although most previous studies (e.g., \cite{LampinenZ99Part1,Liao10,LinLCZ18,LiuWHJ22}) did not use any PCMs, our observation suggests the importance of PCMs in DE for mixed-integer black-box optimization.

As shown in \pref{tab:best_pcm}, the best PCM significantly depends on the combination of the mutation strategy and method.
Roughly speaking, P-CoBi and P-Co perform the best in many cases, followed by P-Sin.
P-CoBi works especially well for low dimensions, i.e., $n \in \{5, 10\}$.
P-CaRS, P-EPS, P-c, P-j, P-JA, and P-SHA show the best performance in a few cases.
\pref{tab:best_pcm} suggests that P-Co is suitable when using the Lamarckian repair method and exploitative mutation strategies, including best/1, best/2, best/1, current-to-best/1, current-to-$p$best/1, and rand-to-$p$best/1.

Although the previous study~\cite{TanabeF20} reported the \emph{poor} performance of P-Co for \emph{numerical} black-box optimization, our results show the \emph{excellent} performance of P-Co for \emph{mixed-integer} black-box optimization.
In contrast, P-SHA is one of the state-of-the-art PCMs in DE~\cite{TanabeF20}.
P-SHA has also been used in state-of-the-art DE algorithms, including a number of L-SHADE-based algorithms~\cite{TanabeF14,BrestMB17}.
However, as shown in Figures \ref{fig:vs_de_Baldwin_r1}--\ref{fig:vs_de_Lamarckian}, P-SHA perform well on the \texttt{bbob-mixint} suite only at the early stage of the search.
As seen from \pref{tab:best_pcm}, P-SHA performs the best in only one case.
This observation suggests that replacing P-SHA with P-Co, P-CoBi, or P-j may improve the performance of L-SHADE$_{\text{ACO}}$~\cite{LinDS17}.

\vspace{0.2em}
\noindent \textit{4.1.5 \hspace{1em} On the best configuration.}
Although we focus on PCMs, it is interesting to discuss which configuration performs the best.
\pref{tab:best_config} shows the top 3 out of 160 DE configurations on the \texttt{bbob-mixint} suite for each $n$, where the 160 configurations include the 9 PCMs and NOPCM, 8 mutation strategies, and the two repair methods ($10 \times 8 \times 2 = 160$).
In \pref{tab:best_config}, a tuple represents a DE configuration that consists of a PCM, mutation strategy, and repair method.
Similar to the above discussion, as seen from \pref{tab:best_config}, the configurations including P-CoBi and P-Co perform well for low dimensions.
In contrast, the configurations including P-j and P-Sin show the best performance for high dimensions.
Although the Lamarckian repair method is included in most of the top three configurations, our results show that the Baldwinian repair method is suitable for P-Sin and P-j, especially for high dimensions.
Thus, there is no clear winner between the two repair methods.
Interestingly, the classical rand/1 and rand/2 strategies are included in 9 out of the top 18 configurations.
Since the best/1 and best/2 strategies are not found in \pref{tab:best_config}, we can say that too exploitative mutation strategies are not suitable.
This may be because the \texttt{bbob-mixint} functions have many plateaus from the point of view of DE due to the use of the rounding operator. 
Although \textit{the behavior analysis of DE is beyond the scope of this paper}, it is an avenue for future research".

\begin{tcolorbox}[title=Answers to RQ1, sharpish corners, top=2pt, bottom=2pt, left=4pt, right=4pt, boxrule=0.5pt]
Although most previous studies (e.g., \cite{LampinenZ99Part1,Liao10,LinLCZ18,LiuWHJ22}) did not use any PCM, our results demonstrated the importance of PCMs in DE for mixed-integer black-box optimization.
We found that the best PCM significantly depends on the combination of the mutation strategy and repair method.
Some of our results are inconsistent with the results shown in \cite{TanabeF20} for numerical black-box optimization.
For example, we demonstrated that P-Co performs significantly better than other PCMs on the \texttt{bbob-mixint} suite, especially when using the Lamarckian repair method and exploitative mutation strategies.
In contrast, our results show the unsuitability of P-SHA, one of the state-of-the-art PCMs, for mixed-integer black-box optimization.

\end{tcolorbox}

\subsection{Comparison with CMA-ES}
\label{sec:vs_sota}


As mentioned in \pref{sec:introduction}, the previous studies~\cite{TusarBH19,HamanoSNS22a,MartySAHH23} demonstrated that the extended versions of CMA-ES outperform DE with no PCM (\texttt{DE-scipy}~\cite{TusarBH19} described later).
However, the results in \pref{sec:vs_pcms} show that the use of PCM can significantly improve the performance of DE.
Thus, it is interesting to compare DE with a suitable PCM with the CMA-ES variants.


We consider the comparison with the following three extensions of CMA-ES: \texttt{CMA-ES-pycma}~\cite{TusarBH19}, \texttt{CMA-ESwM}~\cite{HamanoSNS22a}, and \texttt{cmaIH1e-1}~\cite{MartySAHH23}.
Both \texttt{CMA-ES-pycma} and \texttt{cmaIH1e-1} are the pycma~\cite{hansen2019pycma} implementations of CMA-ES with simple integer handling.
However, the pycma version of \texttt{cmaIH1e-1} is newer than that of \texttt{CMA-ES-pycma}.
Although the previous study~\cite{MartySAHH23} investigated three versions of CMA-ES, its results showed that \texttt{cmaIH1e-1} was the best performer among them.
\texttt{CMA-ESwM} is the CMA-ES with margin~\cite{HamanoSNS22}, which uses a lower bound on the marginal probability for each integer variable.
In addition, we consider the SciPy version of DE (\texttt{DE-scipy}~\cite{TusarBH19}). 
We used the benchmarking results of the four optimizers provided by the COCO data archive (\url{https://numbbo.github.io/data-archive}).


\pref{fig:vs_cma} shows the comparison of two DE algorithms with P-j and P-CoBi with the three CMA-ES variants on the 24 \texttt{bbob-mixint} functions for $n \in \{10, 80, 160\}$.
Here, the benchmarking data of \texttt{CMA-ES-pycma} and \texttt{cmaIH1e-1} for $n=160$ are not available. 
In \pref{fig:vs_cma}, P-j uses the rand/2 strategy and Baldwinian repair method, and P-CoBi uses the rand/1 strategy and Lamarckian repair method.
As shown in \pref{tab:best_config}, P-CoBi and P-j with these configurations perform the best for $n=5$ and $n=160$, respectively.
\pref{supfig:vs_cma} shows the results for all $n$, where it is similar to \pref{fig:vs_cma}.

As shown in \pref{fig:vs_cma}, for all $n$, P-j and P-CoBi are outperformed by the CMA-ES variants for smaller budgets.
In contrast, P-j and P-CoBi perform significantly better than the CMA-ES variants for larger budgets.
Although the configuration of P-CoBi is suitable for low dimensions, it outperforms the CMA-ES variants even for $n \geq 80$.
As expected, the performance of P-j and P-CoBi is significantly better than that of \texttt{DE-scipy} for high dimensions.

\begin{figure*}[t]
\newcommand{\width}{0.235}
\centering
\subfloat[Error value]{
\includegraphics[width=\width\textwidth]{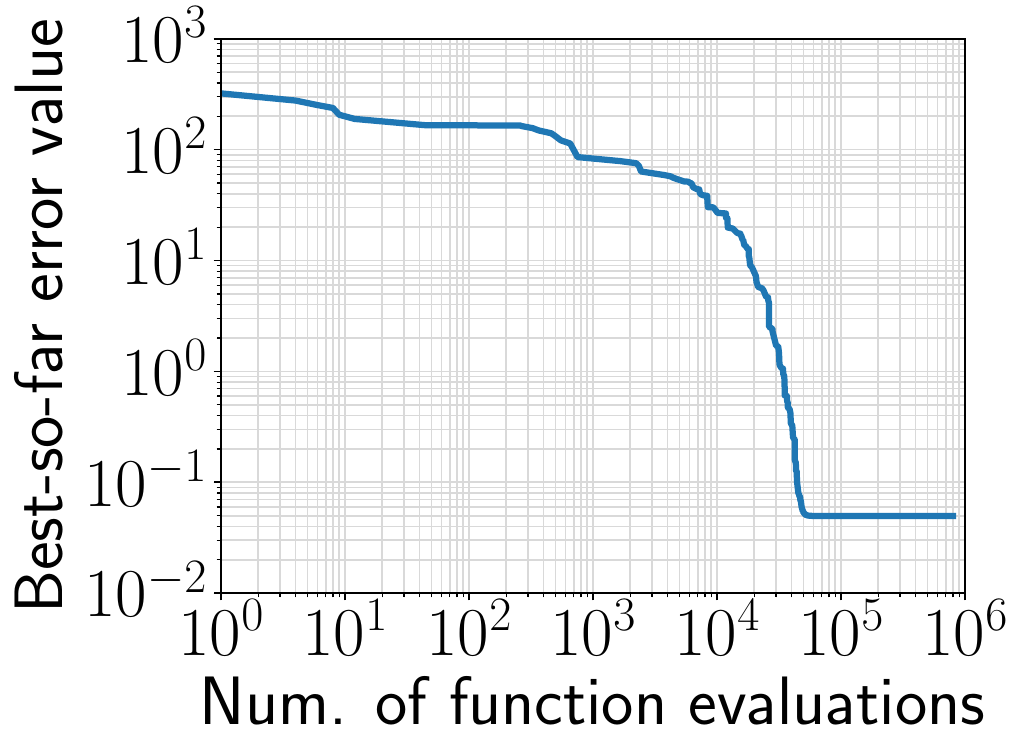}
}
\subfloat[\texttt{div} (left) and \texttt{nsame} (right) values]{
\includegraphics[width=0.26\textwidth]{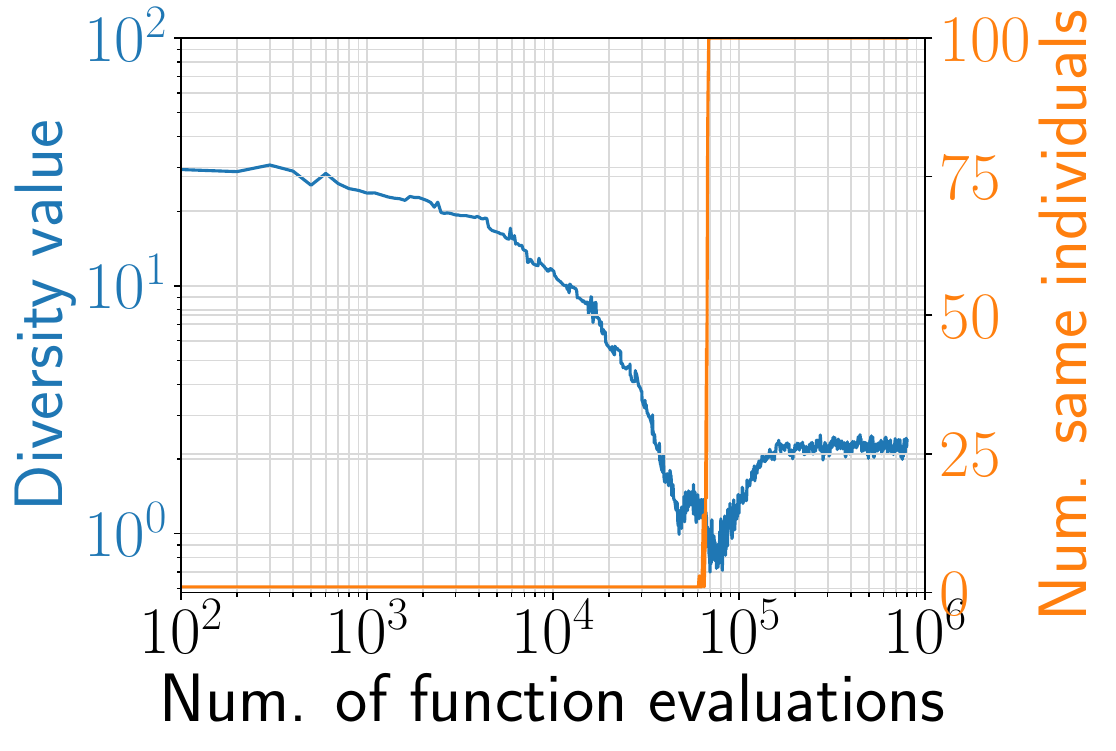}
}
\subfloat[Memories $\vec{m}_s$ and $\vec{m}_c$]{
\includegraphics[width=\width\textwidth]{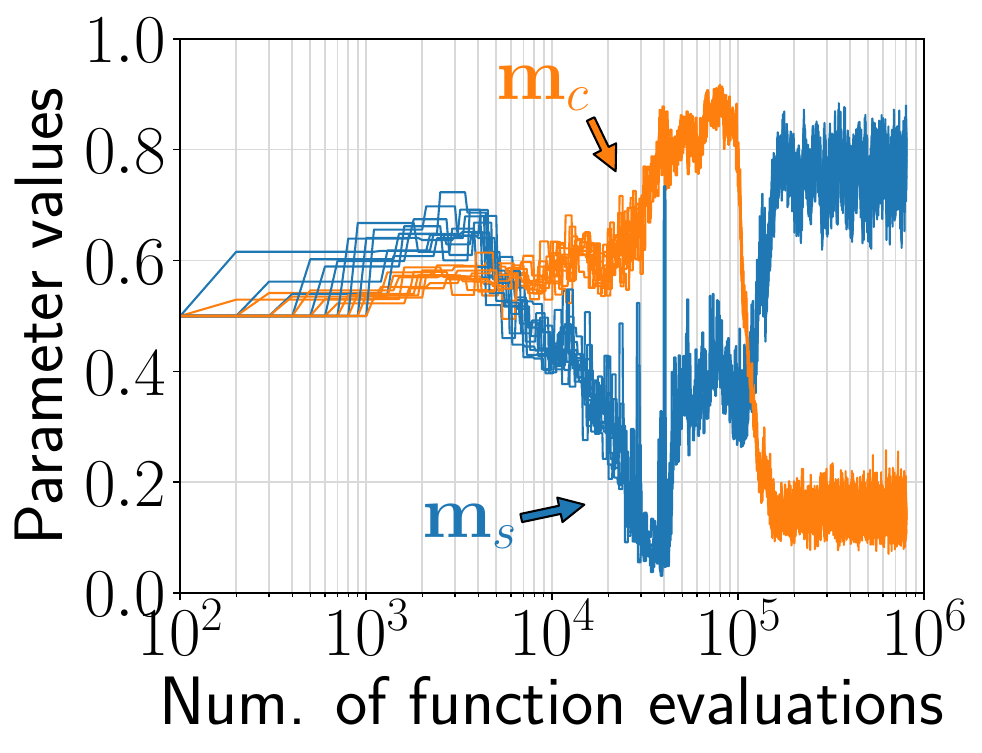}
}
\subfloat[Mean succ. parameters]{
\includegraphics[width=\width\textwidth]{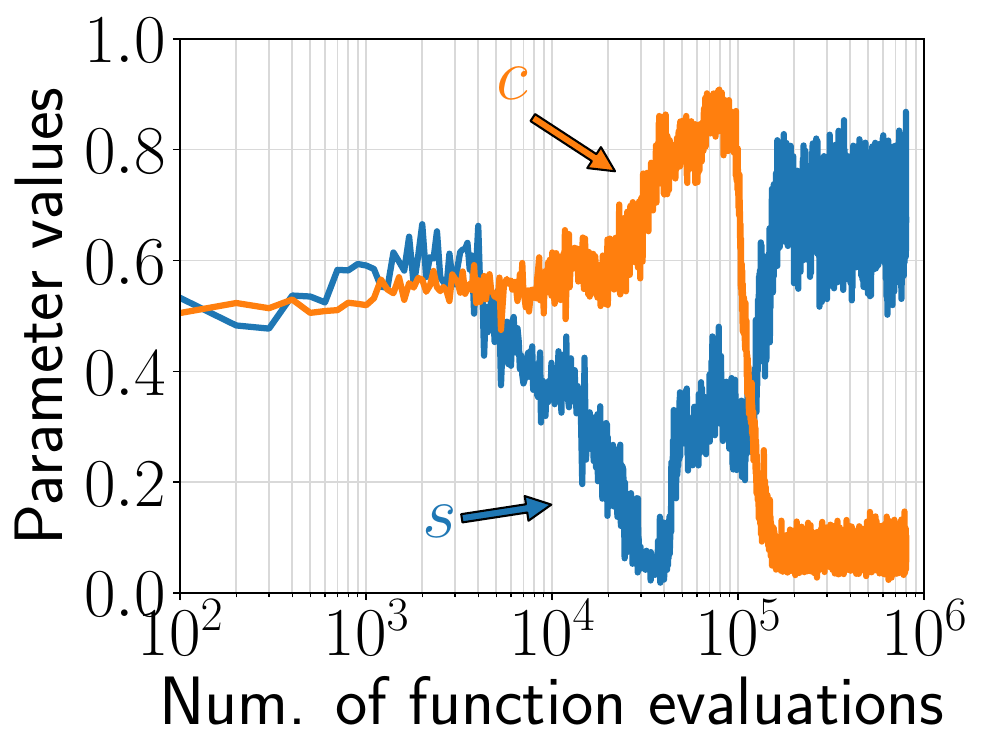}
}
\caption{Analysis results of a typical single run of P-SHA on $f_3$ with $n=80$.}
\label{fig:analysis_psha}
\end{figure*}

\pref{supfig:vs_cma_each_group}(a)--(e) show the comparison on the five function groups for $n=80$, respectively.
As expected, P-j and P-CoBi perform well on the separable functions ($f_1, ...,  f_5$).
In addition, P-CoBi shows the best performance on the functions with high conditioning ($f_{10}, ..., f_{14}$) and multimodal functions with weak global structure ($f_{20}, ...,  f_{24}$) at $10^4$ function evaluations.
P-j also performs the best the functions with low conditioning ($f_{6}, ..., f_{9}$) at $10^4$ function evaluations.
Similar to \pref{fig:vs_cma}, for any function group, the CMA-ES variants outperform P-j and P-CoBi within about $10^3 n$ function evaluations.
In addition, \texttt{cmaIH1e-1} is the best performer on the multimodal functions with adequate global structure ($f_{15}, ..., f_{19}$).
Thus, no optimizer dominates others on any function at any time.
These observations suggest that automated algorithm selection~\cite{KerschkeT19,KerschkeHNT19} with an algorithm portfolio consisting of DE and CMA-ES is a promising approach for mixed-integer black-box optimization.


\begin{tcolorbox}[title=Answers to RQ2, sharpish corners, top=2pt, bottom=2pt, left=4pt, right=4pt, boxrule=0.5pt]

Our results show that DE algorithms with suitable PCMs (P-j and P-CoBi) can perform significantly better than the three CMA-ES variants with integer handling on the \texttt{bbob-mixint} suite, especially for high dimensions and larger budgets of function evaluations.
However, we observed that the DE algorithms are generally outperformed by the CMA-ES variants for smaller budgets of function evaluations and some particular \texttt{bbob-mixint} functions.
This complementarity between DE and CMA-ES suggests a promising possibility of automated algorithm selection.

\end{tcolorbox}

\subsection{How does P-SHA fail?}
\label{sec:analysis_pcm_shade}

Despite the high performance of P-SHA for numerical black-box optimization, the results in \pref{sec:vs_pcms} indicate the poor performance of P-SHA on the \texttt{bbob-mixint} suite.
This section investigates the behavior of P-SHA to find out why it did not work well.


\pref{fig:analysis_psha} shows some analysis results of P-SHA with the rand/1 strategy and Baldwinian repair method on $f_3$ (the separable Rastrigin function) with $n=80$.
Since $f_3$ is separable, it is easy for DE to solve $f_3$.
Nevertheless, P-SHA found the optimal solution in only 3 out of 15 runs.
P-SHA also shows the third worst performance.
For the sake of reference, \pref{supfig:analysis_pja} shows the results of P-JA, which found the optimal solution in all 15 runs.

\pref{fig:analysis_psha}(a) shows the error value $|f(\vec{x}_{\mathrm{bsf}}) - f(\vec{x}^*)|$ of the best solution found so far $\vec{x}_{\mathrm{bsf}}$ by P-SHA in a typical single run.
As seen from \pref{fig:analysis_psha}(a), the improvement of $\vec{x}_{\mathrm{bsf}}$ stops at about $92\,000$ function evaluations.


\pref{fig:analysis_psha}(b) shows a diversity value (\texttt{div}) and the number of individuals with the same objective value (\texttt{nsame}) as the best individual $\vec{x}_{\mathrm{best}} = \argmin_{\vec{x} \in \set{P}} f(\vec{x})$ in the population $\set{P}$ for each iteration.
Here, we calculated the \texttt{div} and \texttt{nsame} values of $\set{P}$ as follows: $\texttt{div}(\set{P})=\frac{1}{\mu}\sum_{\vec{x} \in \set{P} \setminus \{\vec{x}_{\mathrm{best}}\}} \|\vec{x} - \vec{x}_{\mathrm{best}} \|$ and $\texttt{nsame}(\set{P}) = |\{\vec{x} \, | \, \vec{x} \in \set{P} \: \text{s.t.} $ $\: f(\vec{x}) = f(\vec{x}_{\mathrm{best}})\}|$.
A small $\texttt{div}(\set{P})$ value means that most individuals in $\set{P}$ are close to the best individual $\vec{x}_{\mathrm{best}}$ in the solution space.
A small $\texttt{nsame}(\set{P})$ value means that most individuals in $\set{P}$ are at a plateau induced by the rounding operator.
Since $\texttt{div}$ and $\texttt{nsame}$ can be calculated only after $\mu$ function evaluations, \pref{fig:analysis_psha}(b) starts from $\mu$ function evaluations, where $\mu=100$.
On the one hand, as seen from \pref{fig:analysis_psha}(b), the $\texttt{nsame}$ value suddenly becomes $100$ at about $92\,000$ function evaluations. 
Here, $\texttt{nsame}(\set{P})=100$ means that all $100$ individuals in $\set{P}$ are in the same plateau.
On the other hand, the $\texttt{div}$ value increases after the $\texttt{nsame}$ value becomes $100$.
These results indicate that individuals in $\set{P}$ explore the solution space even after getting stuck on a plateau in the objective space.


\pref{fig:analysis_psha}(c) shows the 10 elements of the two memories $\vec{m}_s$ and $\vec{m}_c$ for the adaptation of the scale factor $s$ and crossover rate $c$, respectively.
\pref{fig:analysis_psha}(d) also shows the mean of successful $s$ and $c$ values for each iteration. 
On the separable Rastrigin function, adaptive PCMs generally generate large $s$ and small $c$ values to handle the multimodality and exploit the separability~\cite{BrestGBMZ06,ZhangS09,TanabeF13b}.
In fact, as seen from \pref{supfig:analysis_pja}(c), P-JA adjusts $m_s$ and $m_c$ to large and small values during the search, respectively.
However, as shown in \pref{fig:analysis_psha}(c), P-SHA adjusts $\vec{m}_s$ and $\vec{m}_c$ to small and large values \emph{until} about $92\,000$ function evaluations, respectively.
This kind of parameter adaptation can be found when addressing unimodal functions, e.g., the Sphere function~\cite{BrestGBMZ06,ZhangS09,TanabeF13b}.
Thus, this adaptation of $s$ and $c$ in P-SHA fails on $f_3$. 
This failed parameter adaptation causes the stagnation of the search as shown in Figures \ref{fig:analysis_psha}(a) and (b).
Interestingly, as seen from \pref{fig:analysis_psha}(c), P-SHA correctly adjusts $\vec{m}_s$ and $\vec{m}_c$ to large and small values \emph{after} about $92\,000$ function evaluations, respectively.
Since the population has already stagnated at a plateau after about $92\,000$ function evaluations, this improvement of parameter adaptation in P-SHA is too late.



A previous study~\cite{TanabeF14gecco} showed the pathological behavior of some adaptive PCMs on functions whose search space characteristics of variables are different from each other.
The \texttt{bbob-mixint} functions can be considered to be the same as those kinds of functions investigated in \cite{TanabeF14gecco} due to the existence of integer variables.
In addition, P-SHA has a high tracking performance with respect to successful parameters~\cite{TanabeF17}.
These reasons suggest that P-SHA was particularly influenced by the properties of the \texttt{bbob-mixint} functions.
%




\begin{tcolorbox}[title=Answers to RQ3, sharpish corners, top=2pt, bottom=2pt, left=4pt, right=4pt, boxrule=0.5pt]

We showed the failed parameter adaptation of the two memories $\vec{m}_s$ and $\vec{m}_c$ in P-SHA on $f_3$ (the separable Rastrigin function), which causes the stagnation of the search.
Interestingly, the parameter adaptation in P-SHA works well after all individuals in the population reach the same plateau.
Although most PCMs can address $f_3$ easily, P-SHA struggles on $f_3$.
This is mainly due to the property of the \texttt{bbob-mixint} functions and the high tracking performance of P-SHA.

\end{tcolorbox}

\section{Conclusion}
\label{sec:conclusion}

We have investigated the performance of the 9 PCMs in DE with the 8 mutation strategies and 2 repair methods on the 24 \texttt{bbob-mixint} functions.
Although most previous studies on DE for mixed-integer black-box optimization did not use any PCM, our results show that the use of PCMs can significantly improve the performance of DE (\pref{sec:vs_pcms}).
We have demonstrated that the best PCM depends on the choice of the mutation strategy and the repair method.
Unlike the results for numerical black-box optimization reported in ~\cite{TanabeF20}, our results show that P-SHA is not suitable for mixed-integer black-box optimization.
In contrast, we observed that some simple PCMs (e.g., P-Co, P-CoBi, P-j, and P-Sin) work well on the \texttt{bbob-mixint} suite.
We have also shown that the DE algorithms with suitable PCMs perform significantly better than the CMA-ES variants with integer handling for larger budgets of function evaluations (\pref{sec:vs_sota}).
Finally, we have investigated how the parameter adaptation in P-SHA fails (\pref{sec:analysis_pcm_shade}).



We believe that our findings contribute to the design of an efficient DE algorithm for mixed-integer black-box optimization.
For example, our results suggest the promise of incorporating either P-Co, P-CoBi, P-j, or P-Sin into a new DE algorithm.
%
Fitness landscape analysis on the \texttt{bbob-mixint} suite is necessary for a better understanding of the behavior of DE algorithms.
Decomposed components in this work can be straightforwardly used for automatic configuration \cite{LopezIbanez16} of DE.
Algorithm selection for mixed-integer black-box optimization is also promising based on our observation of the complementarity between DE and CMA-ES.


\begin{acks}
  This work was supported by JSPS KAKENHI Grant Number \seqsplit{23H00491}.

\end{acks}

\bibliographystyle{ACM-Reference-Format}
\bibliography{reference} 


\clearpage

\begin{figure*}
\centering
\fontsize{30pt}{100pt}\selectfont{\textbf{Supplement}}
\end{figure*}




\appendix

\setcounter{figure}{0}
\setcounter{table}{0}

\renewcommand{\thesection}{S.\arabic{section}}
\renewcommand{\thetable}{S.\arabic{table}}
\renewcommand{\thefigure}{S.\arabic{figure}}
\renewcommand\thealgocf{S.\arabic{algocf}} 
\renewcommand{\theequation}{S.\arabic{equation}}

\makeatletter
\renewcommand{\@biblabel}[1]{[S.#1]}
\renewcommand{\@cite}[1]{[S.#1]}
\makeatother

\definecolor{c1}{RGB}{150,150,150}
\definecolor{c2}{RGB}{220,220,220}

\clearpage

\IncMargin{0.5em}
\begin{algorithm*}[t]
\small
\SetSideCommentRight
$t \leftarrow 1$, initialize $\set{P} =\left\{ \vec{x}_1, \ldots, \vec{x}_{\mu}\right\}$ randomly, $\set{A} \leftarrow \emptyset$ \;
%
\While{The termination criteria are not met}{
  \For{$i \in \{1, ..., \mu\}$}{
    $\vec{v}_{i} \leftarrow$ Apply differential mutation with $s$ to individuals in $\set{P}$\;
    $\vec{u}_{i} \leftarrow$ Apply binomial crossover with $c$ to $\vec{x}_{i}$ and $\vec{v}_{i}$\;
  }
  \For{$i \in \{1, \ldots, \mu\}$}{
    \If{$f(\vec{u}_{i}) \leq f(\vec{x}_{i})$} {
      $\set{A} \leftarrow \set{A} \cup \{\vec{x}_{i}\}$\;
      $\vec{x}_{i} \leftarrow \vec{u}_{i}$\;
    }
  }
  \lIf{$|\set{A}| > $ $a$} {
    Delete randomly selected individuals in $\set{A}$ unless $|\set{A}| < a$
  }
  $t \leftarrow t+1$\;
}
\caption{The basic DE algorithm with no PCM (NOPCM)}
\label{alg:de_no}
\end{algorithm*}\DecMargin{0.5em}

\IncMargin{0.5em}
\begin{algorithm*}[t]
\small
\SetSideCommentRight
$t \leftarrow 1$, initialize $\set{P} =\left\{ \vec{x}_1, \ldots, \vec{x}_{\mu}\right\}$ randomly, $\set{A} \leftarrow \emptyset$ \;
%
\While{The termination criteria are not met}{
  \For{$i \in \{1, ..., \mu\}$}{
    $\langle s_i, c_i\rangle \leftarrow$ Randomly select one from three parameter pairs: $\langle 1, 0.1\rangle$, $\langle 1, 0.9\rangle$, and $\langle 0.8, 0.2\rangle$.    
  }
  \For{$i \in \{1, ..., \mu\}$}{
    $\vec{v}_{i} \leftarrow$ Apply differential mutation with $s_i$ to individuals in $\set{P}$\;
    $\vec{u}_{i} \leftarrow$ Apply binomial crossover with $c_i$ to $\vec{x}_{i}$ and $\vec{v}_{i}$\;
  }
  \For{$i \in \{1, \ldots, \mu\}$}{
    \If{$f(\vec{u}_{i}) \leq f(\vec{x}_{i})$} {
      $\set{A} \leftarrow \set{A} \cup \{\vec{x}_{i}\}$\;
      $\vec{x}_{i} \leftarrow \vec{u}_{i}$\;
    }
  }
  \lIf{$|\set{A}| > a$} {
    Delete randomly selected $|\set{A}| - a$ individuals in $\set{A}$
  }
  $t \leftarrow t+1$\;
}
\caption{The basic DE algorithm with P-Co}
\label{alg:de_pco}
\end{algorithm*}\DecMargin{0.5em}

\IncMargin{0.5em}
\begin{algorithm*}[t]
\small
\SetSideCommentRight
$t \leftarrow 1$, initialize $\set{P} =\left\{ \vec{x}_1, \ldots, \vec{x}_{\mu}\right\}$ randomly, $\set{A} \leftarrow \emptyset$ \;
%
\While{The termination criteria are not met}{
  $s \leftarrow \frac{1}{2} \left(\frac{t}{t^{\rm max}} ({\sin}(2 \pi \omega t)) + 1 \right)$\;
  $c \leftarrow \frac{1}{2} \left(\frac{t}{t^{\rm max}} ({\sin}(2 \pi \omega t + \pi)) + 1 \right)$\;  
  \For{$i \in \{1, ..., \mu\}$}{
    $\vec{v}_{i} \leftarrow$ Apply differential mutation with $s$ to individuals in $\set{P}$\;
    $\vec{u}_{i} \leftarrow$ Apply binomial crossover with $c$ to $\vec{x}_{i}$ and $\vec{v}_{i}$\;
  }
  \For{$i \in \{1, \ldots, \mu\}$}{
    \If{$f(\vec{u}_{i}) \leq f(\vec{x}_{i})$} {
      $\set{A} \leftarrow \set{A} \cup \{\vec{x}_{i}\}$\;
      $\vec{x}_{i} \leftarrow \vec{u}_{i}$\;
    }
  }
  \lIf{$|\set{A}| > a$} {
    Delete randomly selected $|\set{A}| - a$ individuals in $\set{A}$
  }
  $t \leftarrow t+1$\;
}
\caption{The basic DE algorithm with P-Sin}
\label{alg:de_psin}
\end{algorithm*}\DecMargin{0.5em}

\IncMargin{0.5em}
\begin{algorithm*}[t]
\small
\SetSideCommentRight
$t \leftarrow 1$, initialize $\set{P} =\left\{ \vec{x}_1, \ldots, \vec{x}_{\mu}\right\}$ randomly, $\set{A} \leftarrow \emptyset$ \;
%
\While{The termination criteria are not met}{
  \lFor{$i \in \{1, ..., \mu\}$}{
    $s_i \leftarrow \mathrm{randu}[0.5, 0.55]$
  }
  $c \leftarrow$ Randomly select one from $\{0.5, 0.6, 0.7, 0.8, 0.9\}$\;
  \For{$i \in \{1, ..., \mu\}$}{
    $\vec{v}_{i} \leftarrow$ Apply differential mutation with $s_i$ to individuals in $\set{P}$\;
    $\vec{u}_{i} \leftarrow$ Apply binomial crossover with $c$ to $\vec{x}_{i}$ and $\vec{v}_{i}$\;
  }
  \For{$i \in \{1, \ldots, \mu\}$}{
    \If{$f(\vec{u}_{i}) \leq f(\vec{x}_{i})$} {
      $\set{A} \leftarrow \set{A} \cup \{\vec{x}_{i}\}$\;
      $\vec{x}_{i} \leftarrow \vec{u}_{i}$\;
    }
  }
  \lIf{$|\set{A}| > a$} {
    Delete randomly selected $|\set{A}| - a$ individuals in $\set{A}$
  }
  $t \leftarrow t+1$\;
}
\caption{The basic DE algorithm with P-CaRS}
\label{alg:de_pcars}
\end{algorithm*}\DecMargin{0.5em}

\IncMargin{0.5em}
\begin{algorithm*}[t]
\small
\SetSideCommentRight
$t \leftarrow 1$, initialize $\set{P} =\left\{ \vec{x}_1, \ldots, \vec{x}_{\mu}\right\}$ randomly, $\set{A} \leftarrow \emptyset$ \;
\lFor{$i \in \{1, ..., \mu\}$}{
  $s_i \leftarrow 0.5$,   $c_i \leftarrow 0.9$
}
%
\While{The termination criteria are not met}{
  \For{$i \in \{1, ..., \mu\}$}{
    \lIf{$\mathrm{randu}[0,1] < \tau_s$} {
      $s^{\mathrm{trial}}_i \leftarrow \mathrm{randu}[0.1,1]$
    }
    \lElse{
      $s^{\mathrm{trial}}_i \leftarrow s_i$
      }
    \lIf{$\mathrm{randu}[0,1] < \tau_c$} {
      $c^{\mathrm{trial}}_i \leftarrow \mathrm{randu}[0,1]$
    }
    \lElse{
      $c^{\mathrm{trial}}_i \leftarrow c_i$
      }  
  }
  \For{$i \in \{1, ..., \mu\}$}{
    $\vec{v}_{i} \leftarrow$ Apply differential mutation with $s^{\mathrm{trial}}_i$ to individuals in $\set{P}$\;
    $\vec{u}_{i} \leftarrow$ Apply binomial crossover with $c^{\mathrm{trial}}_i$ to $\vec{x}_{i}$ and $\vec{v}_{i}$\;
  }
  \For{$i \in \{1, \ldots, \mu\}$}{
    \If{$f(\vec{u}_{i}) \leq f(\vec{x}_{i})$} {
      $\set{A} \leftarrow \set{A} \cup \{\vec{x}_{i}\}$\;
      $\vec{x}_{i} \leftarrow \vec{u}_{i}$\;
      $s_i \leftarrow s^{\mathrm{trial}}_i$, $c_i \leftarrow c^{\mathrm{trial}}_i$\;      
    }
  }
  \lIf{$|\set{A}| > $ $a$} {
    Delete randomly selected individuals in $\set{A}$ unless $|\set{A}| < a$
  }
  $t \leftarrow t+1$\;
}
\caption{The basic DE algorithm with P-j}
\label{alg:de_pj}
\end{algorithm*}\DecMargin{0.5em}

\IncMargin{0.5em}
\begin{algorithm*}[t]
\small
\SetSideCommentRight
\SetKwRepeat{Do}{do}{while}%
$t \leftarrow 1$, initialize $\set{P} =\left\{ \vec{x}_1, \ldots, \vec{x}_{\mu}\right\}$ randomly, $\set{A} \leftarrow \emptyset$ \;
$m_s \leftarrow 0.5$, $m_c \leftarrow 0.5$\;
\While{The termination criteria are not met}{
  \For{$i \in \{1, ..., \mu\}$}{
  \Do{$s_i \leq 0$}{
    $s_i \leftarrow$ Randomly select a value from $C (m_s, 0.1)$\;
  }
  $s_i \leftarrow \min \{s_i, 1\}$\;
  $c_i \leftarrow$ Randomly select a value from $N (m_c, 0.1)$\;
  \lIf{$c_i \notin [0,1]$} {
    $c_i \leftarrow$ Replace with 0 or 1 closest to $c_i$
    }
  }
  \For{$i \in \{1, ..., \mu\}$}{
    $\vec{v}_{i} \leftarrow$ Apply differential mutation with $s_i$ to individuals in $\set{P}$\;
    $\vec{u}_{i} \leftarrow$ Apply binomial crossover with $c_i$ to $\vec{x}_{i}$ and $\vec{v}_{i}$\; 
  }
    $\Theta_s \leftarrow \emptyset$, $\Theta_c \leftarrow \emptyset$\;
  \For{$i \in \{1, \ldots, \mu\}$}{
    \If{$f(\vec{u}_{i}) \leq f(\vec{x}_{i})$} {
      $\set{A} \leftarrow \set{A} \cup \{\vec{x}_{i}\}$\;
      $\vec{x}_{i} \leftarrow \vec{u}_{i}$\;
      $\Theta_s \leftarrow \Theta_s \cup \{s_i\}$, $\Theta_c \leftarrow \Theta_c \cup \{c_i\}$\;
    }
  }
  \lIf{$|\set{A}| > $ $a$} {
    Delete randomly selected individuals in $\set{A}$ unless $|\set{A}| < a$
  }
  \If{$\Theta_s \neq \emptyset$ and $\Theta_c \neq \emptyset$} {
    $m_{s} \leftarrow (1 - \alpha) m_{s} + \alpha  \mathrm{Lmean}(\Theta_{s})$\;
    $m_{c} \leftarrow (1 - \alpha) m_{c} + \alpha \mathrm{mean}(\Theta_{c})$\;        
  }  
  $t \leftarrow t+1$\;
}
\caption{The basic DE algorithm with P-JA}
\label{alg:de_pja}
\end{algorithm*}\DecMargin{0.5em}

\IncMargin{0.5em}
\begin{algorithm*}[t]
\small
\SetSideCommentRight
\SetKwRepeat{Do}{do}{while}%
$t \leftarrow 1$, initialize $\set{P} =\left\{ \vec{x}_1, \ldots, \vec{x}_{\mu}\right\}$ randomly, $\set{A} \leftarrow \emptyset$ \;
$k \leftarrow 1$\;
\For{$i \in \{1, ..., h\}$}{
  $m_{s,i} \leftarrow 0.5$, $m_{c,i} \leftarrow 0.5$\;
}
\While{The termination criteria are not met}{
  \For{$i \in \{1, ..., \mu\}$}{
    $r \leftarrow$ Randomly select a value from $\{1, \ldots, h\}$\;
    \Do{$s_i \leq 0$}{
    $s_i \leftarrow$ Randomly select a value from $C (m_{s,r}, 0.1)$\;
  }
  $s_i \leftarrow \min \{s_i, 1\}$\;
  $c_i \leftarrow$ Randomly select a value from $N (m_{c,r}, 0.1)$\;
  \lIf{$c_i \notin [0,1]$} {
    $c_i \leftarrow$ Replace with 0 or 1 closest to $c_i$
    }
  }
  \For{$i \in \{1, ..., \mu\}$}{
    $\vec{v}_{i} \leftarrow$ Apply differential mutation with $s_i$ to individuals in $\set{P}$\;
    $\vec{u}_{i} \leftarrow$ Apply binomial crossover with $c_i$ to $\vec{x}_{i}$ and $\vec{v}_{i}$\; 
  }
    $\Theta_s \leftarrow \emptyset$, $\Theta_c \leftarrow \emptyset$\;
  \For{$i \in \{1, \ldots, \mu\}$}{
    \If{$f(\vec{u}_{i}) \leq f(\vec{x}_{i})$} {
      $\set{A} \leftarrow \set{A} \cup \{\vec{x}_{i}\}$\;
      $\vec{x}_{i} \leftarrow \vec{u}_{i}$\;
      $\Theta_s \leftarrow \Theta_s \cup \{s_i\}$, $\Theta_c \leftarrow \Theta_c \cup \{c_i\}$\;
    }
  }
  \lIf{$|\set{A}| > $ $a$} {
    Delete randomly selected individuals in $\set{A}$ unless $|\set{A}| < a$
  }
  \If{$\Theta_s \neq \emptyset$ and $\Theta_c \neq \emptyset$} {
    $m_{s,k} \leftarrow \mathrm{Lmean}(\Theta_{s})$\;
    $m_{c,k} \leftarrow \mathrm{Lmean}(\Theta_{c})$\;    
    $k \leftarrow k+1$\;
    \lIf{$k > h$} {
      $k \leftarrow 1$
    }
  }  
  $t \leftarrow t+1$\;
}
\caption{The basic DE algorithm with P-SHA}
\label{alg:de_pja}
\end{algorithm*}\DecMargin{0.5em}

\IncMargin{0.5em}
\begin{algorithm*}[t]
\small
\SetSideCommentRight
$t \leftarrow 1$, initialize $\set{P} =\left\{ \vec{x}_1, \ldots, \vec{x}_{\mu}\right\}$ randomly, $\set{A} \leftarrow \emptyset$ \;
\For{$i \in \{1, ..., \mu\}$}{
  $s_i \leftarrow$ Randomly select one from $\{0.4, 0.5, 0.6, 0.7, 0.8, 0.9\}$\;
  $c_i \leftarrow$ Randomly select one from $\{0.1, 0.2, 0.3, 0.4, 0.5, 0.6, 0.7, 0.8, 0.9\}$\;  
}
%
\While{The termination criteria are not met}{
  \For{$i \in \{1, ..., \mu\}$}{
    $\vec{v}_{i} \leftarrow$ Apply differential mutation with $s_i$ to individuals in $\set{P}$\;
    $\vec{u}_{i} \leftarrow$ Apply binomial crossover with $c_i$ to $\vec{x}_{i}$ and $\vec{v}_{i}$\;
  }
  \For{$i \in \{1, \ldots, \mu\}$}{
    \uIf{$f(\vec{u}_{i}) \leq f(\vec{x}_{i})$} {
      $\set{A} \leftarrow \set{A} \cup \{\vec{x}_{i}\}$\;
      $\vec{x}_{i} \leftarrow \vec{u}_{i}$\;
    }
    \Else{
      $s_i \leftarrow$ Randomly select one from $\{0.4, 0.5, 0.6, 0.7, 0.8, 0.9\}$\;
      $c_i \leftarrow$ Randomly select one from $\{0.1, 0.2, 0.3, 0.4, 0.5, 0.6, 0.7, 0.8, 0.9\}$\;
      }
  }
  \lIf{$|\set{A}| > $ $a$} {
    Delete randomly selected individuals in $\set{A}$ unless $|\set{A}| < a$
  }
  $t \leftarrow t+1$\;
}
\caption{The basic DE algorithm with P-EPS}
\label{alg:de_peps}
\end{algorithm*}\DecMargin{0.5em}

\IncMargin{0.5em}
\begin{algorithm*}[t]
\small
\SetSideCommentRight
$t \leftarrow 1$, initialize $\set{P} =\left\{ \vec{x}_1, \ldots, \vec{x}_{\mu}\right\}$ randomly, $\set{A} \leftarrow \emptyset$ \;
\For{$i \in \{1, ..., \mu\}$}{
  $\langle s_i, c_i\rangle \leftarrow$ Generate the scale factor and crossover rate by using \pref{alg:pcobi_gen}\;
}
%
\While{The termination criteria are not met}{
  \For{$i \in \{1, ..., \mu\}$}{
    $\vec{v}_{i} \leftarrow$ Apply differential mutation with $s_i$ to individuals in $\set{P}$\;
    $\vec{u}_{i} \leftarrow$ Apply binomial crossover with $c_i$ to $\vec{x}_{i}$ and $\vec{v}_{i}$\;
  }
  \For{$i \in \{1, \ldots, \mu\}$}{
    \uIf{$f(\vec{u}_{i}) \leq f(\vec{x}_{i})$} {
      $\set{A} \leftarrow \set{A} \cup \{\vec{x}_{i}\}$\;
      $\vec{x}_{i} \leftarrow \vec{u}_{i}$\;
    }
    \Else{
      $\langle s_i, c_i\rangle \leftarrow$ Generate the scale factor and crossover rate by using \pref{alg:pcobi_gen}\;      
      }
  }
  \lIf{$|\set{A}| > $ $a$} {
    Delete randomly selected individuals in $\set{A}$ unless $|\set{A}| < a$
  }
  $t \leftarrow t+1$\;
}
\caption{The basic DE algorithm with P-CoBi}
\label{alg:de_pcobi}
\end{algorithm*}\DecMargin{0.5em}

\IncMargin{0.5em}
\begin{algorithm*}[t]
\small
\SetSideCommentRight
\SetKw{Return}{return}
\uIf{$\mathrm{randu}[0,1] < 0.5$}{
  $s_i \leftarrow$ Randomly select a value from $C (0.65, 0.1)$\;
}
\Else{
  $s_i \leftarrow$ Randomly select a value from $C (1, 0.1)$\;
}
\uIf{$\mathrm{randu}[0,1] < 0.5$}{
  $c_i \leftarrow$ Randomly select a value from $C (0.1, 0.1)$\;
}
\Else{
  $c_i \leftarrow$ Randomly select a value from $C (0.95, 0.1)$\;
}
\Return $\langle s_i, c_i\rangle$\;
\caption{The parameter generation method in P-CoBi}
\label{alg:pcobi_gen}
\end{algorithm*}\DecMargin{0.5em}

\IncMargin{0.5em}
\begin{algorithm*}[t]
\small
\SetSideCommentRight
\SetKwRepeat{Do}{do}{while}%
$t \leftarrow 1$, initialize $\set{P} =\left\{ \vec{x}_1, \ldots, \vec{x}_{\mu}\right\}$ randomly, $\set{A} \leftarrow \emptyset$ \;
$\vec{q}_1 \leftarrow\langle0.5, 0\rangle, \vec{q}_2 \leftarrow\langle0.5, 0.5\rangle,  \vec{q}_3 \leftarrow\langle0.5, 1\rangle, \vec{q}_4 \leftarrow\langle0.8, 0\rangle, \vec{q}_5 \leftarrow\langle0.8, 0.5\rangle,  \vec{q}_6 \leftarrow\langle0.8, 1\rangle, \vec{q}_7 \leftarrow\langle1, 0\rangle, \vec{q}_8 \leftarrow\langle1, 0.5\rangle,  \vec{q}_9 \leftarrow\langle1, 1\rangle$\;
\lFor{$i \in \{1, ..., 9\}$}{
  $o_i \leftarrow 0$
}
\While{The termination criteria are not met}{
  \For{$i \in \{1, ..., \mu\}$}{   
    \For{$j \in \{1, ..., 9\}$}{
      $\tau_{j} \leftarrow \frac{o_{j} + \epsilon}{\sum^9_{k=1} (o_{k} + \epsilon)}$
    }       
    \For{$j \in \{1, ..., 9\}$}{
      \If{$\tau_{j} \leq \delta$} {
        $o_{j} \leftarrow 0$\;
        $\tau_{j} \leftarrow \frac{o_{j} + \epsilon}{\sum^9_{k=1} (o_{k} + \epsilon)}$\;
      }
    }
  }
  \For{$i \in \{1, ..., \mu\}$}{
    $\langle s_i, c_i \rangle \leftarrow $ Randomly select one from $\vec{q}_1, \ldots, \vec{q}_9$ with the propabities $\tau_{1}, \ldots, \tau_{9}$\;    
  }
  \For{$i \in \{1, ..., \mu\}$}{
    $\vec{v}_{i} \leftarrow$ Apply differential mutation with $s_i$ to individuals in $\set{P}$\;
    $\vec{u}_{i} \leftarrow$ Apply binomial crossover with $c_i$ to $\vec{x}_{i}$ and $\vec{v}_{i}$\; 
  }
  \For{$i \in \{1, \ldots, \mu\}$}{
    \If{$f(\vec{u}_{i}) \leq f(\vec{x}_{i})$} {
      $\set{A} \leftarrow \set{A} \cup \{\vec{x}_{i}\}$\;
      $\vec{x}_{i} \leftarrow \vec{u}_{i}$\;
      \If{$\vec{q}_j = \langle s_i, c_i \rangle$} {
        $o_{j} \leftarrow o_{j} + 1$\;
      }             
    }
  }
  \lIf{$|\set{A}| > $ $a$} {
    Delete randomly selected individuals in $\set{A}$ unless $|\set{A}| < a$
  }
  $t \leftarrow t+1$\;
}
\caption{The basic DE algorithm with P-c}
\label{alg:de_pc}
\end{algorithm*}\DecMargin{0.5em}

\clearpage

\begin{figure*}[htbp]
\newcommand{\width}{0.31}
\centering
\subfloat[$n=5$]{
\includegraphics[width=\width\textwidth]{./figs/vs_de/rand_1_mu100_conventional_Baldwin/NOPCM_P-j_P-JA_P-SHA_P-EPS_P-CoB_P-c_et_al/pprldmany_10D_noiselessall.pdf}
}
\subfloat[$n=10$]{
\includegraphics[width=\width\textwidth]{./figs/vs_de/rand_1_mu100_conventional_Baldwin/NOPCM_P-j_P-JA_P-SHA_P-EPS_P-CoB_P-c_et_al/pprldmany_10D_noiselessall.pdf}
}
\subfloat[$n=20$]{
\includegraphics[width=\width\textwidth]{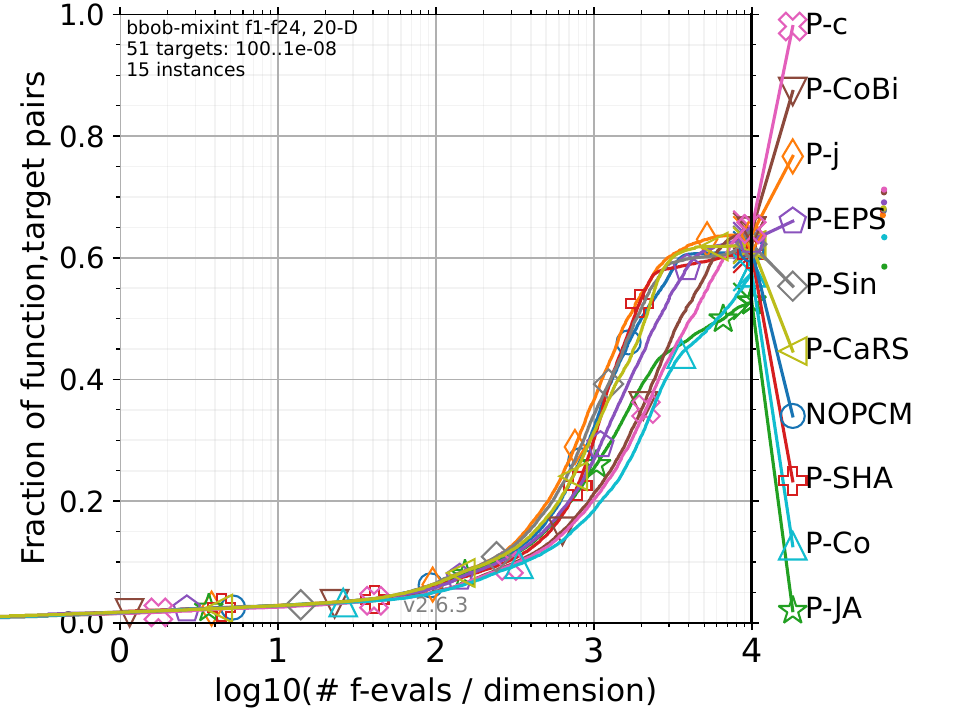}
}
\\
\subfloat[$n=40$]{
\includegraphics[width=\width\textwidth]{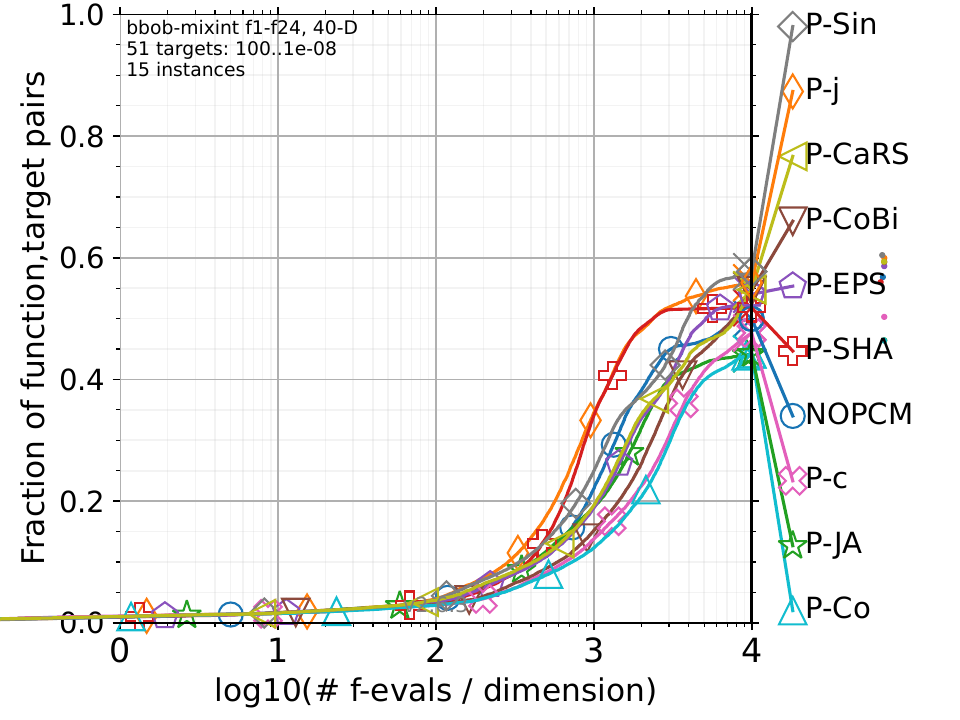}
}
\subfloat[$n=80$]{
\includegraphics[width=\width\textwidth]{./figs/vs_de/rand_1_mu100_conventional_Baldwin/NOPCM_P-j_P-JA_P-SHA_P-EPS_P-CoB_P-c_et_al/pprldmany_80D_noiselessall.pdf}
}
\subfloat[$n=160$]{
\includegraphics[width=\width\textwidth]{./figs/vs_de/rand_1_mu100_conventional_Baldwin/NOPCM_P-j_P-JA_P-SHA_P-EPS_P-CoB_P-c_et_al/pprldmany_160D_noiselessall.pdf}
}
\caption{Comparison of the nine PCMs and NOPCM  with the rand/1 mutation strategy and the Baldwinian repair method on the 24 \texttt{bbob-mixint} functions for $n \in \{5, 10, 20, 40, 80, 160\}$.}
\label{supfig:vs_de_rand_1_Baldwin}
\subfloat[$n=5$]{
\includegraphics[width=\width\textwidth]{./figs/vs_de/rand_1_mu100_conventional_Lamarckian/NOPCM_P-j_P-JA_P-SHA_P-EPS_P-CoB_P-c_et_al/pprldmany_10D_noiselessall.pdf}
}
\subfloat[$n=10$]{
\includegraphics[width=\width\textwidth]{./figs/vs_de/rand_1_mu100_conventional_Lamarckian/NOPCM_P-j_P-JA_P-SHA_P-EPS_P-CoB_P-c_et_al/pprldmany_10D_noiselessall.pdf}
}
\subfloat[$n=20$]{
\includegraphics[width=\width\textwidth]{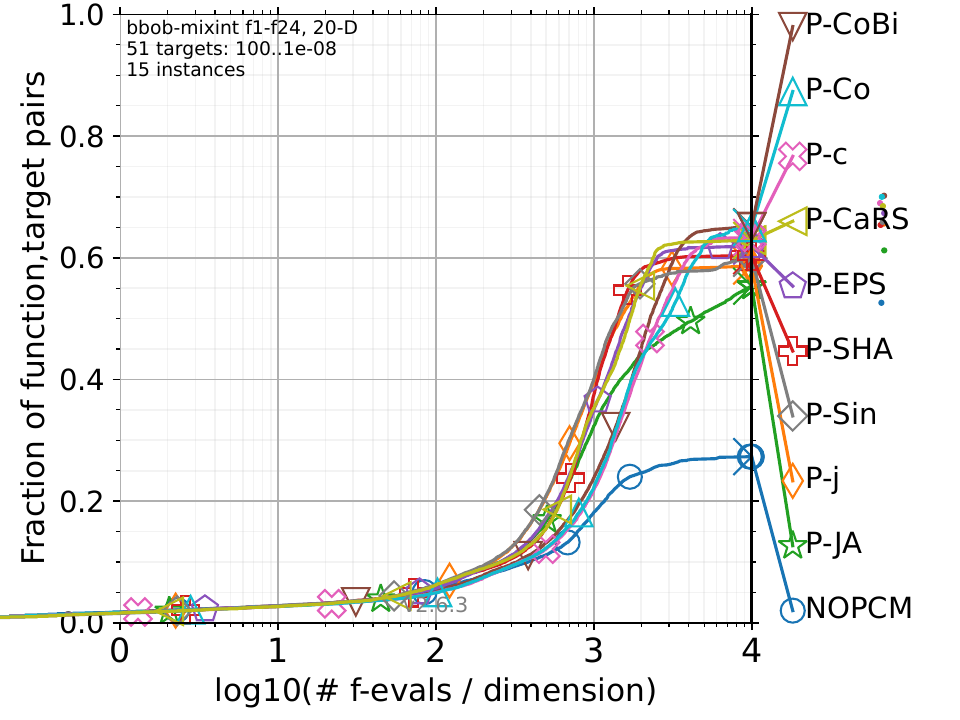}
}
\\
\subfloat[$n=40$]{
\includegraphics[width=\width\textwidth]{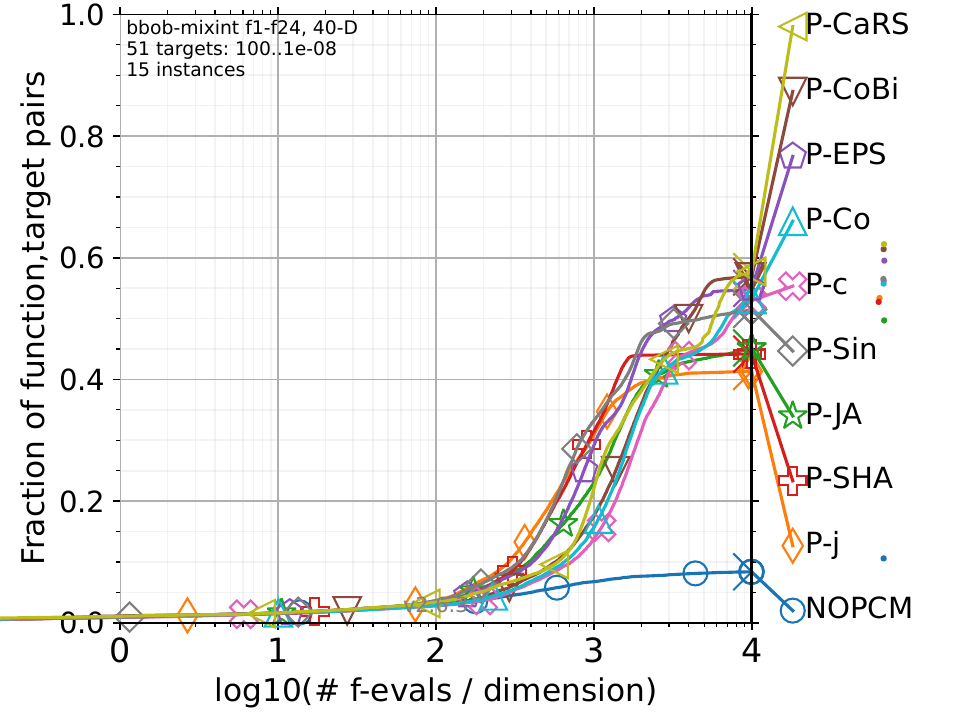}
}
\subfloat[$n=80$]{
\includegraphics[width=\width\textwidth]{./figs/vs_de/rand_1_mu100_conventional_Lamarckian/NOPCM_P-j_P-JA_P-SHA_P-EPS_P-CoB_P-c_et_al/pprldmany_80D_noiselessall.pdf}
}
\subfloat[$n=160$]{
\includegraphics[width=\width\textwidth]{./figs/vs_de/rand_1_mu100_conventional_Lamarckian/NOPCM_P-j_P-JA_P-SHA_P-EPS_P-CoB_P-c_et_al/pprldmany_160D_noiselessall.pdf}
}
\caption{Comparison of the nine PCMs and NOPCM  with the rand/1 mutation strategy and the Lamarckian repair method on the 24 \texttt{bbob-mixint} functions for $n \in \{5, 10, 20, 40, 80, 160\}$.}
\label{supfig:vs_de_rand_1_Lamarckian}
\end{figure*}

\begin{figure*}[htbp]
\newcommand{\width}{0.31}
\centering
\subfloat[$n=5$]{
\includegraphics[width=\width\textwidth]{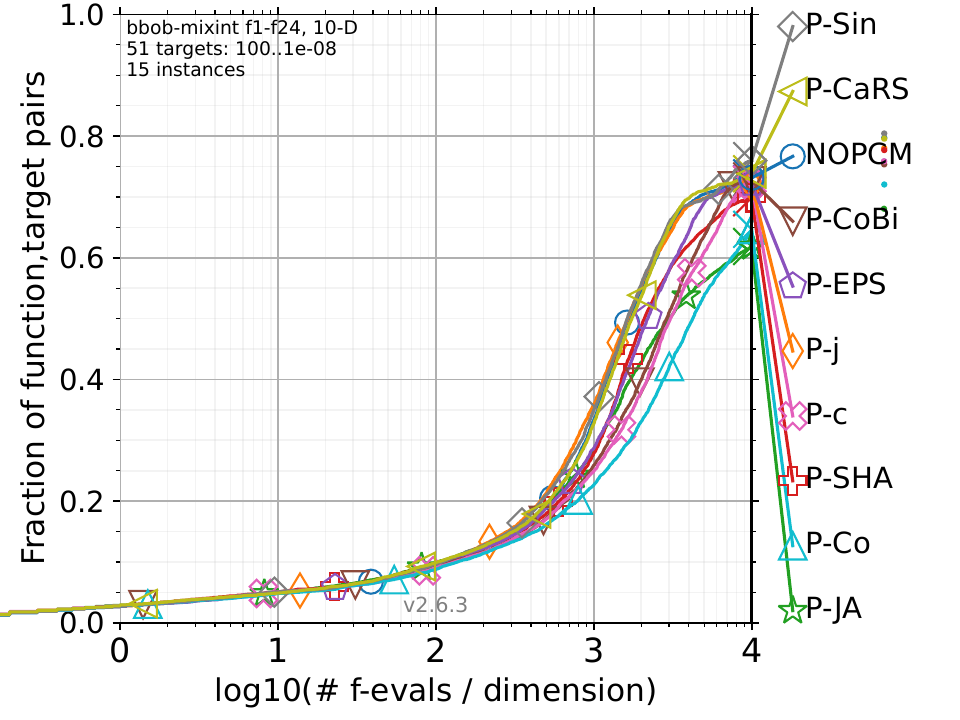}
}
\subfloat[$n=10$]{
\includegraphics[width=\width\textwidth]{./figs/vs_de/rand_2_mu100_conventional_Baldwin/NOPCM_P-j_P-JA_P-SHA_P-EPS_P-CoB_P-c_et_al/pprldmany_10D_noiselessall.pdf}
}
\subfloat[$n=20$]{
\includegraphics[width=\width\textwidth]{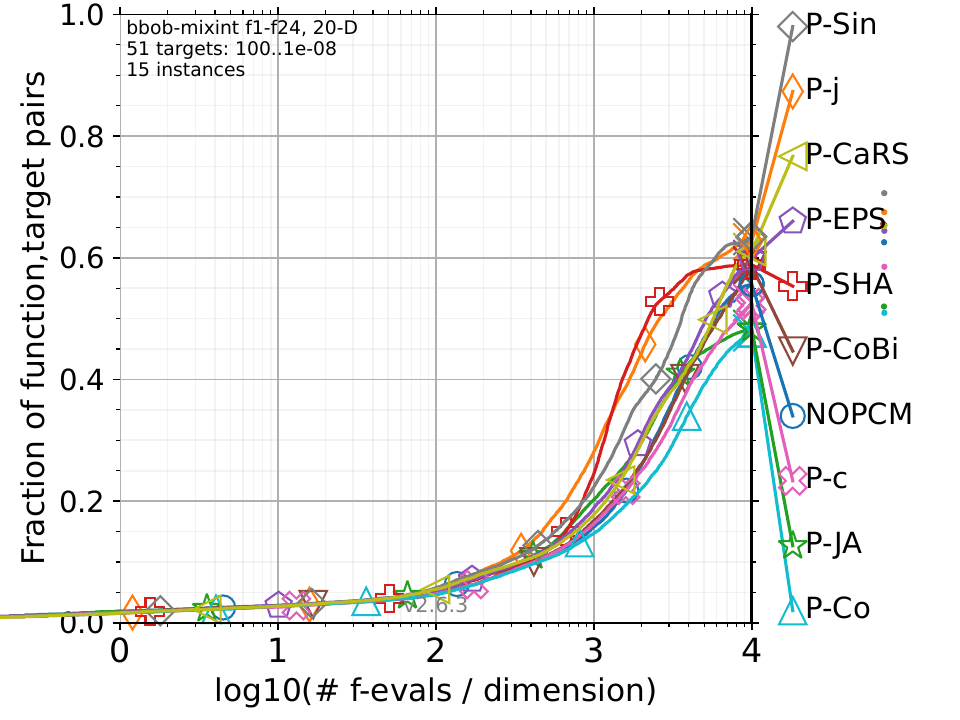}
}
\\
\subfloat[$n=40$]{
\includegraphics[width=\width\textwidth]{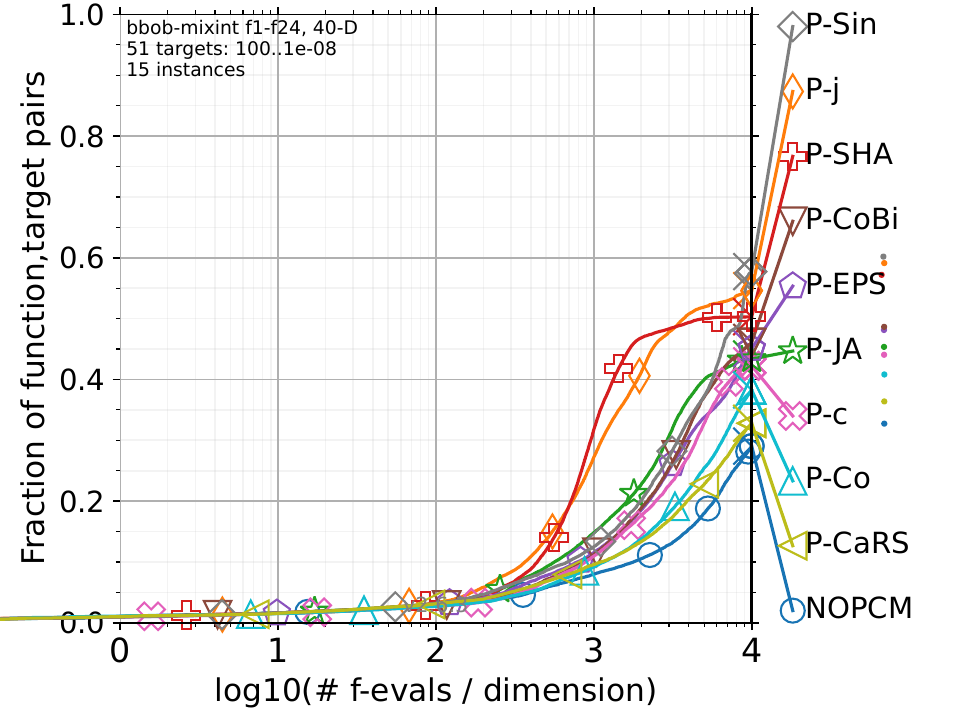}
}
\subfloat[$n=80$]{
\includegraphics[width=\width\textwidth]{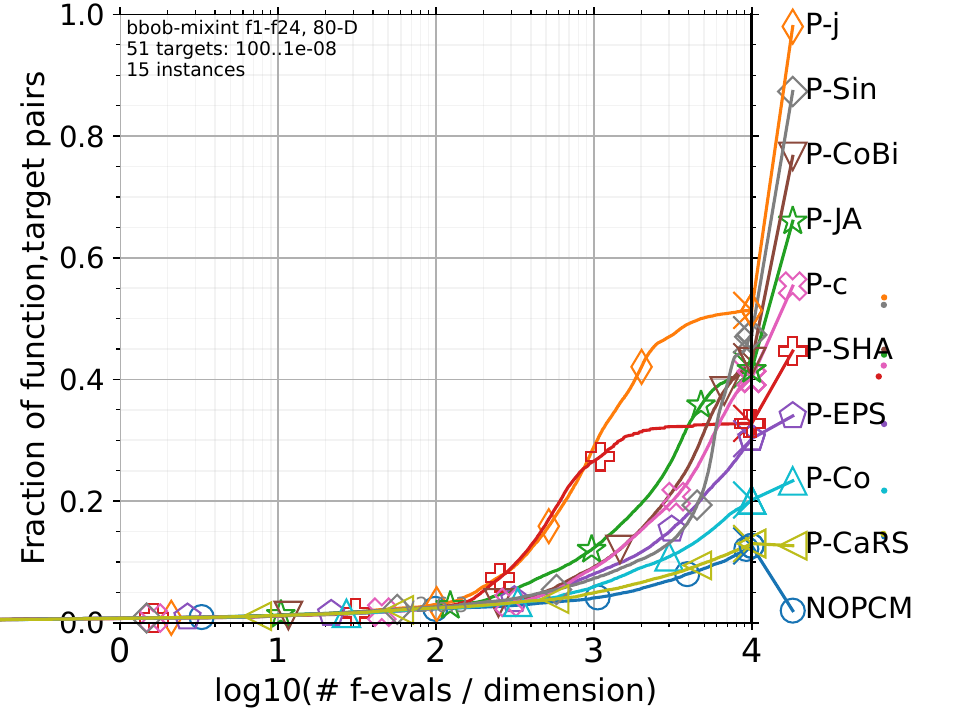}
}
\subfloat[$n=160$]{
\includegraphics[width=\width\textwidth]{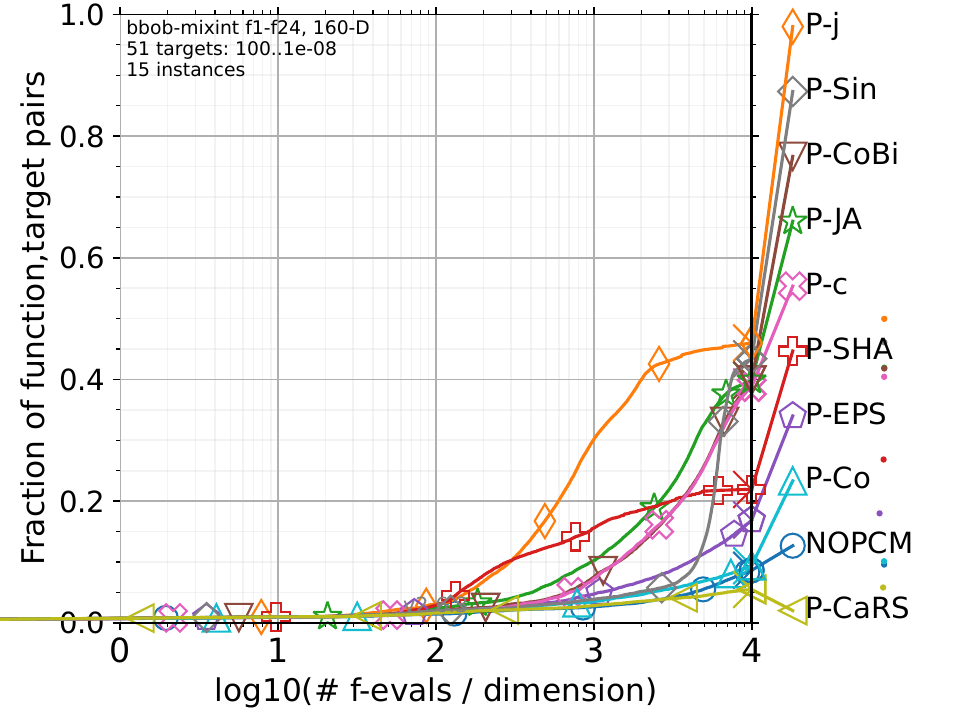}
}
\caption{Comparison of the nine PCMs and NOPCM  with the rand/2 mutation strategy and the Baldwinian repair method on the 24 \texttt{bbob-mixint} functions for $n \in \{5, 10, 20, 40, 80, 160\}$.}
\label{supfig:vs_de_rand_2_Baldwin}
\subfloat[$n=5$]{
\includegraphics[width=\width\textwidth]{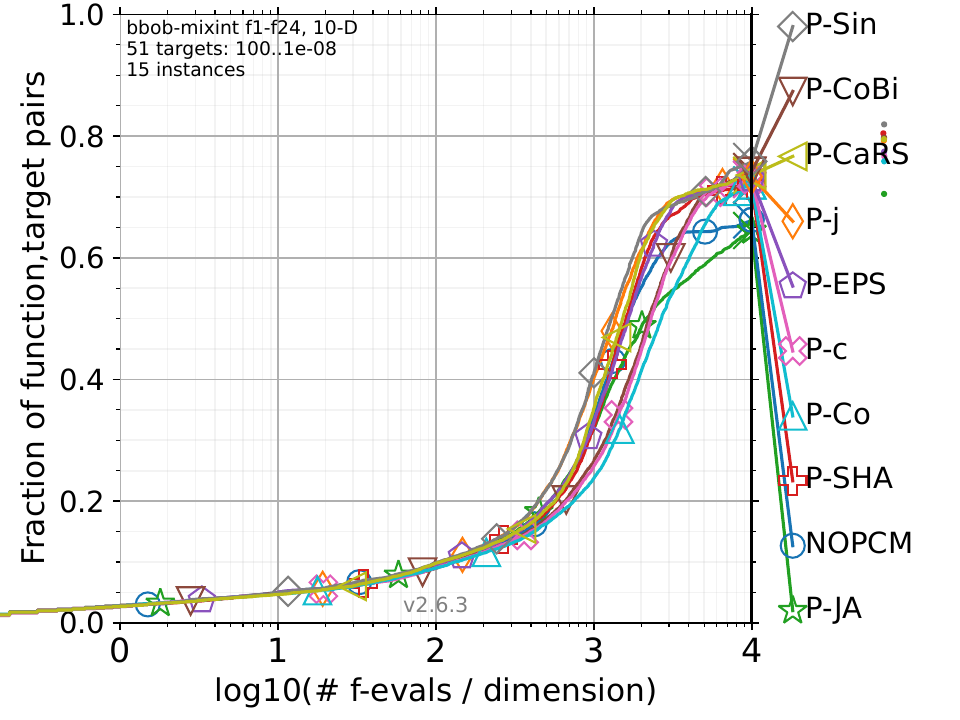}
}
\subfloat[$n=10$]{
\includegraphics[width=\width\textwidth]{./figs/vs_de/rand_2_mu100_conventional_Lamarckian/NOPCM_P-j_P-JA_P-SHA_P-EPS_P-CoB_P-c_et_al/pprldmany_10D_noiselessall.pdf}
}
\subfloat[$n=20$]{
\includegraphics[width=\width\textwidth]{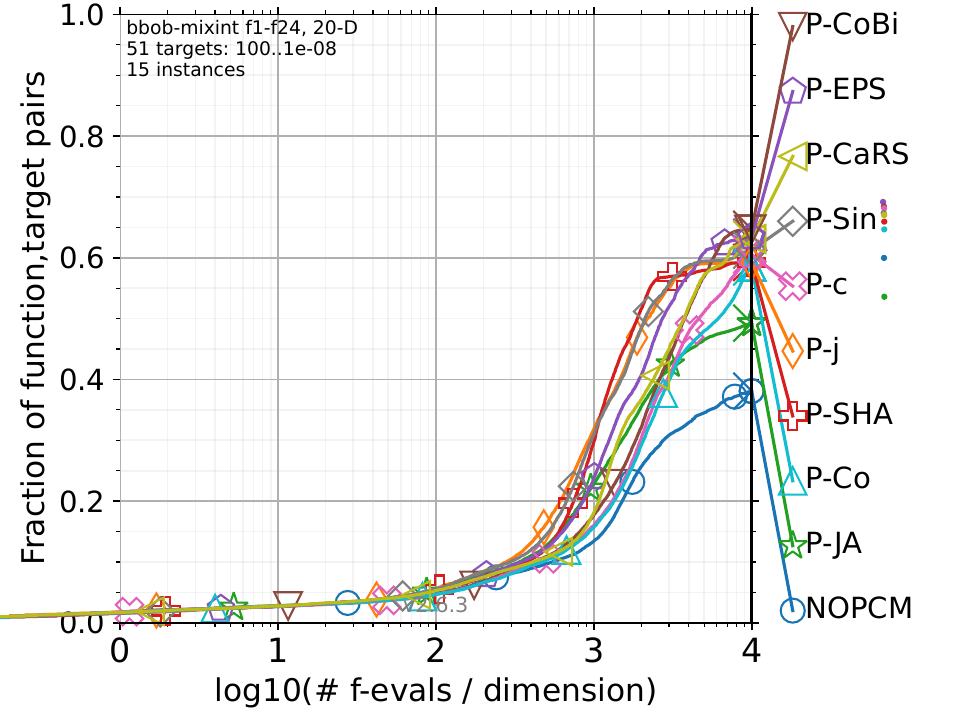}
}
\\
\subfloat[$n=40$]{
\includegraphics[width=\width\textwidth]{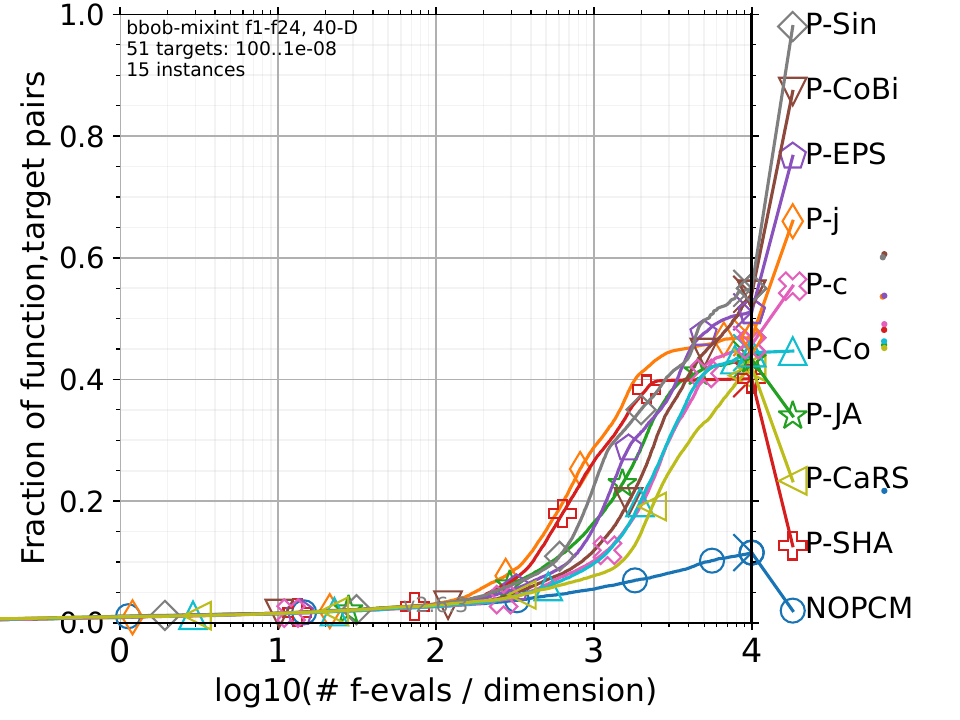}
}
\subfloat[$n=80$]{
\includegraphics[width=\width\textwidth]{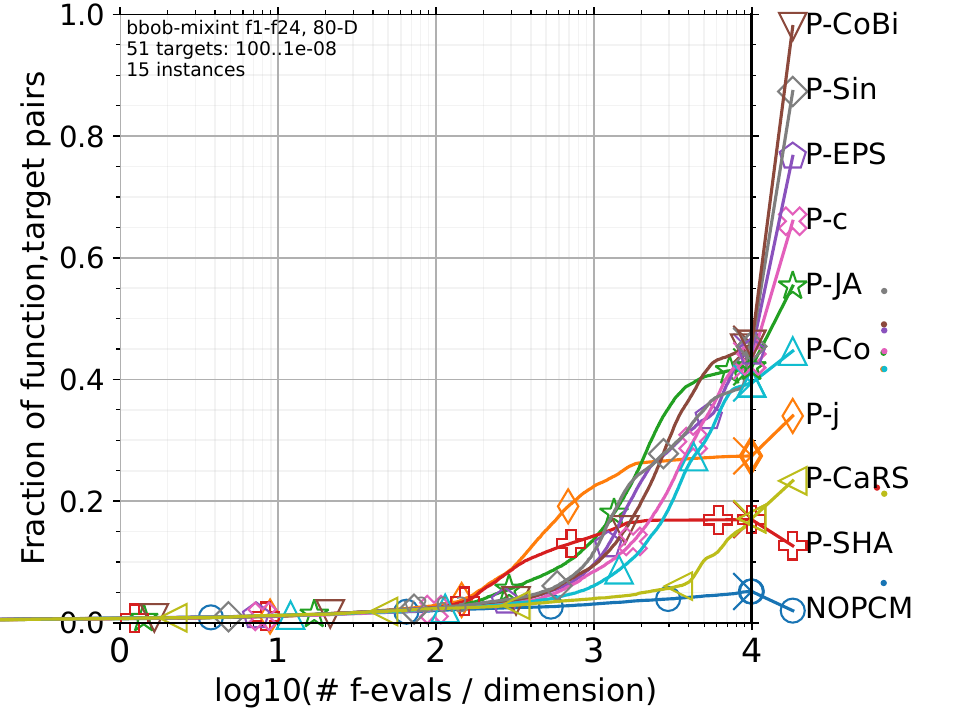}
}
\subfloat[$n=160$]{
\includegraphics[width=\width\textwidth]{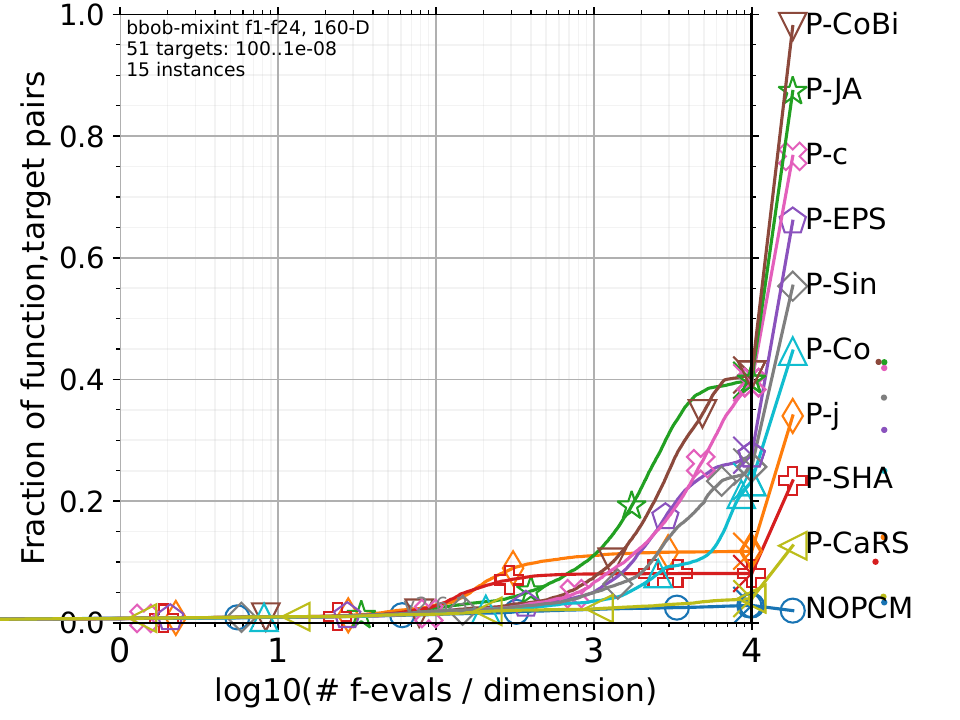}
}
\caption{Comparison of the nine PCMs and NOPCM  with the rand/2 mutation strategy and the Lamarckian repair method on the 24 \texttt{bbob-mixint} functions for $n \in \{5, 10, 20, 40, 80, 160\}$.}
\label{supfig:vs_de_rand_2_Lamarckian}
\end{figure*}

\begin{figure*}[htbp]
\newcommand{\width}{0.31}
\centering
\subfloat[$n=5$]{
\includegraphics[width=\width\textwidth]{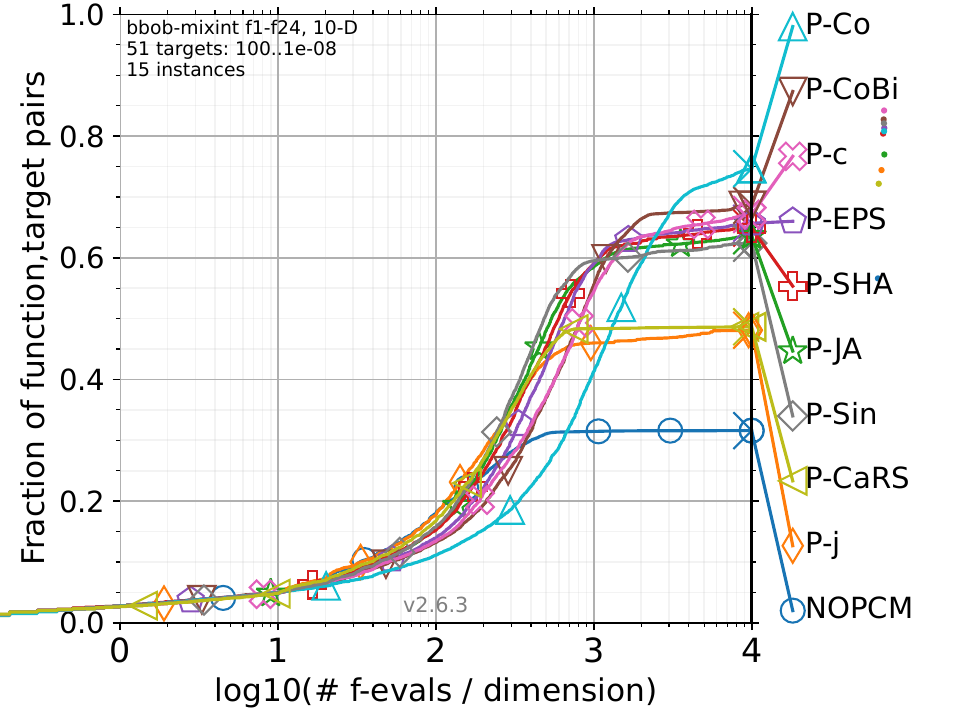}
}
\subfloat[$n=10$]{
\includegraphics[width=\width\textwidth]{./figs/vs_de/best_1_mu100_conventional_Baldwin/NOPCM_P-j_P-JA_P-SHA_P-EPS_P-CoB_P-c_et_al/pprldmany_10D_noiselessall.pdf}
}
\subfloat[$n=20$]{
\includegraphics[width=\width\textwidth]{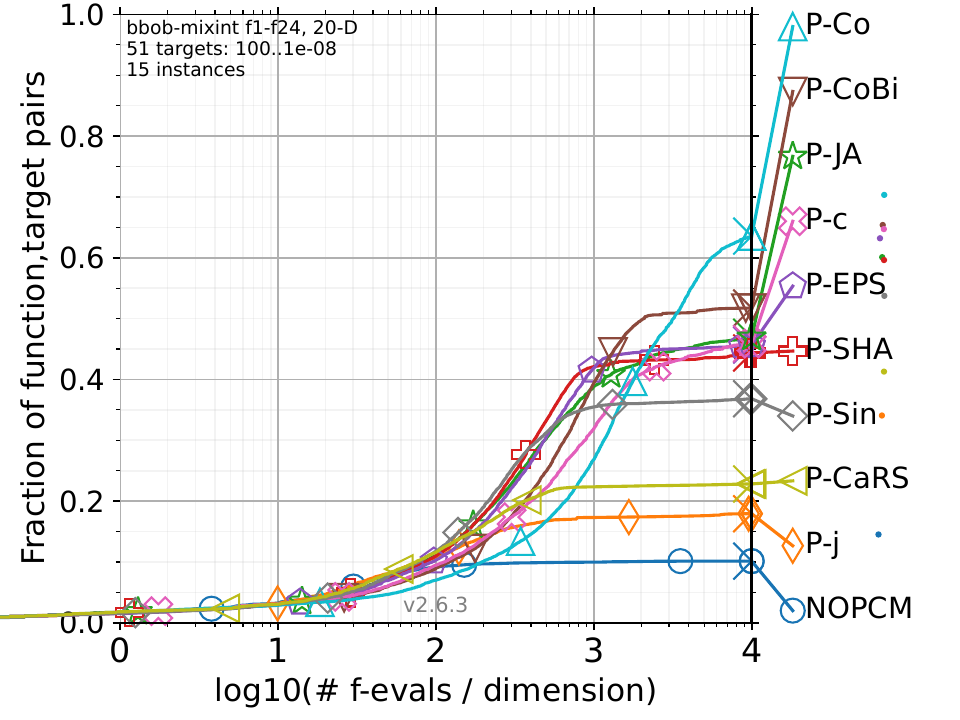}
}
\\
\subfloat[$n=40$]{
\includegraphics[width=\width\textwidth]{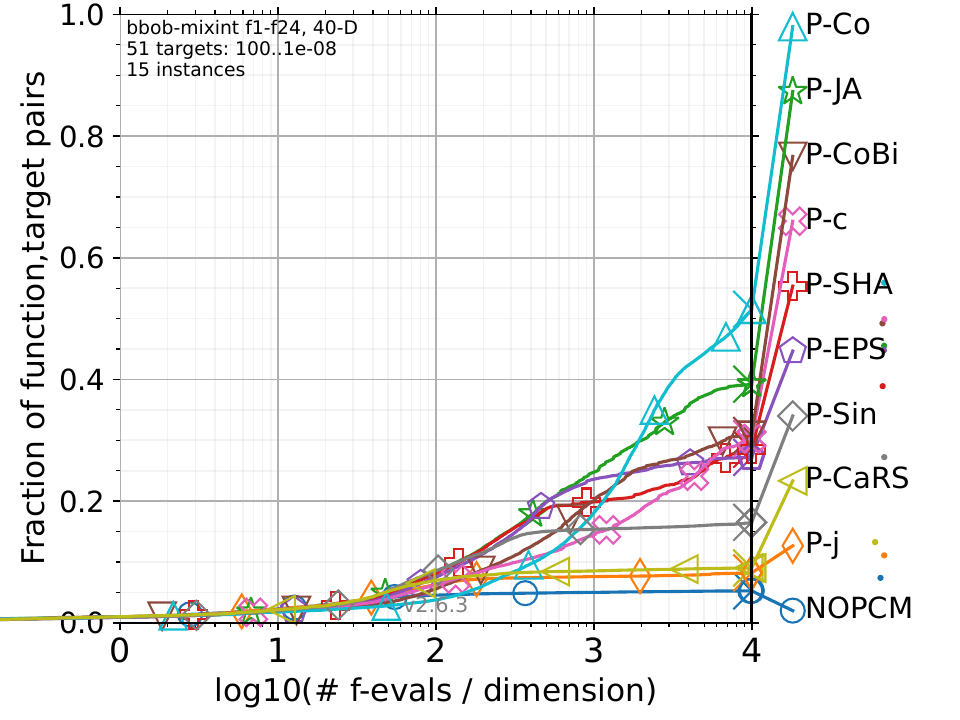}
}
\subfloat[$n=80$]{
\includegraphics[width=\width\textwidth]{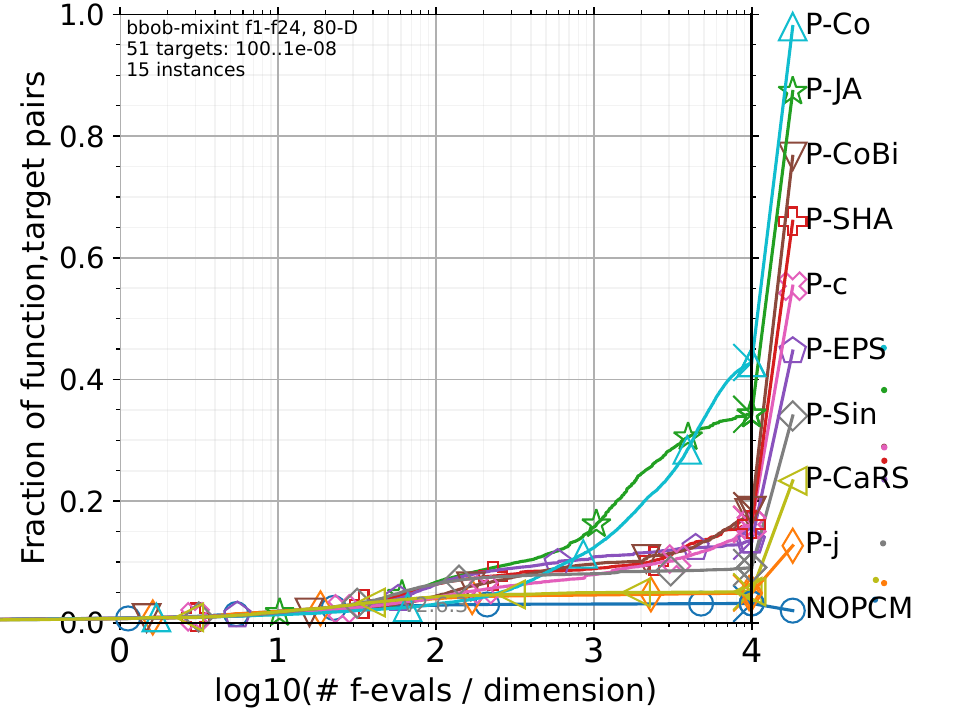}
}
\subfloat[$n=160$]{
\includegraphics[width=\width\textwidth]{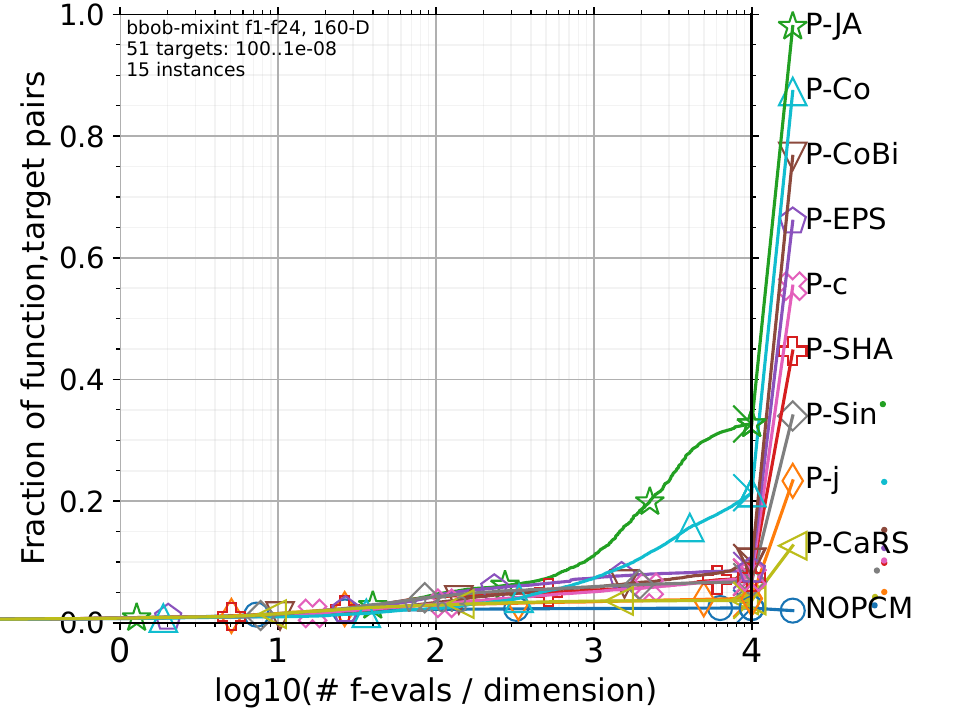}
}
\caption{Comparison of the nine PCMs and NOPCM  with the best/1 mutation strategy and the Baldwinian repair method on the 24 \texttt{bbob-mixint} functions for $n \in \{5, 10, 20, 40, 80, 160\}$.}
\label{supfig:vs_de_best_1_Baldwin}
\subfloat[$n=5$]{
\includegraphics[width=\width\textwidth]{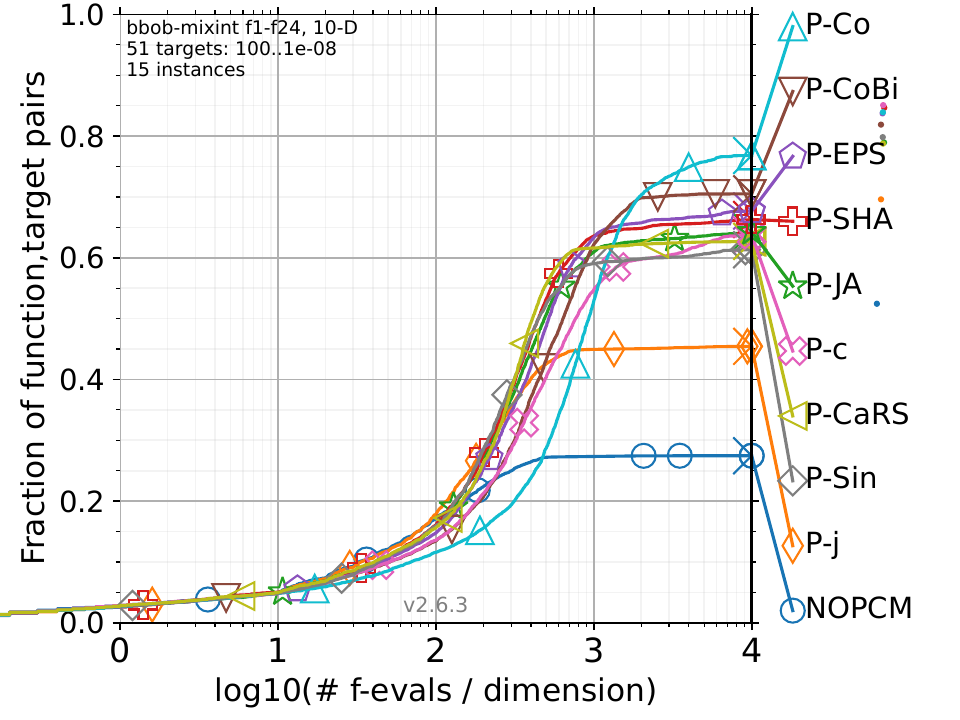}
}
\subfloat[$n=10$]{
\includegraphics[width=\width\textwidth]{./figs/vs_de/best_1_mu100_conventional_Lamarckian/NOPCM_P-j_P-JA_P-SHA_P-EPS_P-CoB_P-c_et_al/pprldmany_10D_noiselessall.pdf}
}
\subfloat[$n=20$]{
\includegraphics[width=\width\textwidth]{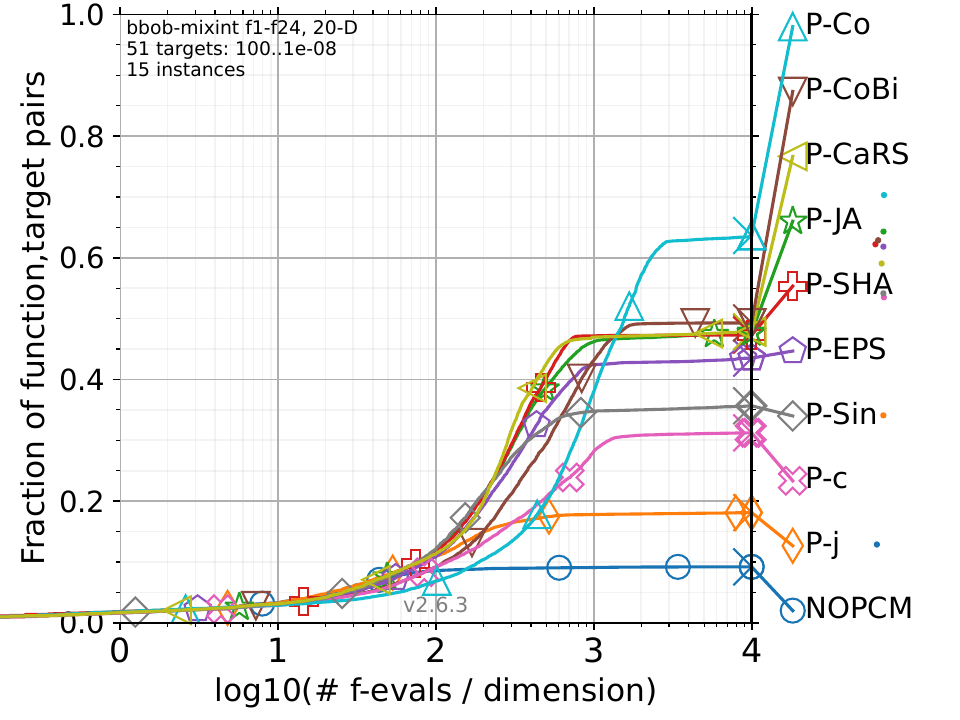}
}
\\
\subfloat[$n=40$]{
\includegraphics[width=\width\textwidth]{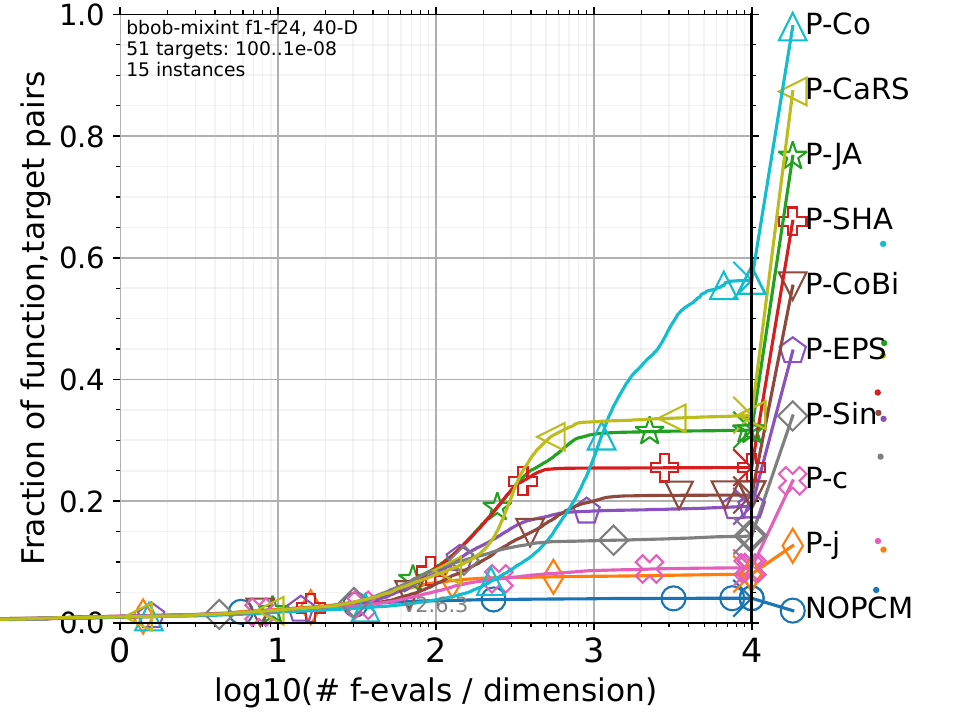}
}
\subfloat[$n=80$]{
\includegraphics[width=\width\textwidth]{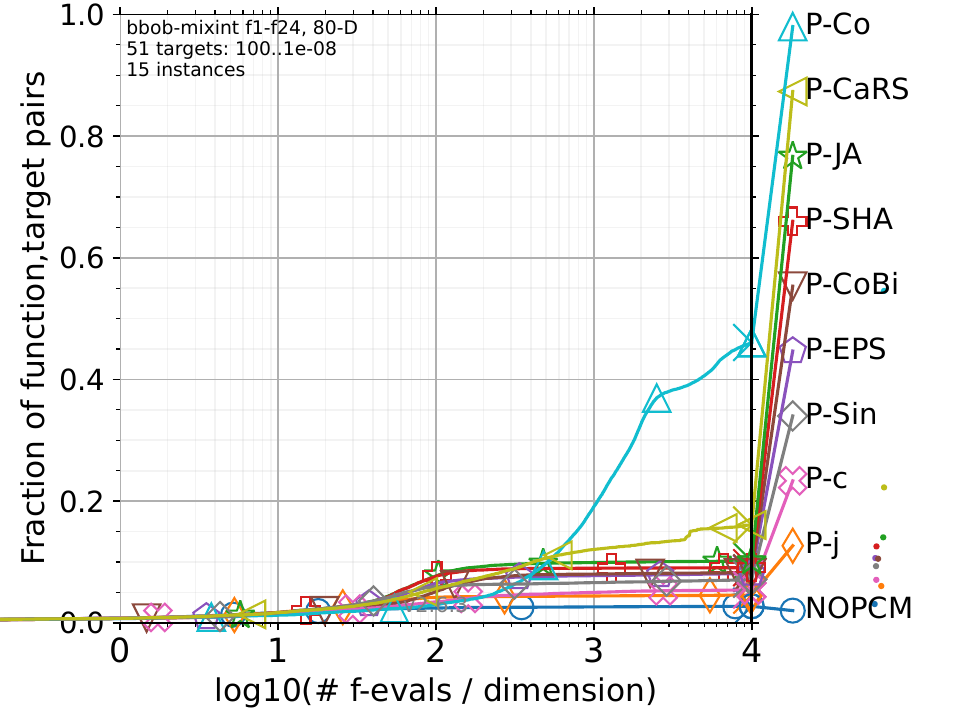}
}
\subfloat[$n=160$]{
\includegraphics[width=\width\textwidth]{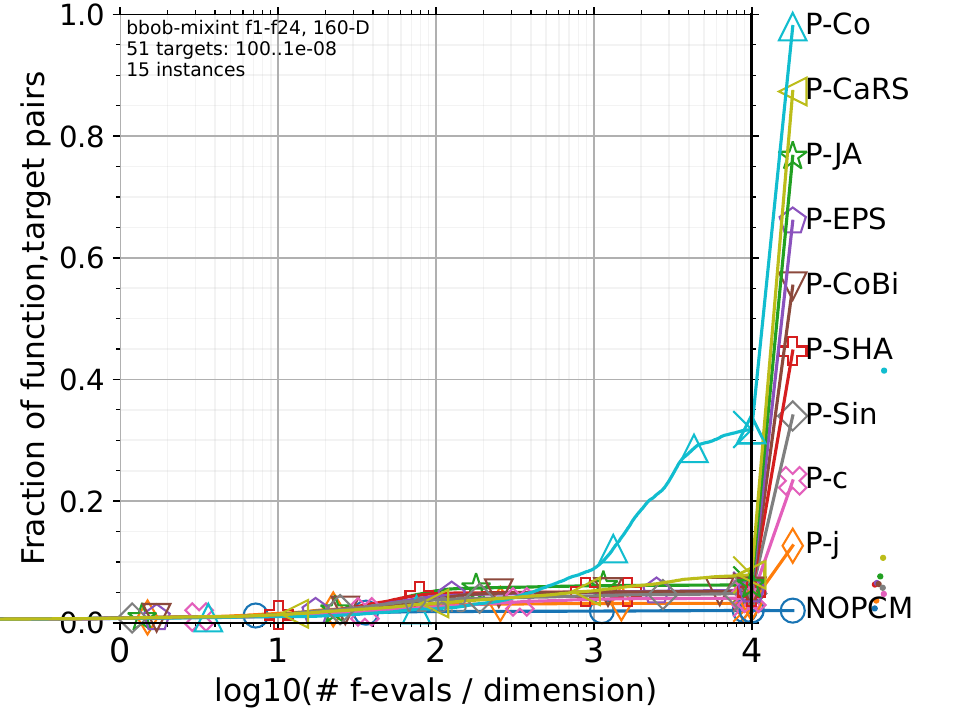}
}
\caption{Comparison of the nine PCMs and NOPCM  with the best/1 mutation strategy and the Lamarckian repair method on the 24 \texttt{bbob-mixint} functions for $n \in \{5, 10, 20, 40, 80, 160\}$.}
\label{supfig:vs_de_best_1_Lamarckian}
\end{figure*}

\begin{figure*}[htbp]
\newcommand{\width}{0.31}
\centering
\subfloat[$n=5$]{
\includegraphics[width=\width\textwidth]{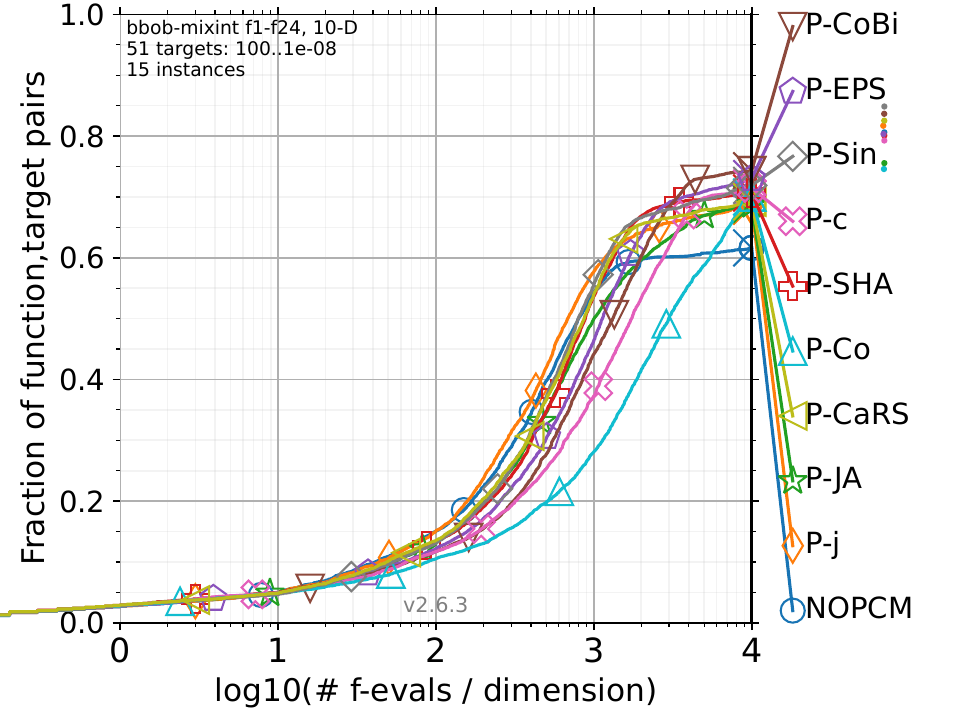}
}
\subfloat[$n=10$]{
\includegraphics[width=\width\textwidth]{./figs/vs_de/best_2_mu100_conventional_Baldwin/NOPCM_P-j_P-JA_P-SHA_P-EPS_P-CoB_P-c_et_al/pprldmany_10D_noiselessall.pdf}
}
\subfloat[$n=20$]{
\includegraphics[width=\width\textwidth]{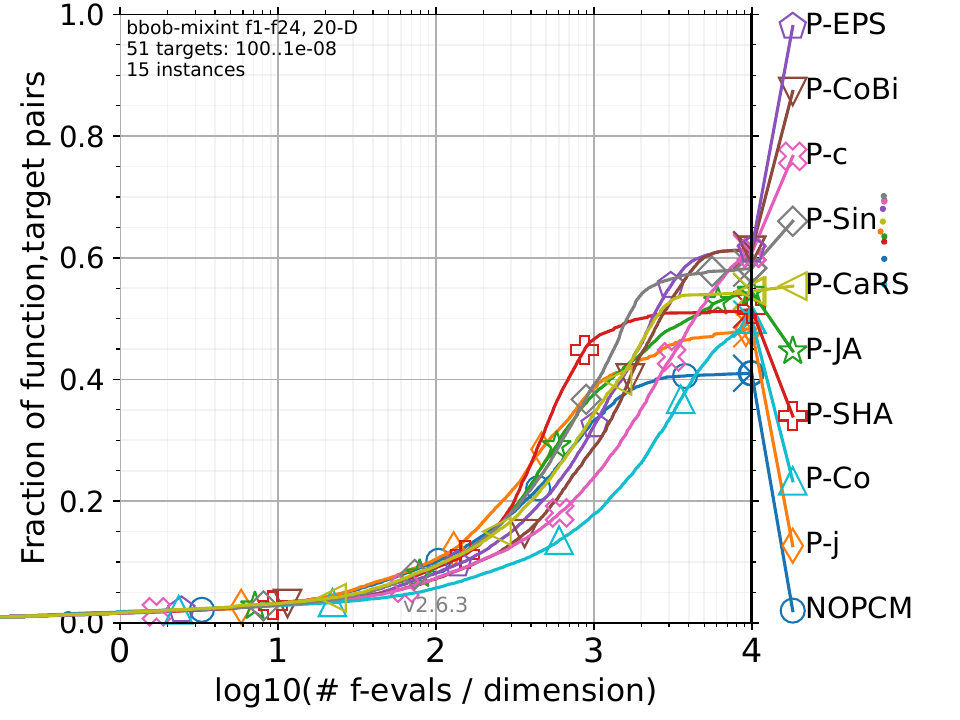}
}
\\
\subfloat[$n=40$]{
\includegraphics[width=\width\textwidth]{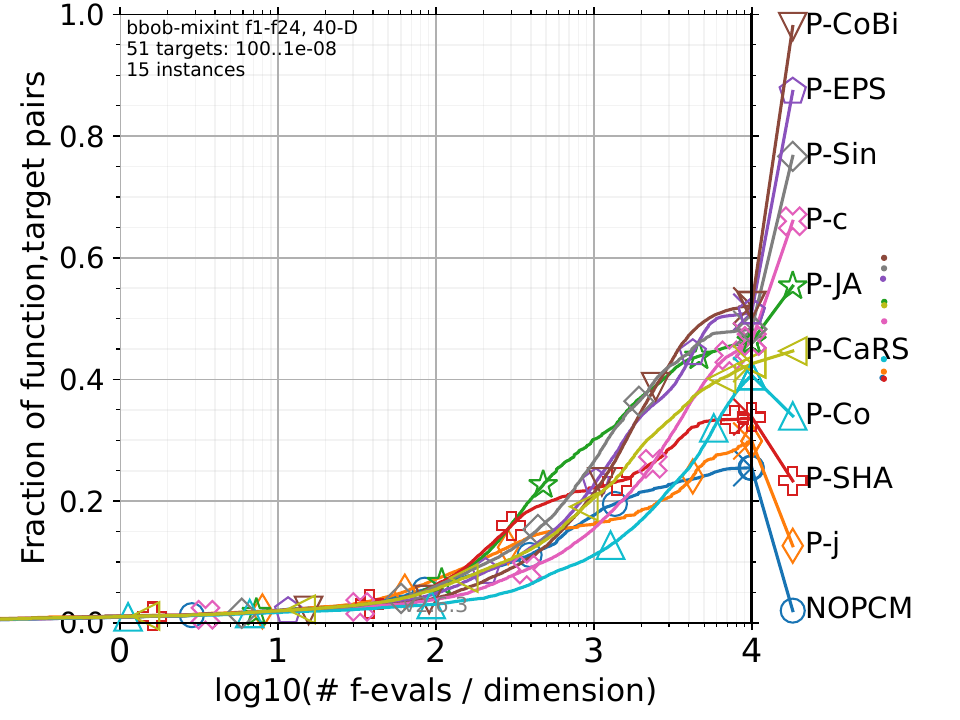}
}
\subfloat[$n=80$]{
\includegraphics[width=\width\textwidth]{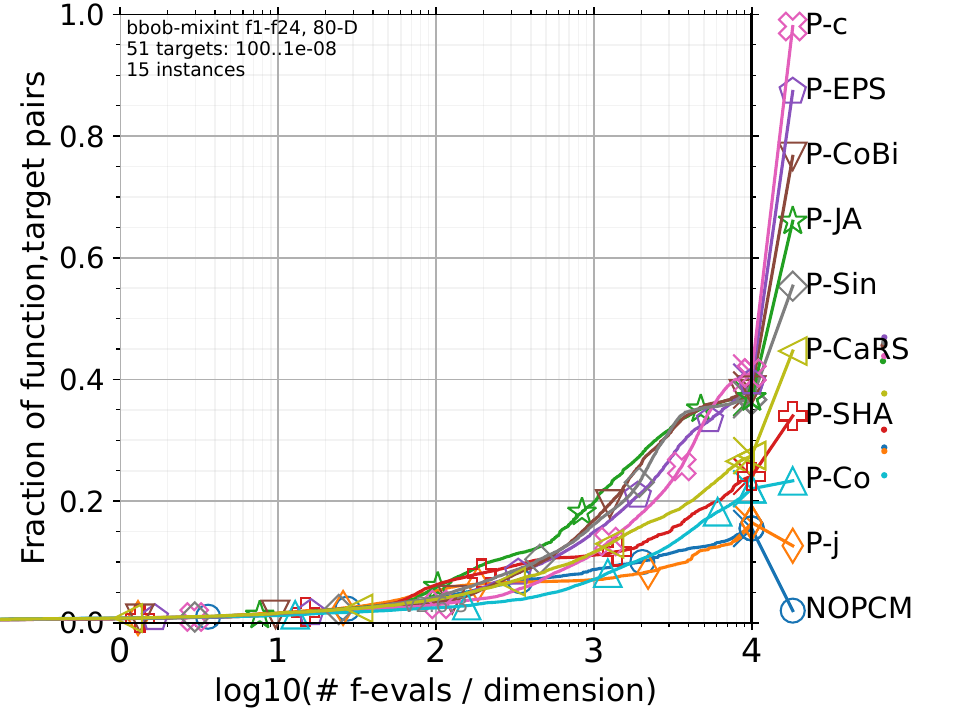}
}
\subfloat[$n=160$]{
\includegraphics[width=\width\textwidth]{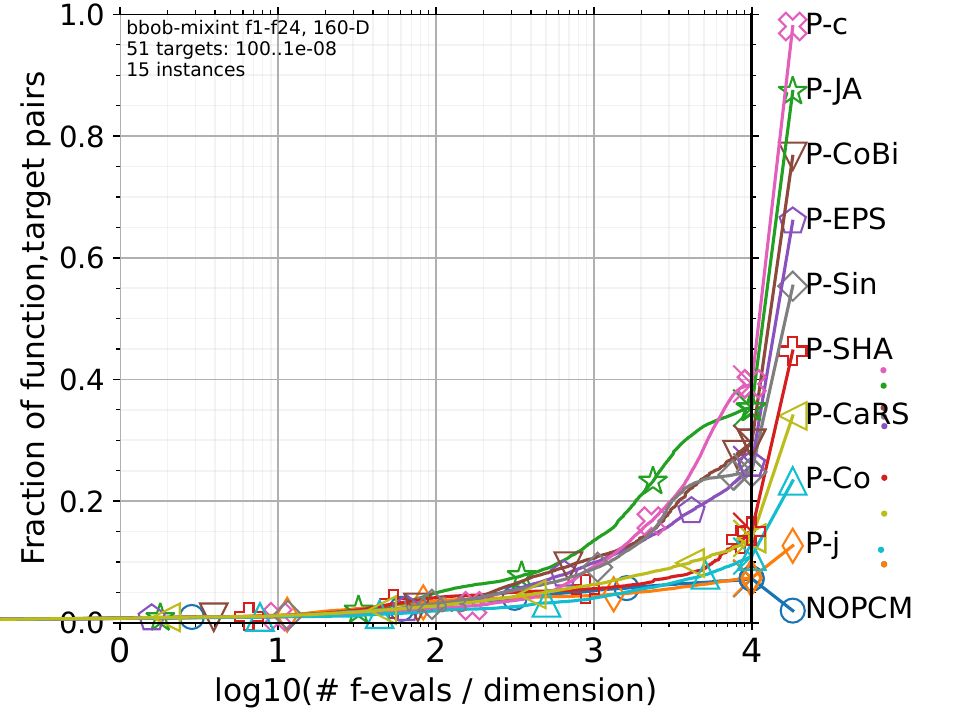}
}
\caption{Comparison of the nine PCMs and NOPCM  with the best/2 mutation strategy and the Baldwinian repair method on the 24 \texttt{bbob-mixint} functions for $n \in \{5, 10, 20, 40, 80, 160\}$.}
\label{supfig:vs_de_best_2_Baldwin}
\subfloat[$n=5$]{
\includegraphics[width=\width\textwidth]{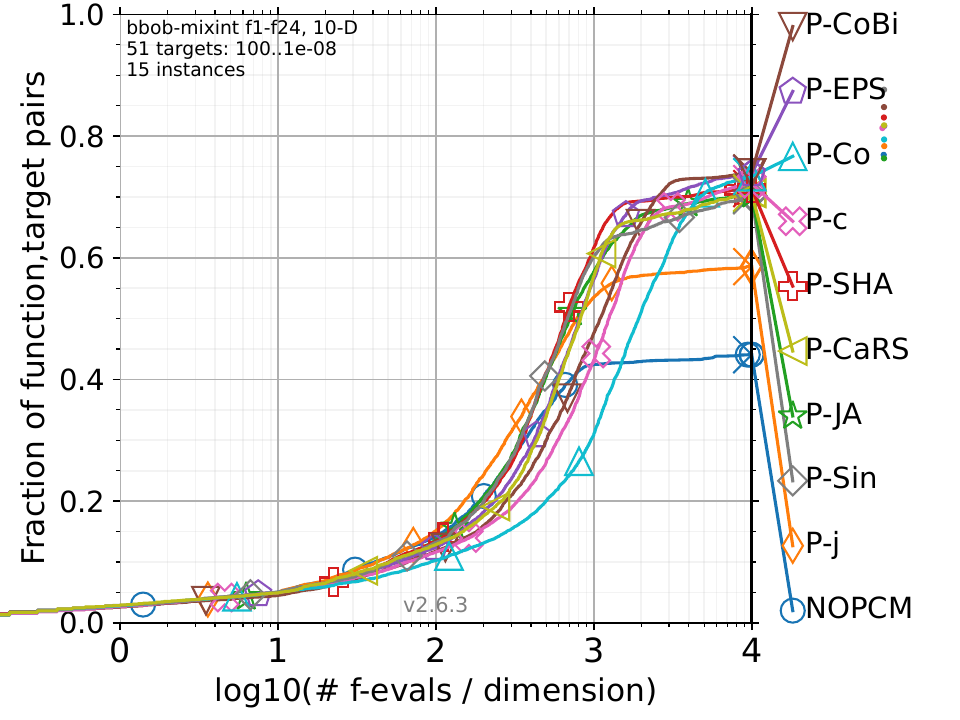}
}
\subfloat[$n=10$]{
\includegraphics[width=\width\textwidth]{./figs/vs_de/best_2_mu100_conventional_Lamarckian/NOPCM_P-j_P-JA_P-SHA_P-EPS_P-CoB_P-c_et_al/pprldmany_10D_noiselessall.pdf}
}
\subfloat[$n=20$]{
\includegraphics[width=\width\textwidth]{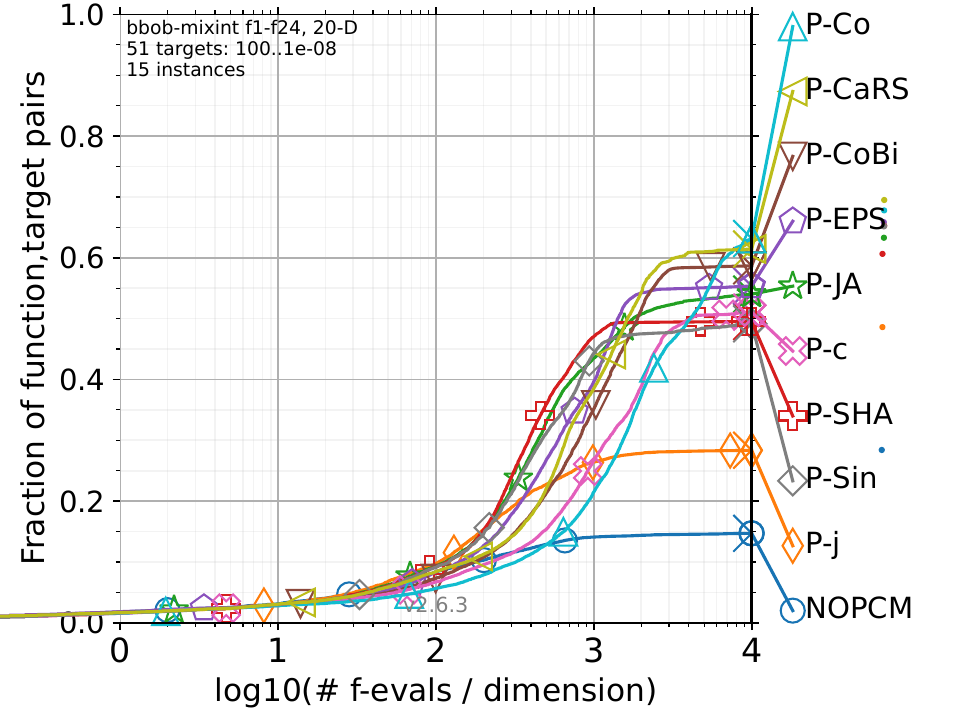}
}
\\
\subfloat[$n=40$]{
\includegraphics[width=\width\textwidth]{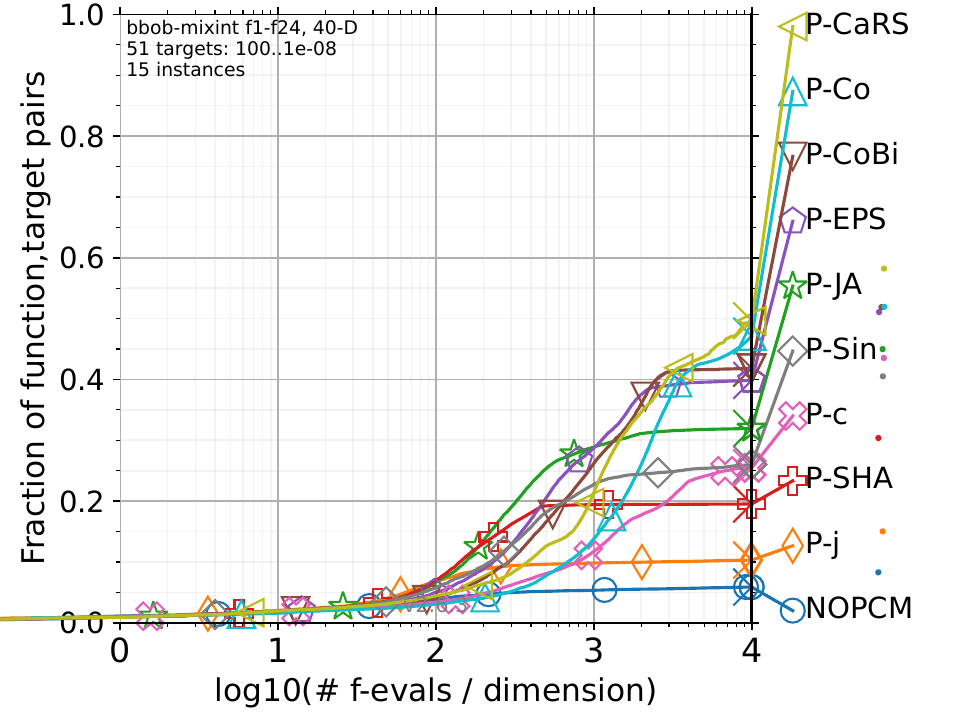}
}
\subfloat[$n=80$]{
\includegraphics[width=\width\textwidth]{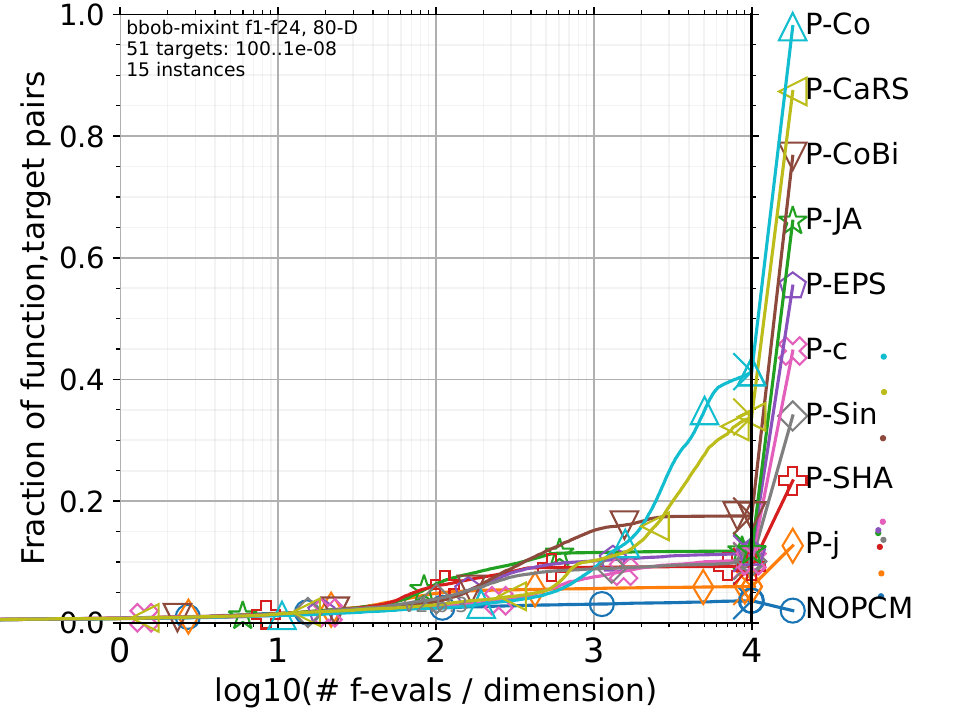}
}
\subfloat[$n=160$]{
\includegraphics[width=\width\textwidth]{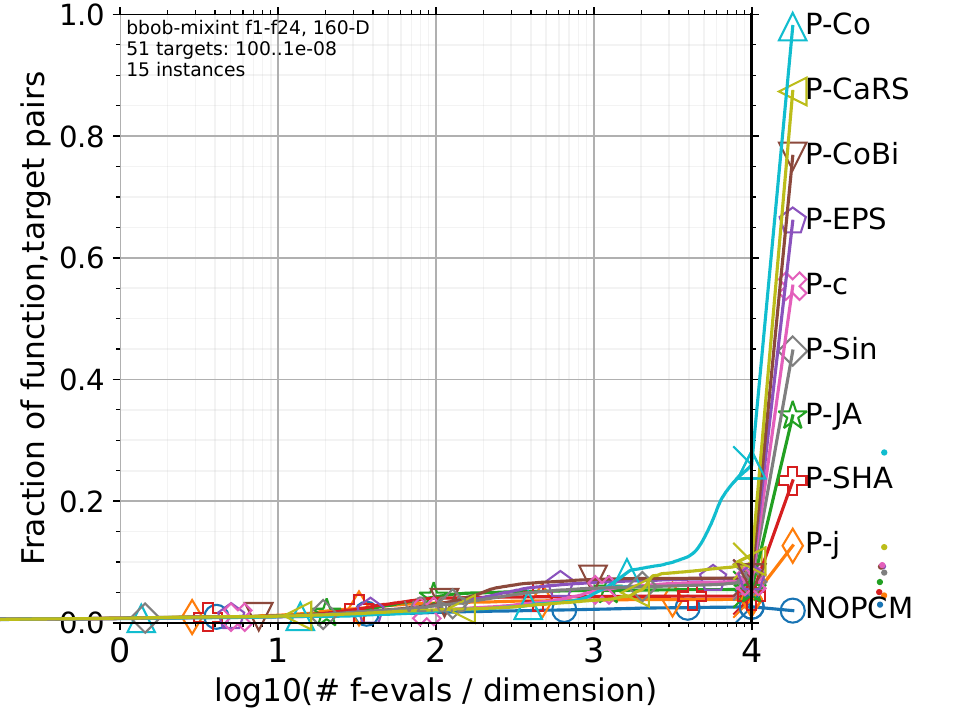}
}
\caption{Comparison of the nine PCMs and NOPCM  with the best/2 mutation strategy and the Lamarckian repair method on the 24 \texttt{bbob-mixint} functions for $n \in \{5, 10, 20, 40, 80, 160\}$.}
\label{supfig:vs_de_best_2_Lamarckian}
\end{figure*}

\begin{figure*}[htbp]
\newcommand{\width}{0.31}
\centering
\subfloat[$n=5$]{
\includegraphics[width=\width\textwidth]{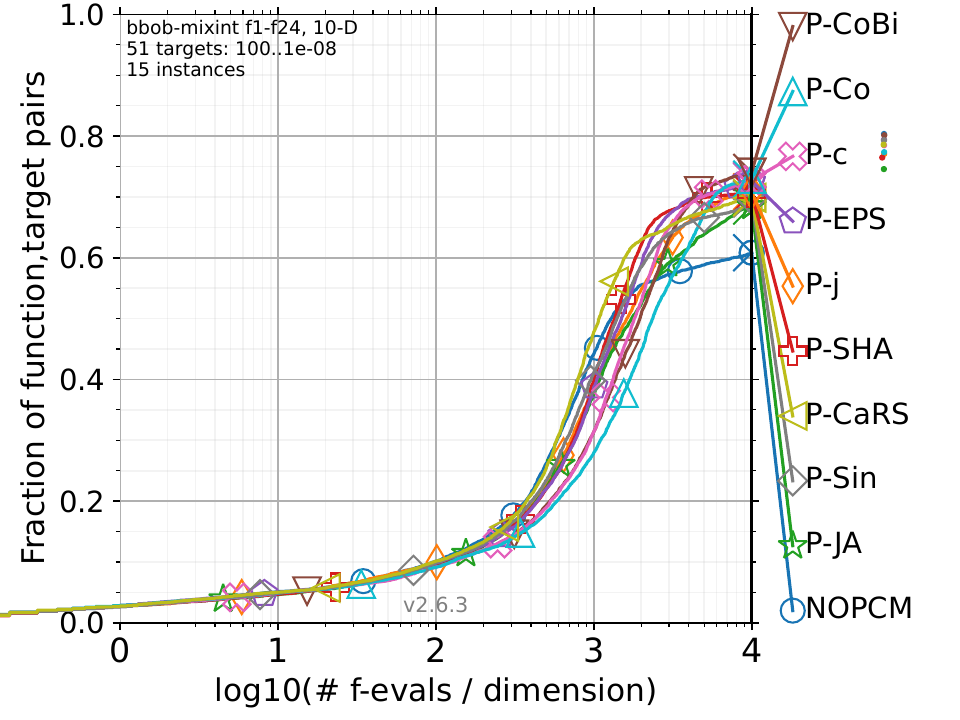}
}
\subfloat[$n=10$]{
\includegraphics[width=\width\textwidth]{./figs/vs_de/current_to_rand_1_mu100_conventional_Baldwin/NOPCM_P-j_P-JA_P-SHA_P-EPS_P-CoB_P-c_et_al/pprldmany_10D_noiselessall.pdf}
}
\subfloat[$n=20$]{
\includegraphics[width=\width\textwidth]{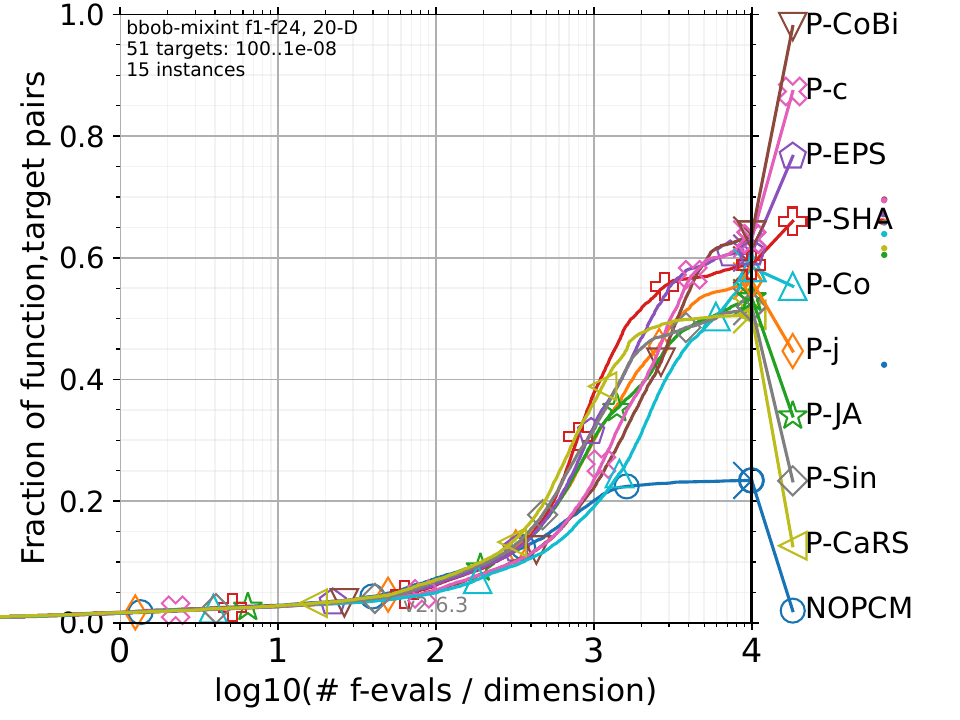}
}
\\
\subfloat[$n=40$]{
\includegraphics[width=\width\textwidth]{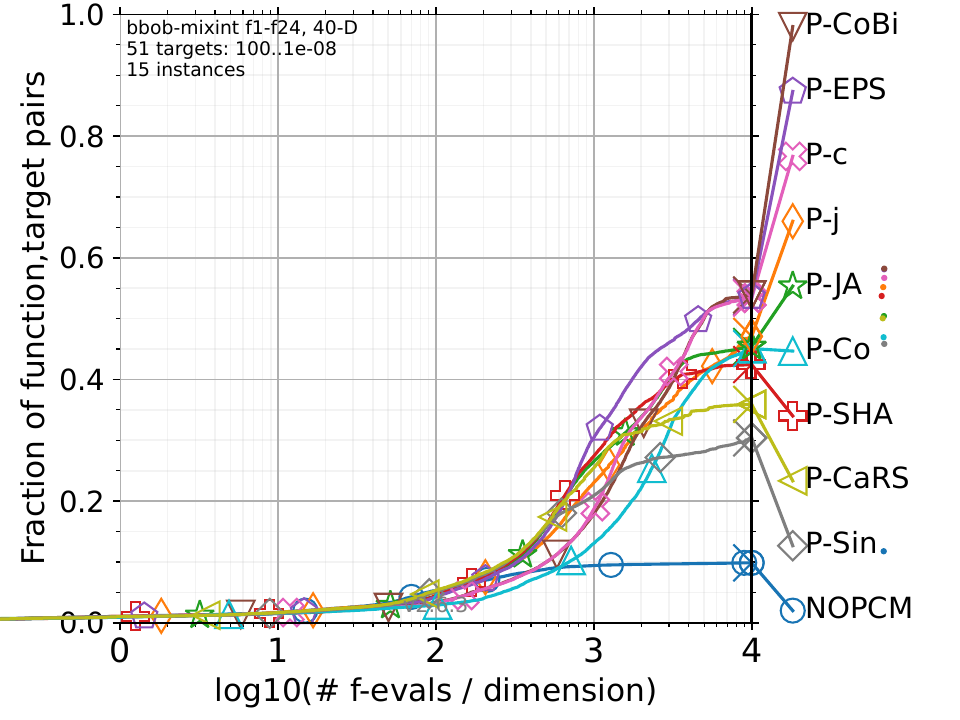}
}
\subfloat[$n=80$]{
\includegraphics[width=\width\textwidth]{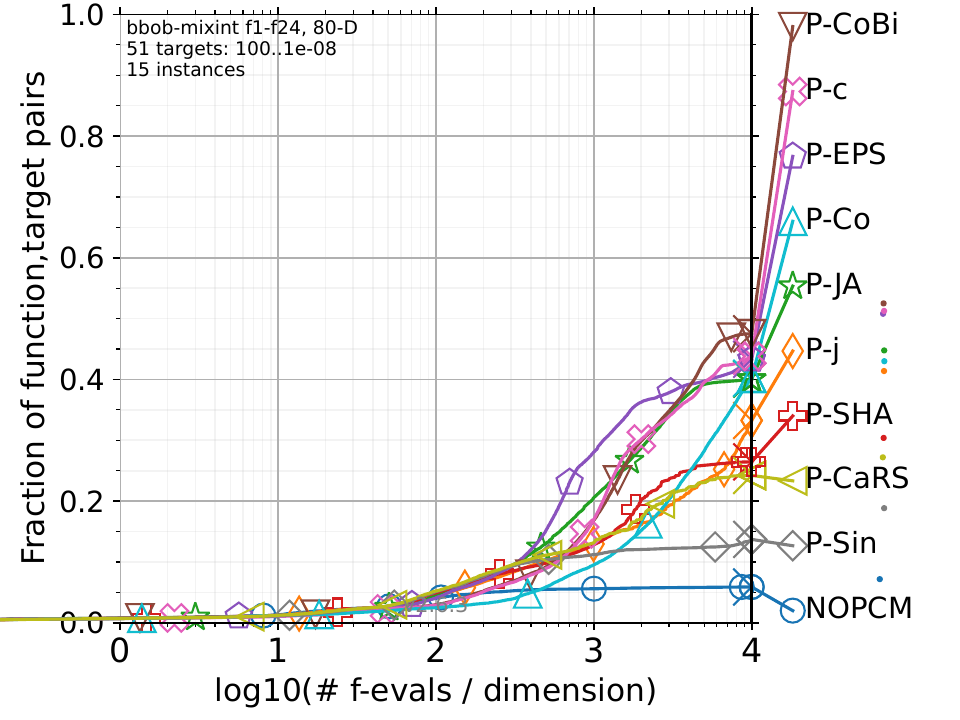}
}
\subfloat[$n=160$]{
\includegraphics[width=\width\textwidth]{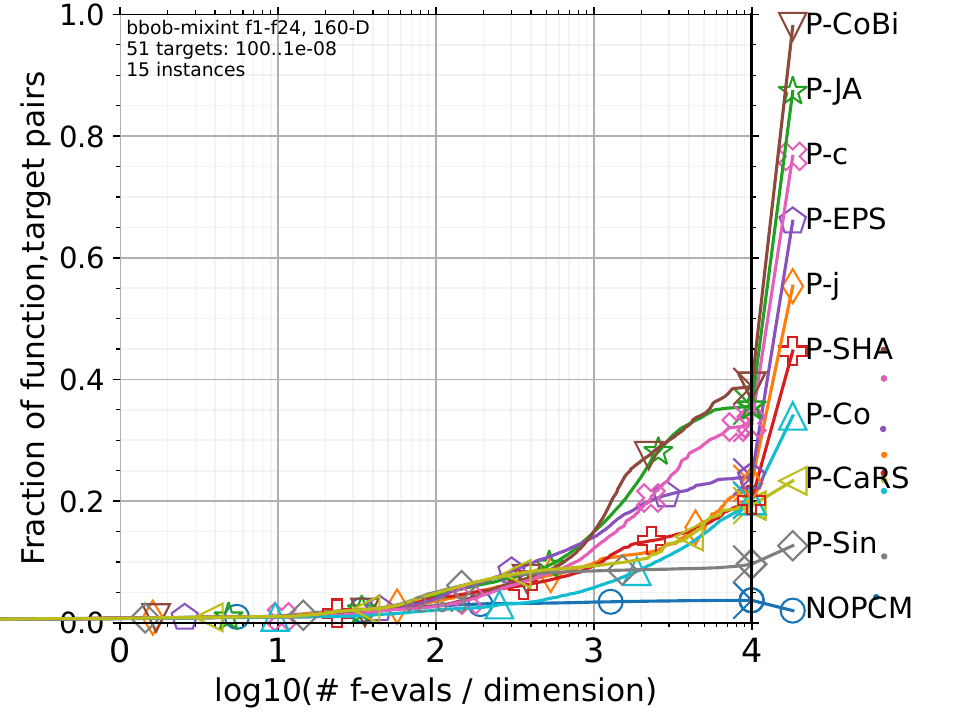}
}
\caption{Comparison of the nine PCMs and NOPCM  with the current-to-rand/1 mutation strategy and the Baldwinian repair method on the 24 \texttt{bbob-mixint} functions for $n \in \{5, 10, 20, 40, 80, 160\}$.}
\label{supfig:vs_de_current_to_rand_1_Baldwin}
\subfloat[$n=5$]{
\includegraphics[width=\width\textwidth]{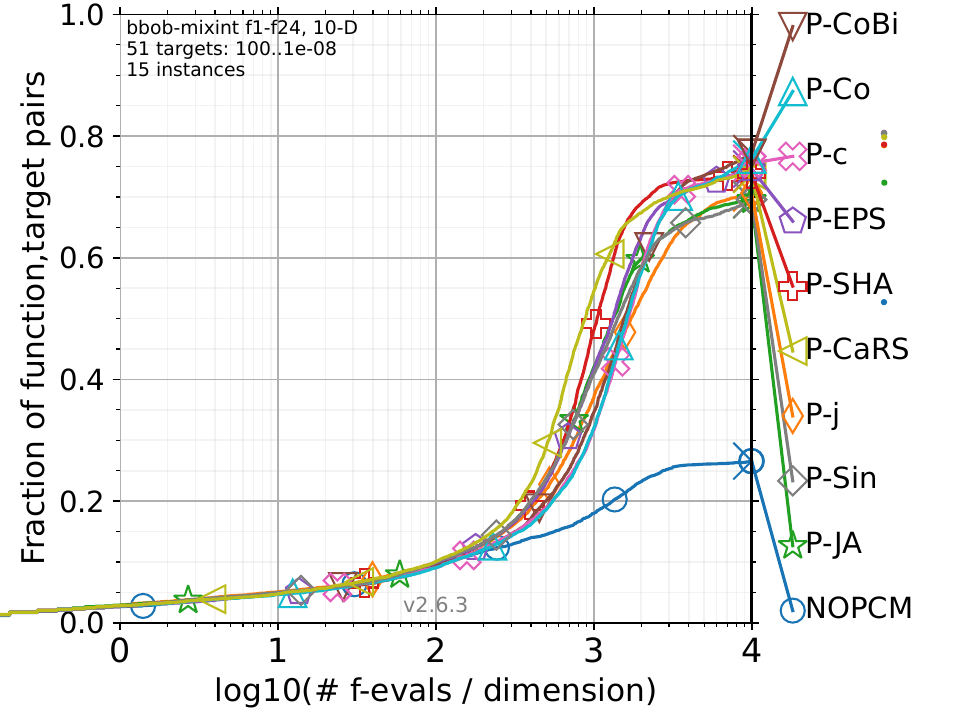}
}
\subfloat[$n=10$]{
\includegraphics[width=\width\textwidth]{./figs/vs_de/current_to_rand_1_mu100_conventional_Lamarckian/NOPCM_P-j_P-JA_P-SHA_P-EPS_P-CoB_P-c_et_al/pprldmany_10D_noiselessall.pdf}
}
\subfloat[$n=20$]{
\includegraphics[width=\width\textwidth]{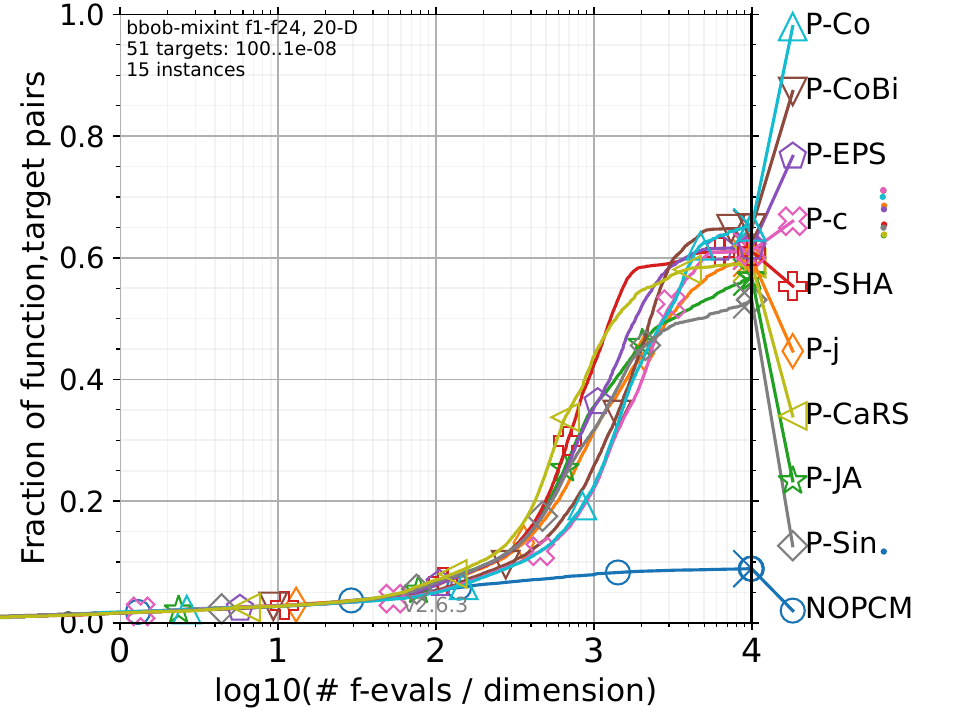}
}
\\
\subfloat[$n=40$]{
\includegraphics[width=\width\textwidth]{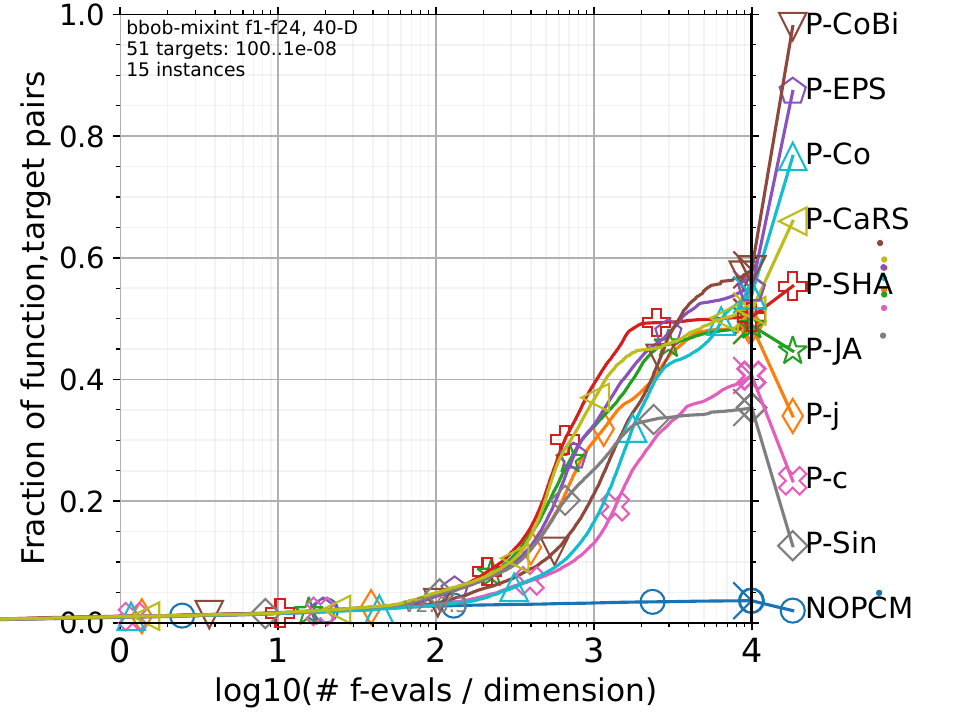}
}
\subfloat[$n=80$]{
\includegraphics[width=\width\textwidth]{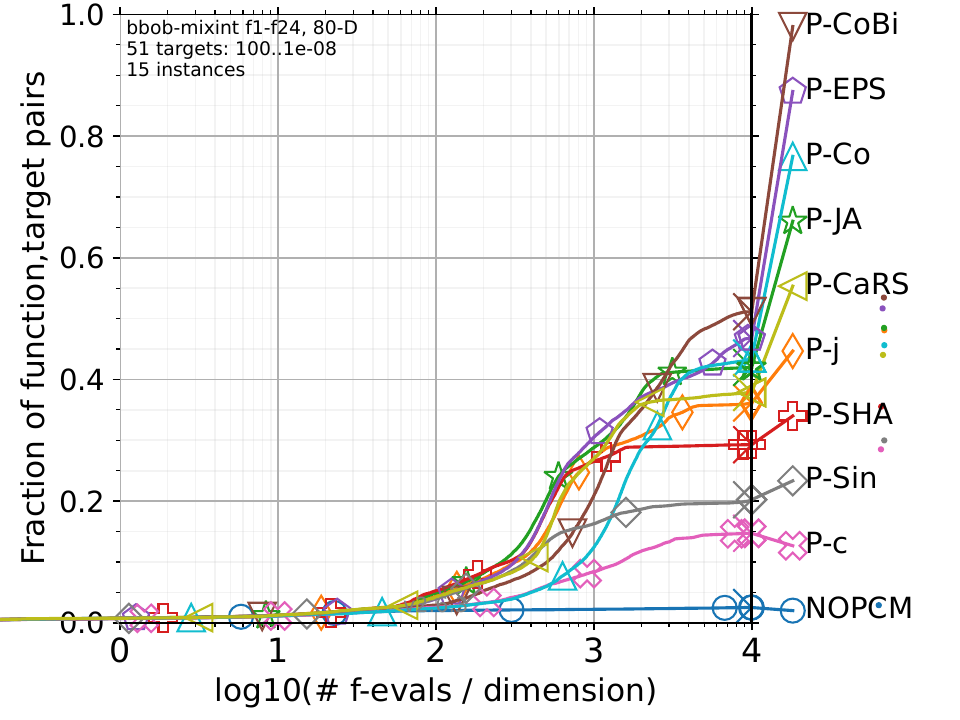}
}
\subfloat[$n=160$]{
\includegraphics[width=\width\textwidth]{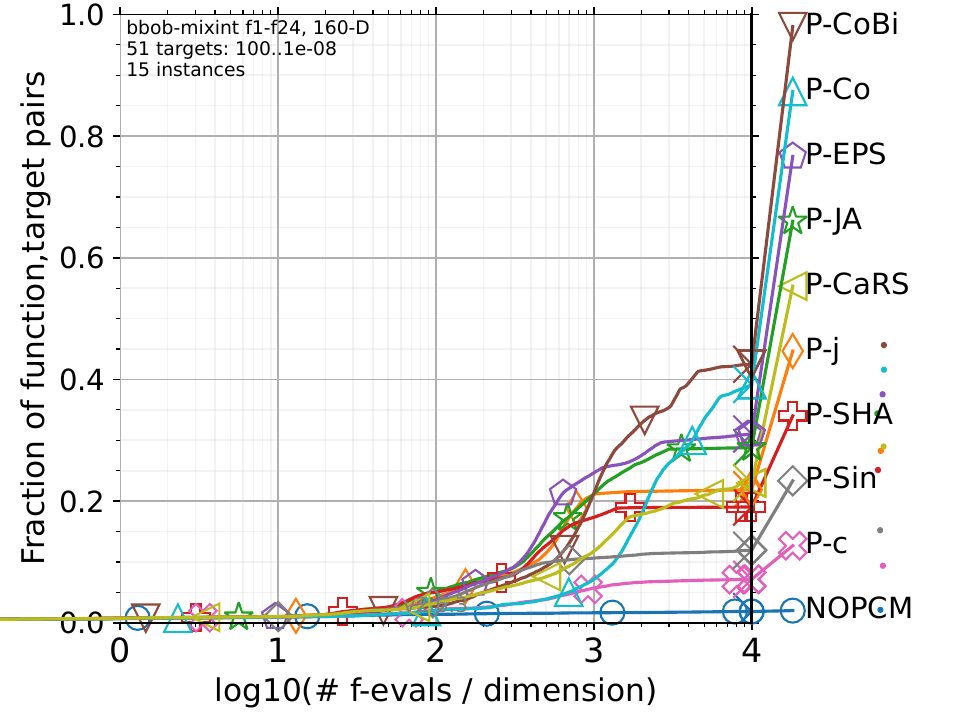}
}
\caption{Comparison of the nine PCMs and NOPCM  with the current-to-rand/1 mutation strategy and the Lamarckian repair method on the 24 \texttt{bbob-mixint} functions for $n \in \{5, 10, 20, 40, 80, 160\}$.}
\label{supfig:vs_de_current_to_rand_1_Lamarckian}
\end{figure*}

\begin{figure*}[htbp]
\newcommand{\width}{0.31}
\centering
\subfloat[$n=5$]{
\includegraphics[width=\width\textwidth]{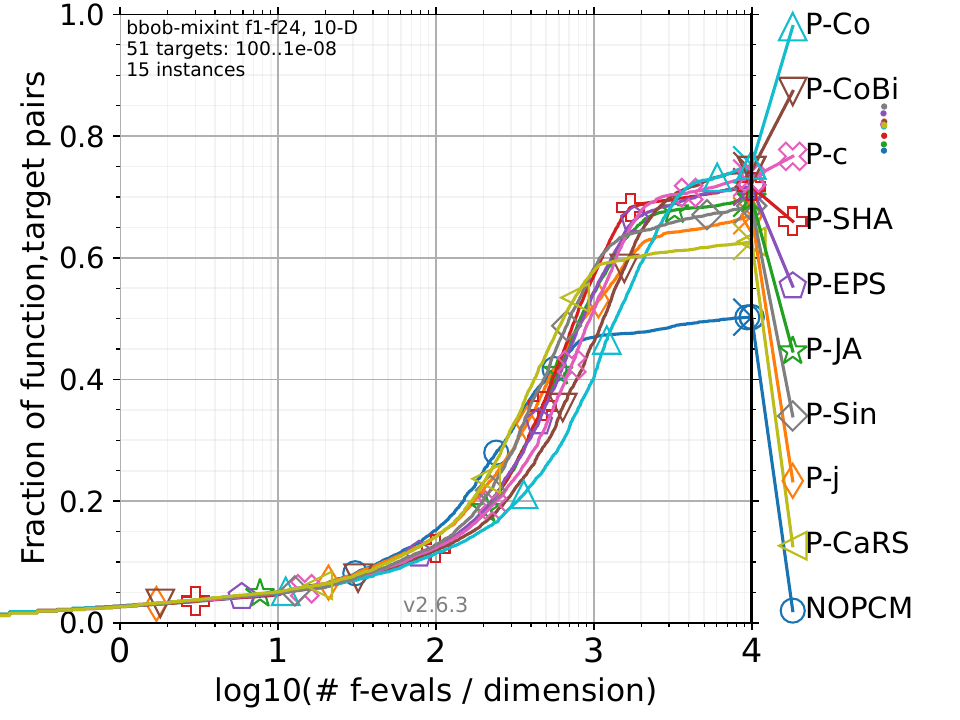}
}
\subfloat[$n=10$]{
\includegraphics[width=\width\textwidth]{./figs/vs_de/current_to_best_1_mu100_conventional_Baldwin/NOPCM_P-j_P-JA_P-SHA_P-EPS_P-CoB_P-c_et_al/pprldmany_10D_noiselessall.pdf}
}
\subfloat[$n=20$]{
\includegraphics[width=\width\textwidth]{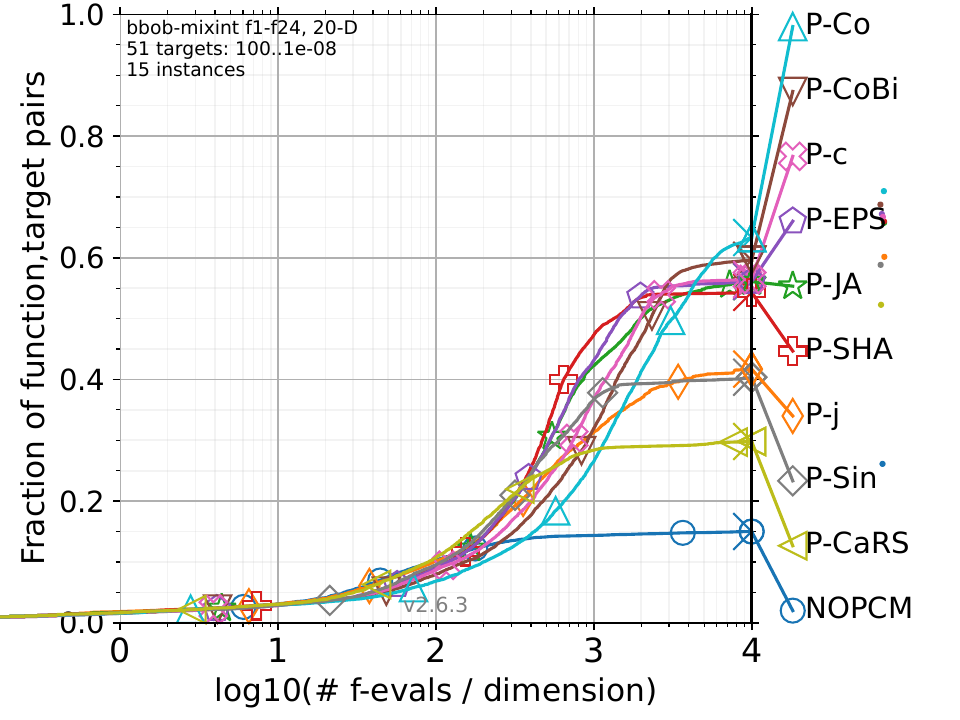}
}
\\
\subfloat[$n=40$]{
\includegraphics[width=\width\textwidth]{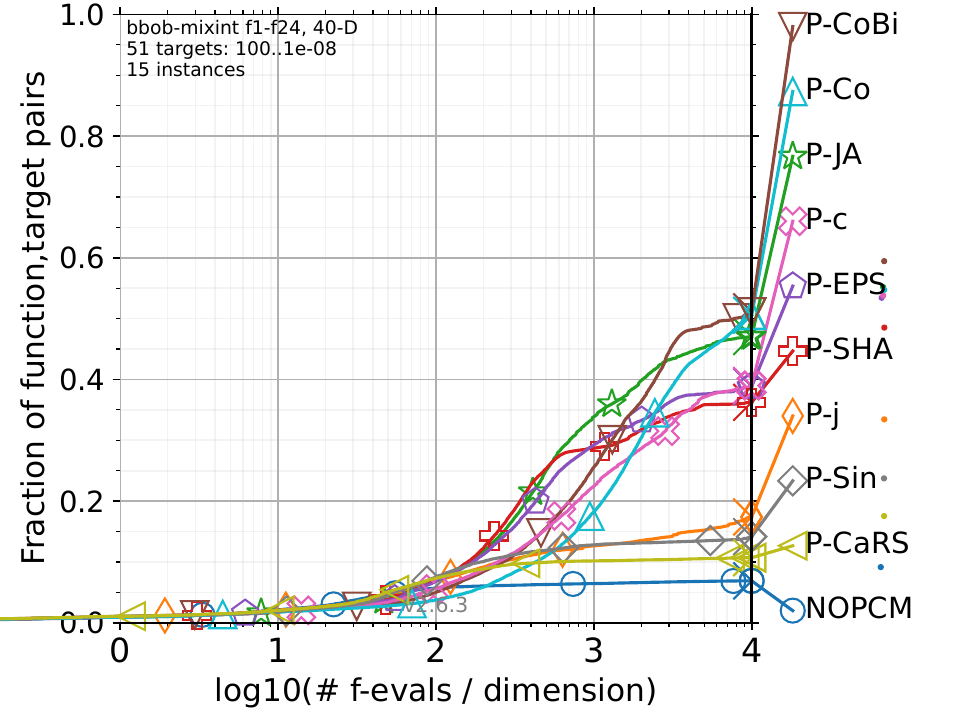}
}
\subfloat[$n=80$]{
\includegraphics[width=\width\textwidth]{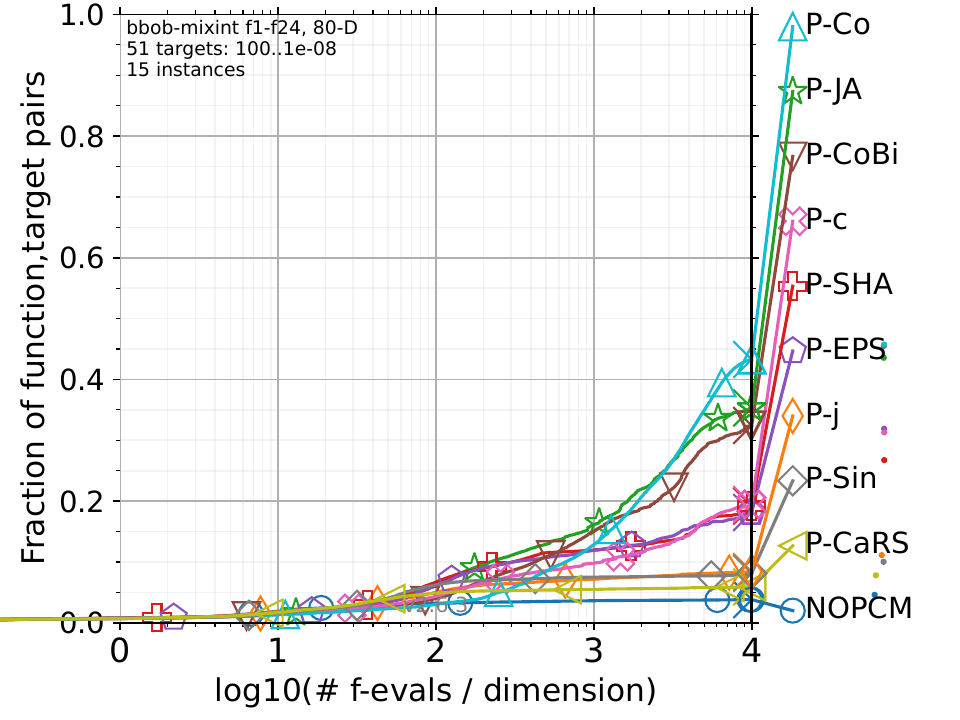}
}
\subfloat[$n=160$]{
\includegraphics[width=\width\textwidth]{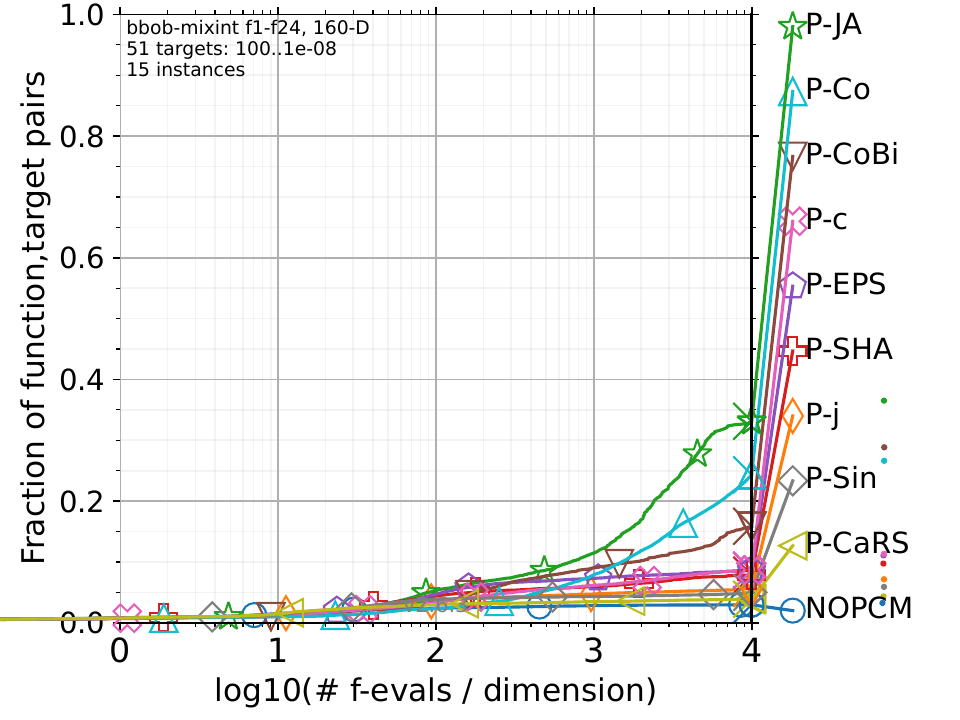}
}
\caption{Comparison of the nine PCMs and NOPCM  with the current-to-best/1 mutation strategy and the Baldwinian repair method on the 24 \texttt{bbob-mixint} functions for $n \in \{5, 10, 20, 40, 80, 160\}$.}
\label{supfig:vs_de_current_to_best_1_Baldwin}
\subfloat[$n=5$]{
\includegraphics[width=\width\textwidth]{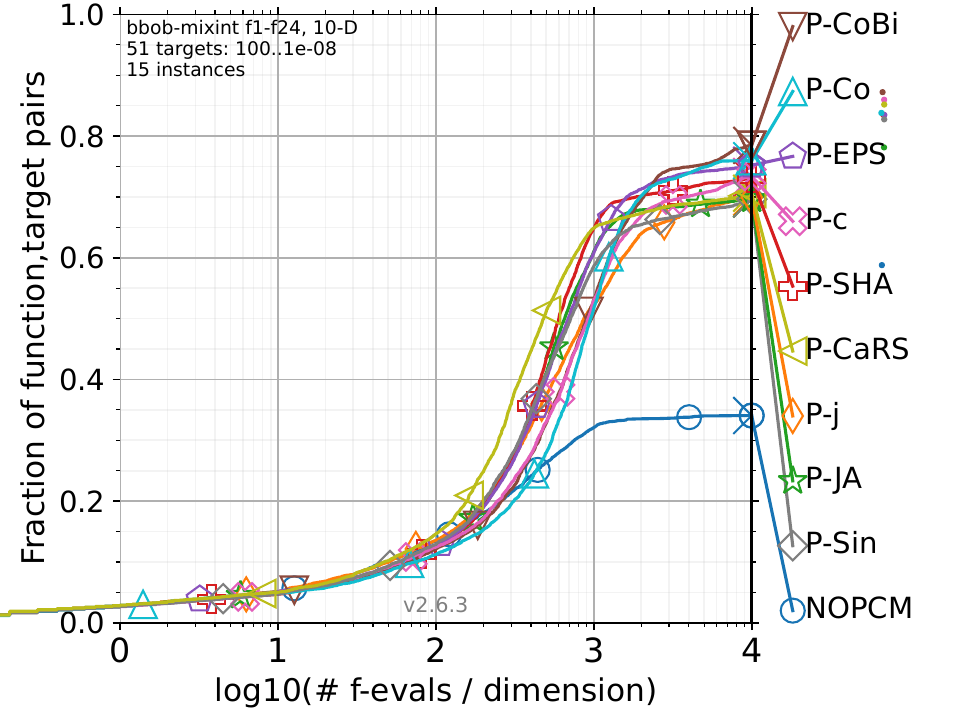}
}
\subfloat[$n=10$]{
\includegraphics[width=\width\textwidth]{./figs/vs_de/current_to_best_1_mu100_conventional_Lamarckian/NOPCM_P-j_P-JA_P-SHA_P-EPS_P-CoB_P-c_et_al/pprldmany_10D_noiselessall.pdf}
}
\subfloat[$n=20$]{
\includegraphics[width=\width\textwidth]{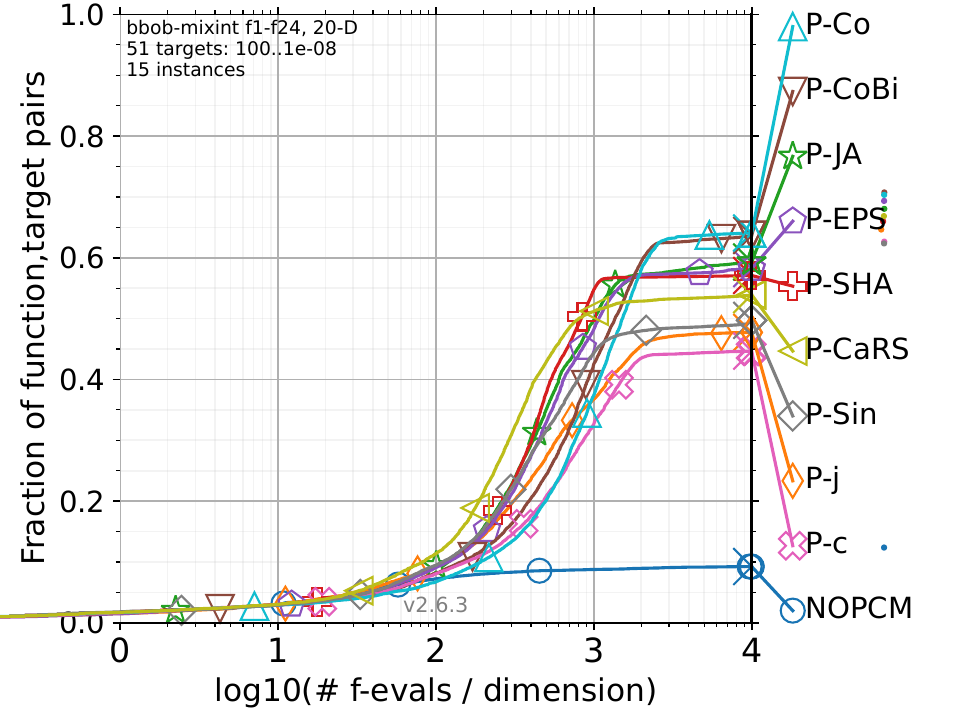}
}
\\
\subfloat[$n=40$]{
\includegraphics[width=\width\textwidth]{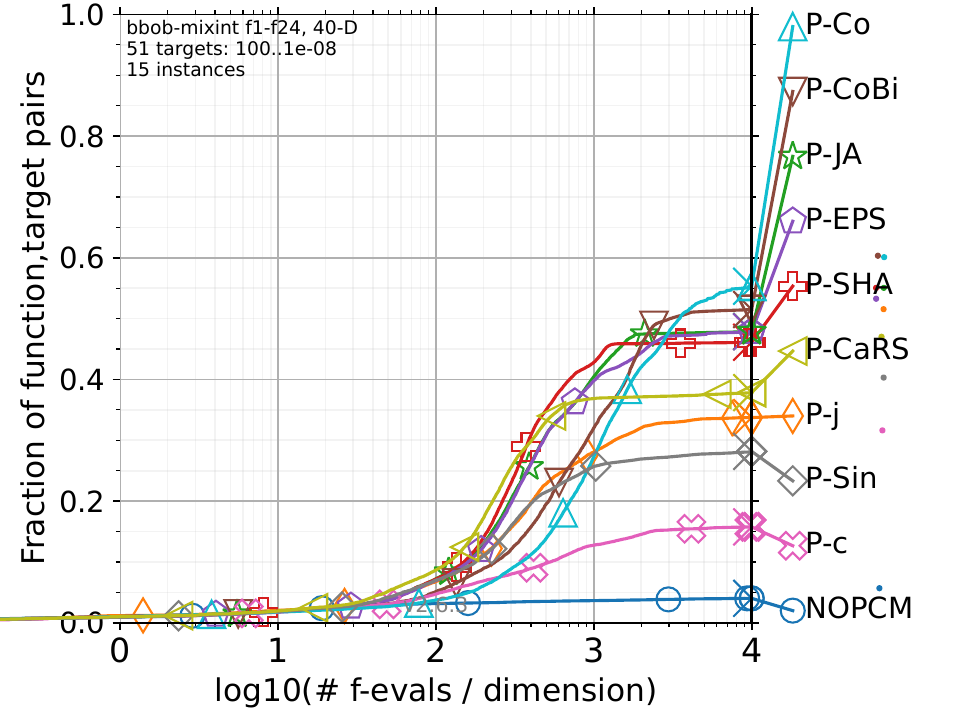}
}
\subfloat[$n=80$]{
\includegraphics[width=\width\textwidth]{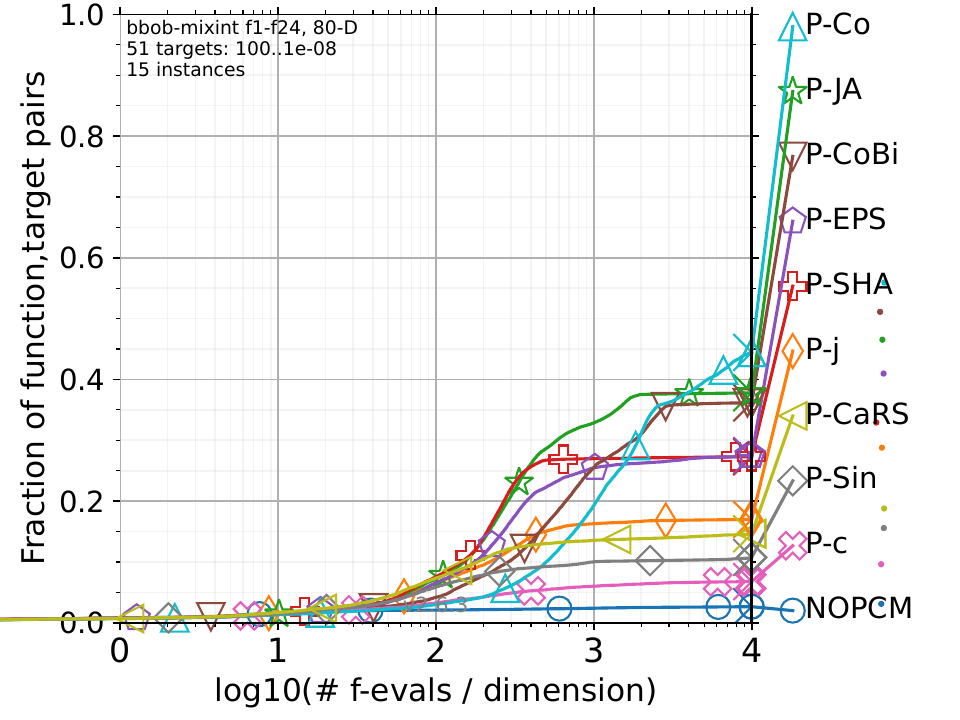}
}
\subfloat[$n=160$]{
\includegraphics[width=\width\textwidth]{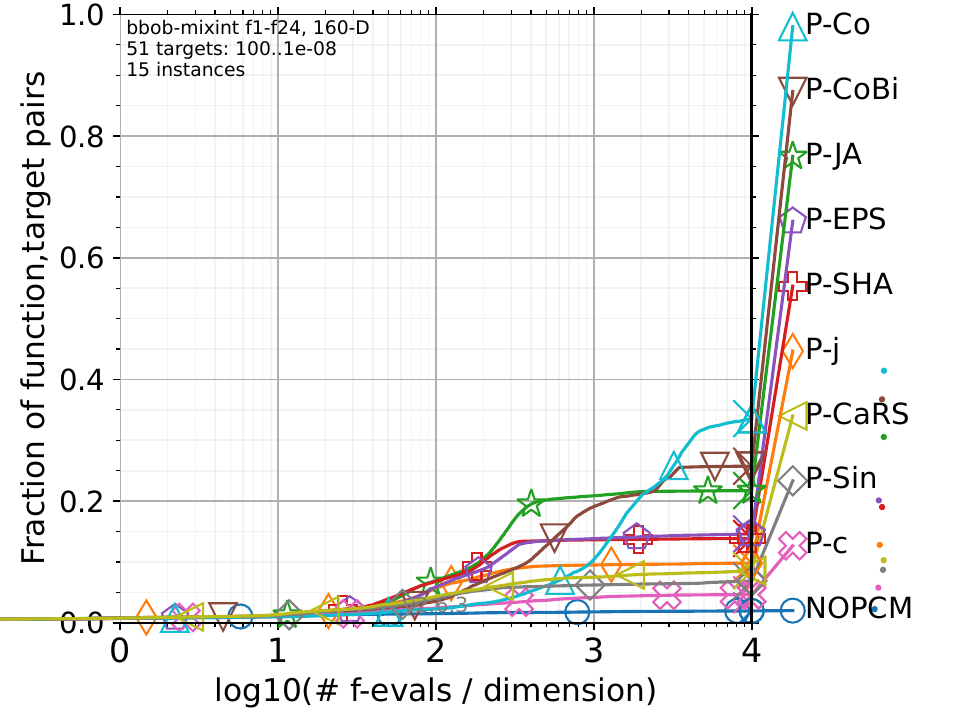}
}
\caption{Comparison of the nine PCMs and NOPCM  with the current-to-best/1 mutation strategy and the Lamarckian repair method on the 24 \texttt{bbob-mixint} functions for $n \in \{5, 10, 20, 40, 80, 160\}$.}
\label{supfig:vs_de_current_to_best_1_Lamarckian}
\end{figure*}

\begin{figure*}[htbp]
\newcommand{\width}{0.31}
\centering
\subfloat[$n=5$]{
\includegraphics[width=\width\textwidth]{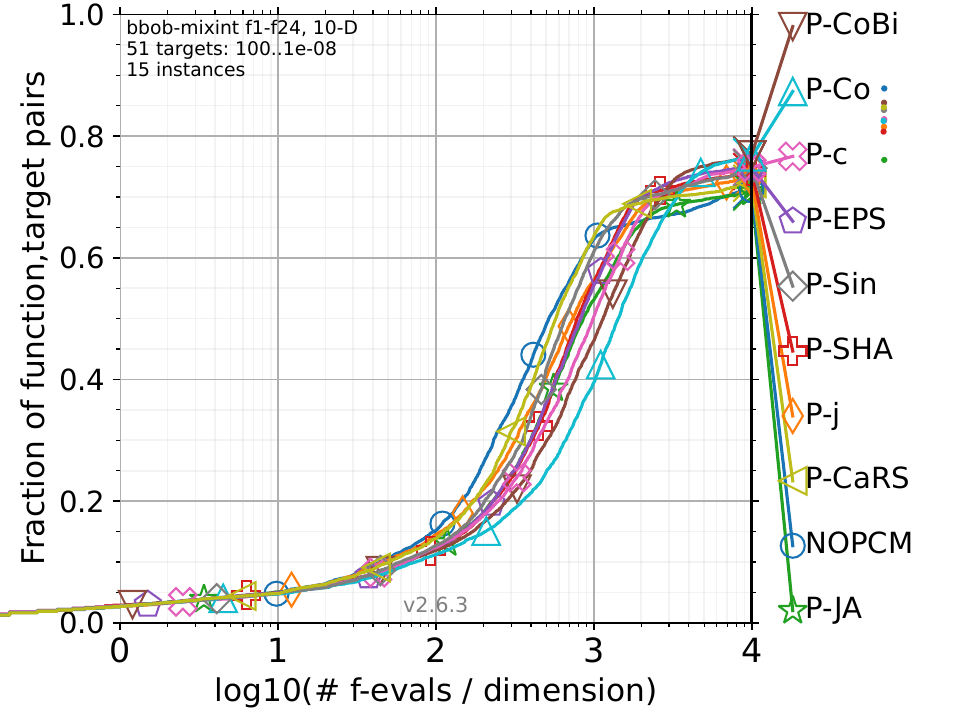}
}
\subfloat[$n=10$]{
\includegraphics[width=\width\textwidth]{./figs/vs_de/current_to_pbest_1_mu100_conventional_Baldwin/NOPCM_P-j_P-JA_P-SHA_P-EPS_P-CoB_P-c_et_al/pprldmany_10D_noiselessall.pdf}
}
\subfloat[$n=20$]{
\includegraphics[width=\width\textwidth]{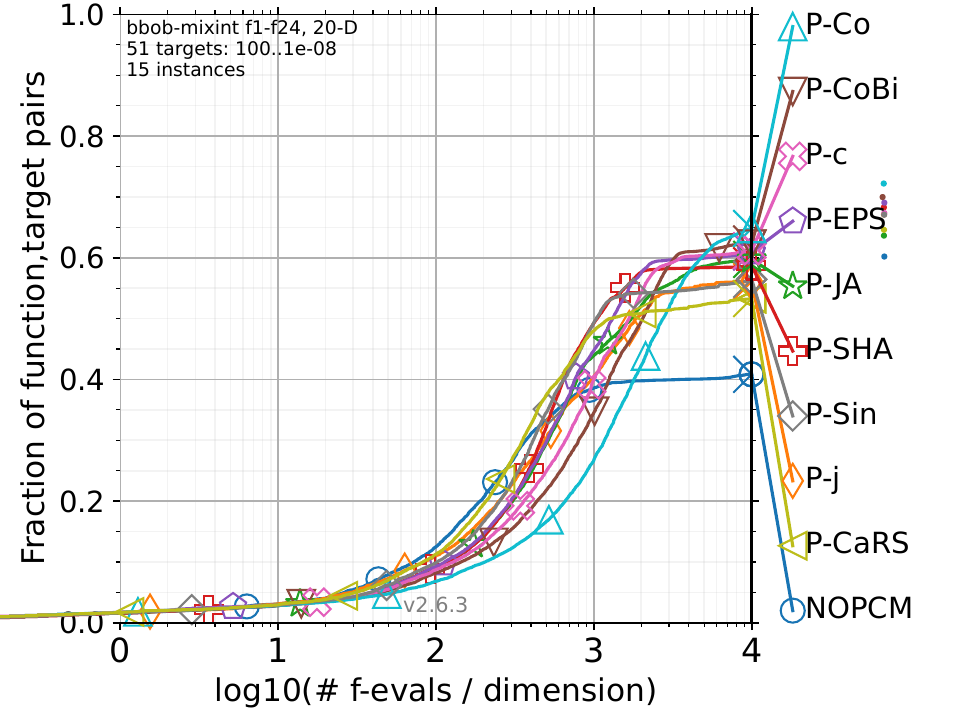}
}
\\
\subfloat[$n=40$]{
\includegraphics[width=\width\textwidth]{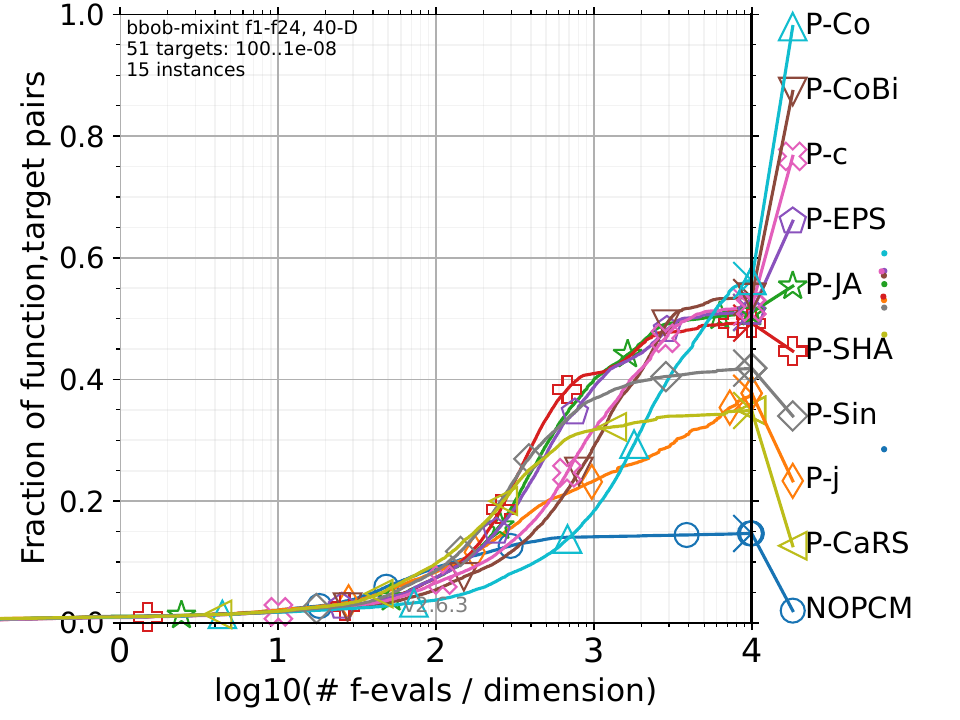}
}
\subfloat[$n=80$]{
\includegraphics[width=\width\textwidth]{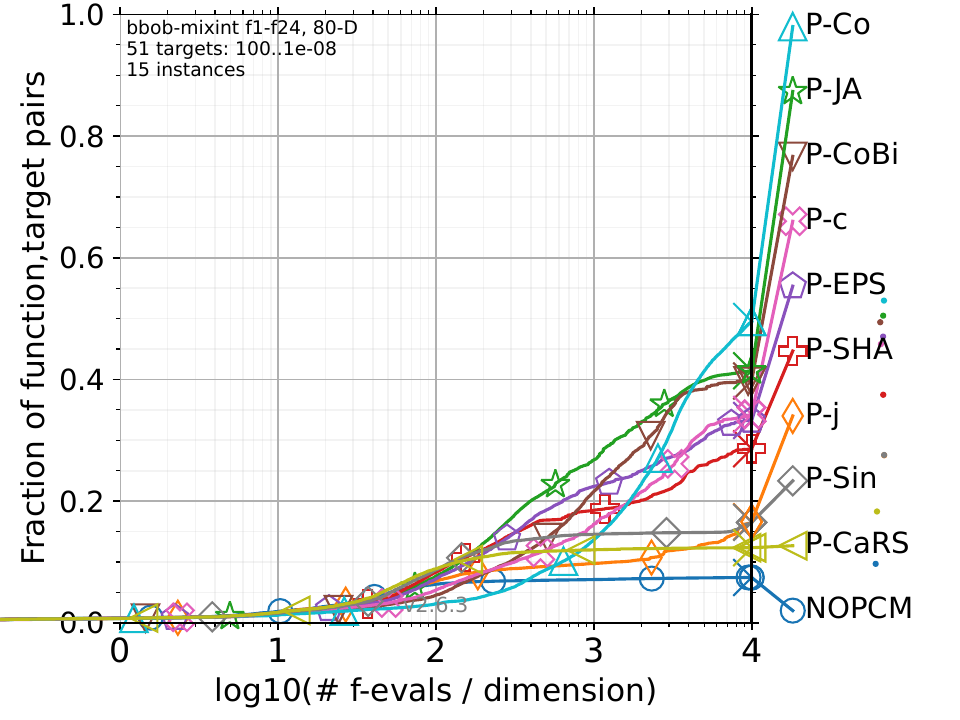}
}
\subfloat[$n=160$]{
\includegraphics[width=\width\textwidth]{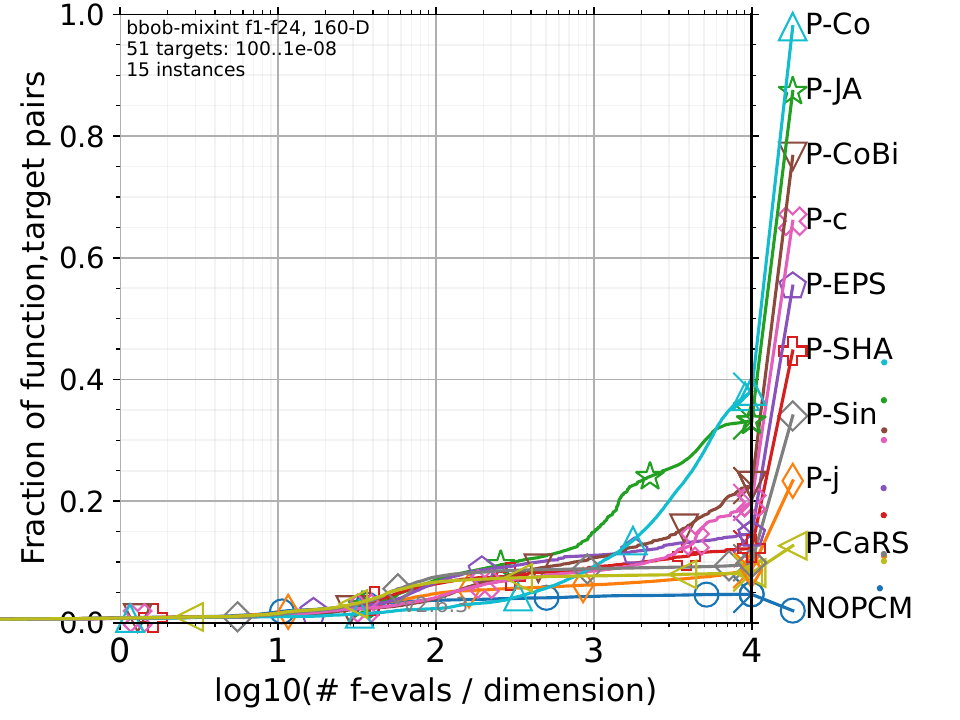}
}
\caption{Comparison of the nine PCMs and NOPCM  with the current-to-$p$best/1 mutation strategy and the Baldwinian repair method on the 24 \texttt{bbob-mixint} functions for $n \in \{5, 10, 20, 40, 80, 160\}$.}
\label{supfig:vs_de_current_to_pbest_1_Baldwin}
\subfloat[$n=5$]{
\includegraphics[width=\width\textwidth]{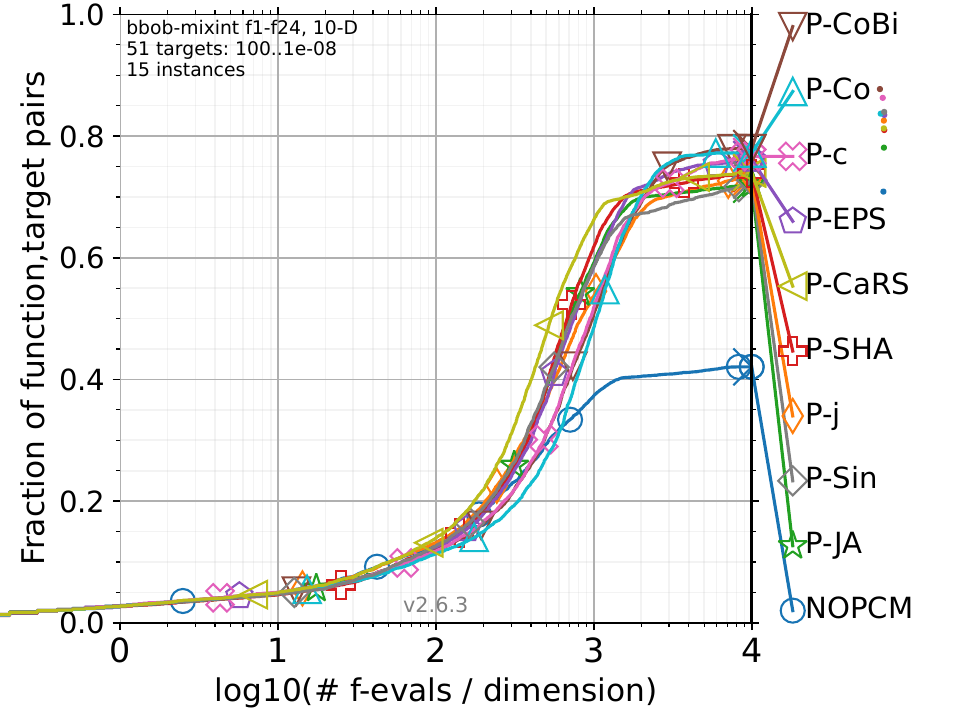}
}
\subfloat[$n=10$]{
\includegraphics[width=\width\textwidth]{./figs/vs_de/current_to_pbest_1_mu100_conventional_Lamarckian/NOPCM_P-j_P-JA_P-SHA_P-EPS_P-CoB_P-c_et_al/pprldmany_10D_noiselessall.pdf}
}
\subfloat[$n=20$]{
\includegraphics[width=\width\textwidth]{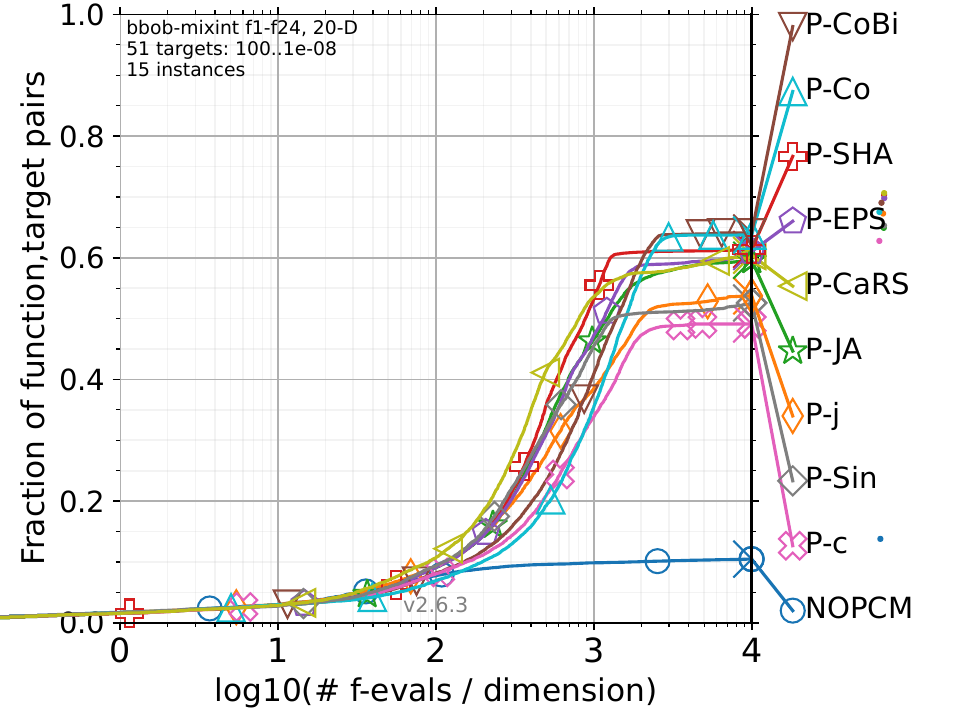}
}
\\
\subfloat[$n=40$]{
\includegraphics[width=\width\textwidth]{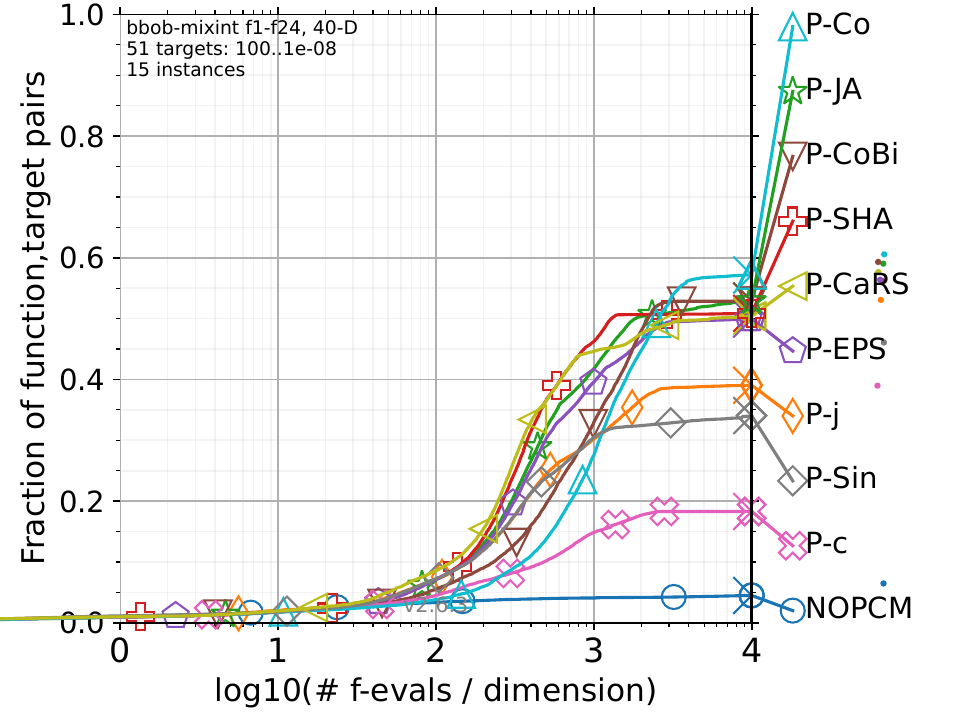}
}
\subfloat[$n=80$]{
\includegraphics[width=\width\textwidth]{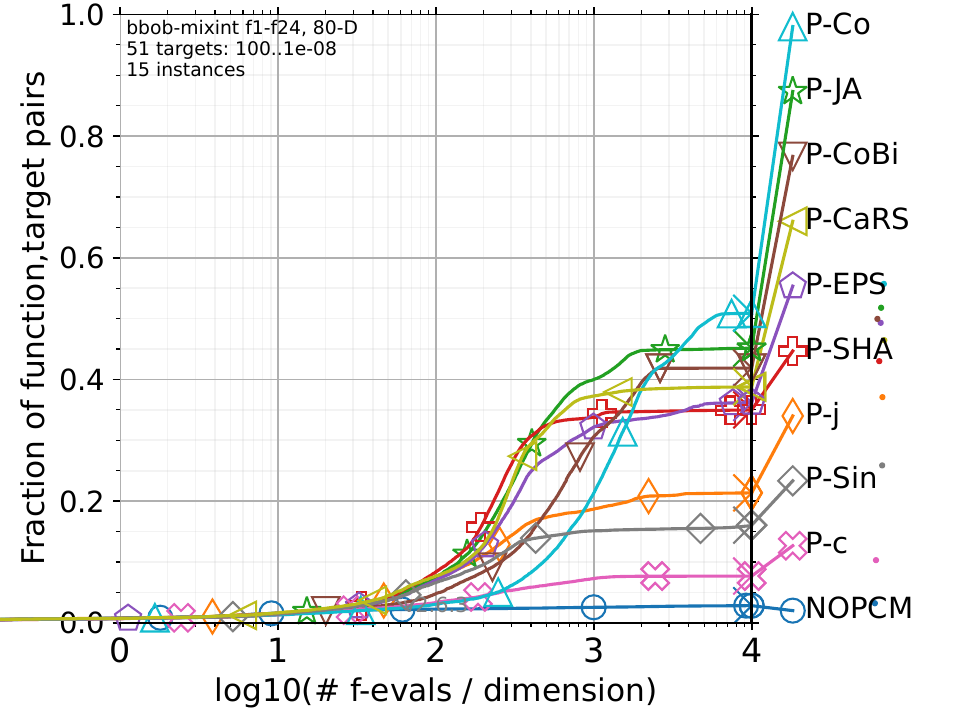}
}
\subfloat[$n=160$]{
\includegraphics[width=\width\textwidth]{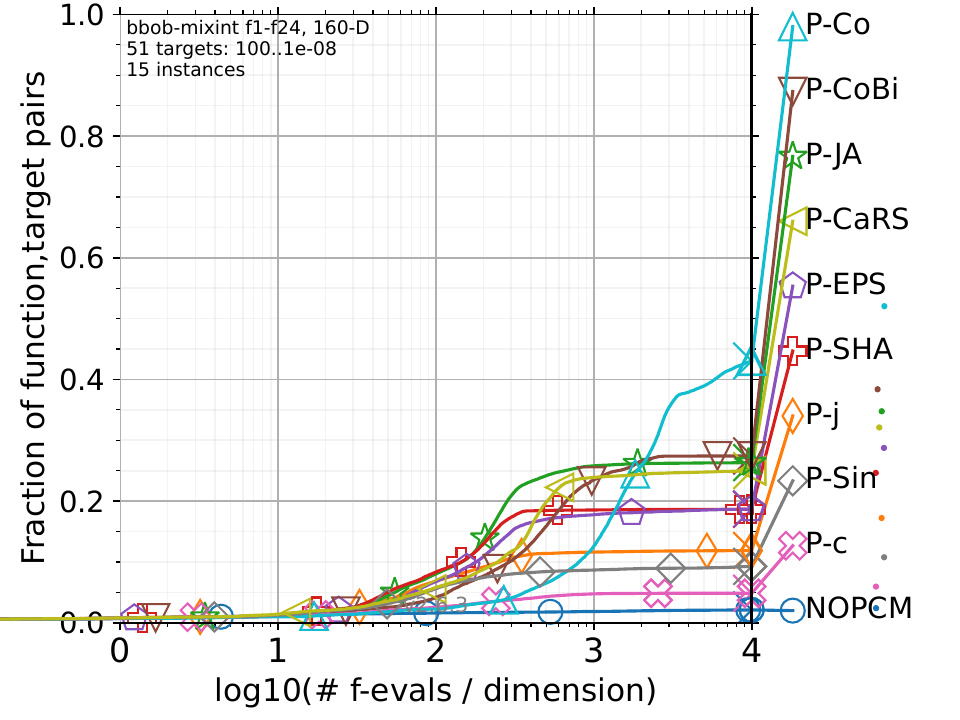}
}
\caption{Comparison of the nine PCMs and NOPCM  with the current-to-$p$best/1 mutation strategy and the Lamarckian repair method on the 24 \texttt{bbob-mixint} functions for $n \in \{5, 10, 20, 40, 80, 160\}$.}
\label{supfig:vs_de_current_to_pbest_1_Lamarckian}
\end{figure*}

\begin{figure*}[htbp]
\newcommand{\width}{0.31}
\centering
\subfloat[$n=5$]{
\includegraphics[width=\width\textwidth]{./figs/vs_de/rand_to_pbest_1_mu100_conventional_Baldwin/NOPCM_P-j_P-JA_P-SHA_P-EPS_P-CoB_P-c_et_al/pprldmany_10D_noiselessall.pdf}
}
\subfloat[$n=10$]{
\includegraphics[width=\width\textwidth]{./figs/vs_de/rand_to_pbest_1_mu100_conventional_Baldwin/NOPCM_P-j_P-JA_P-SHA_P-EPS_P-CoB_P-c_et_al/pprldmany_10D_noiselessall.pdf}
}
\subfloat[$n=20$]{
\includegraphics[width=\width\textwidth]{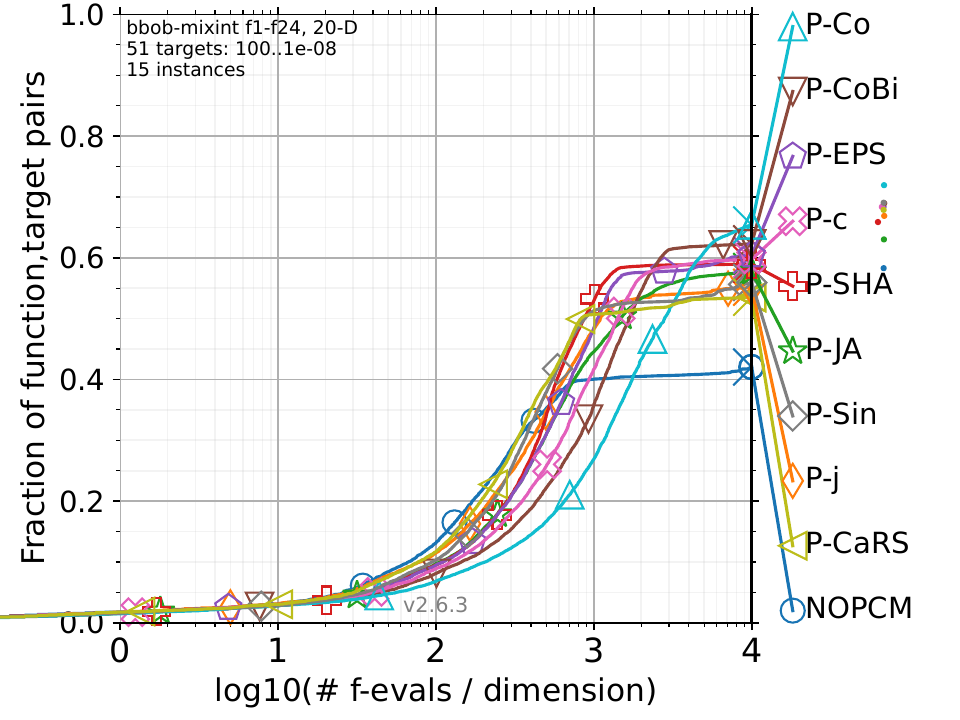}
}
\\
\subfloat[$n=40$]{
\includegraphics[width=\width\textwidth]{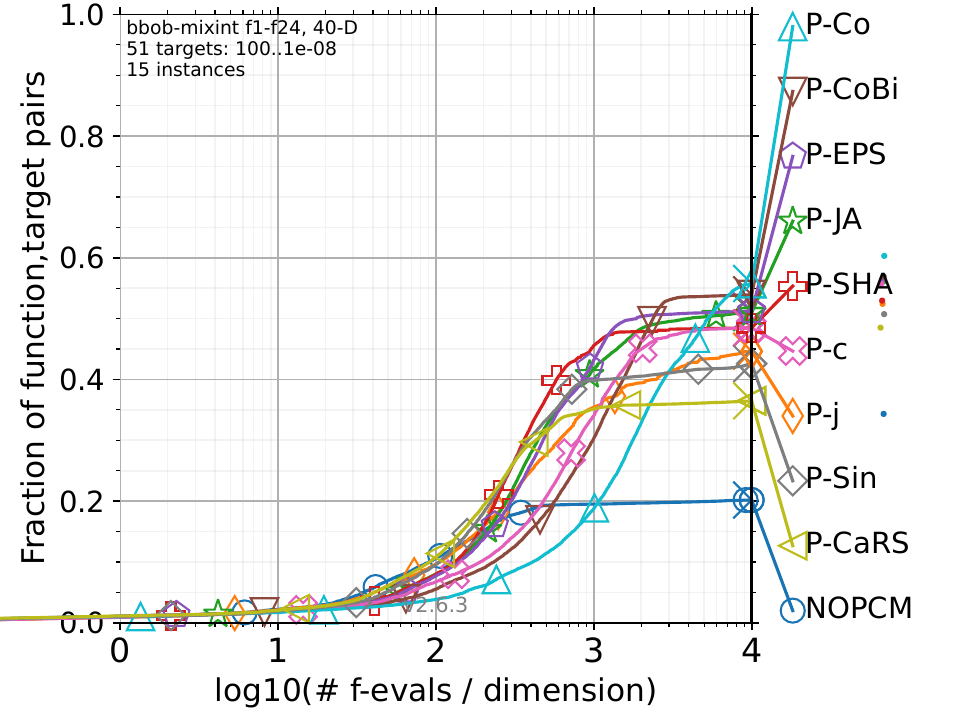}
}
\subfloat[$n=80$]{
\includegraphics[width=\width\textwidth]{./figs/vs_de/rand_to_pbest_1_mu100_conventional_Baldwin/NOPCM_P-j_P-JA_P-SHA_P-EPS_P-CoB_P-c_et_al/pprldmany_80D_noiselessall.pdf}
}
\subfloat[$n=160$]{
\includegraphics[width=\width\textwidth]{./figs/vs_de/rand_to_pbest_1_mu100_conventional_Baldwin/NOPCM_P-j_P-JA_P-SHA_P-EPS_P-CoB_P-c_et_al/pprldmany_160D_noiselessall.pdf}
}
\caption{Comparison of the nine PCMs and NOPCM  with the rand-to-$p$best/1 mutation strategy and the Baldwinian repair method on the 24 \texttt{bbob-mixint} functions for $n \in \{5, 10, 20, 40, 80, 160\}$.}
\label{supfig:vs_de_rand_to_pbest_1_Baldwin}
\subfloat[$n=5$]{
\includegraphics[width=\width\textwidth]{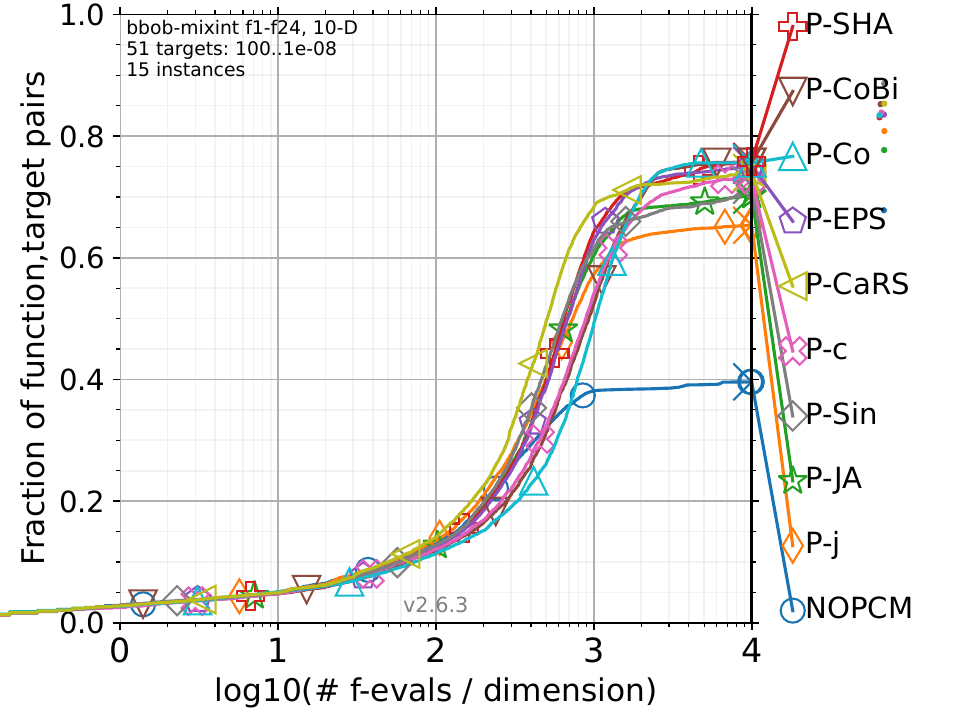}
}
\subfloat[$n=10$]{
\includegraphics[width=\width\textwidth]{./figs/vs_de/rand_to_pbest_1_mu100_conventional_Lamarckian/NOPCM_P-j_P-JA_P-SHA_P-EPS_P-CoB_P-c_et_al/pprldmany_10D_noiselessall.pdf}
}
\subfloat[$n=20$]{
\includegraphics[width=\width\textwidth]{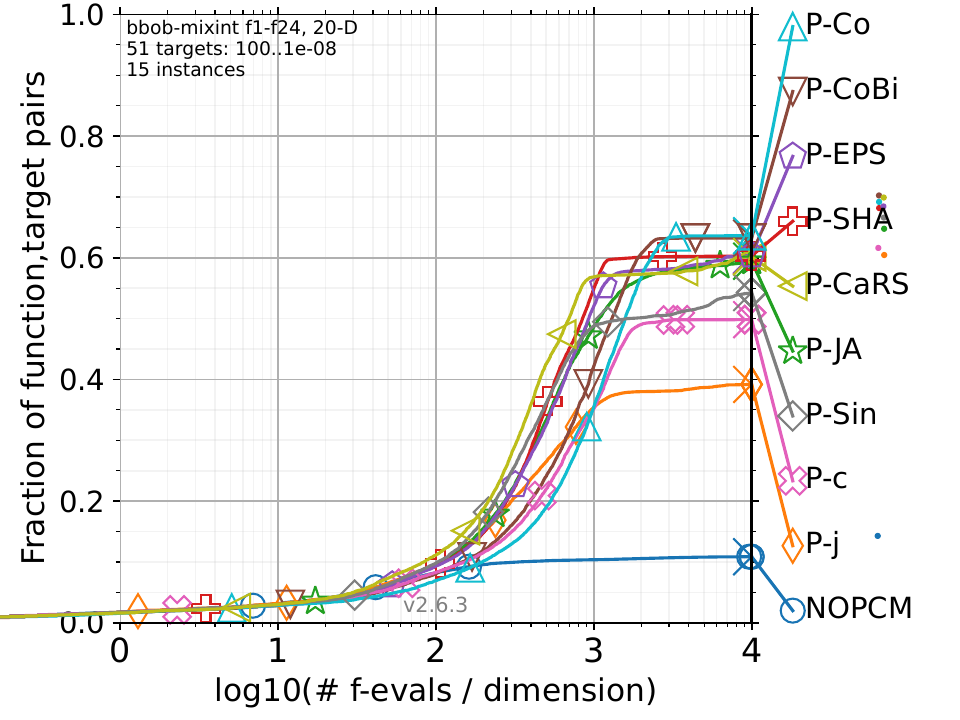}
}
\\
\subfloat[$n=40$]{
\includegraphics[width=\width\textwidth]{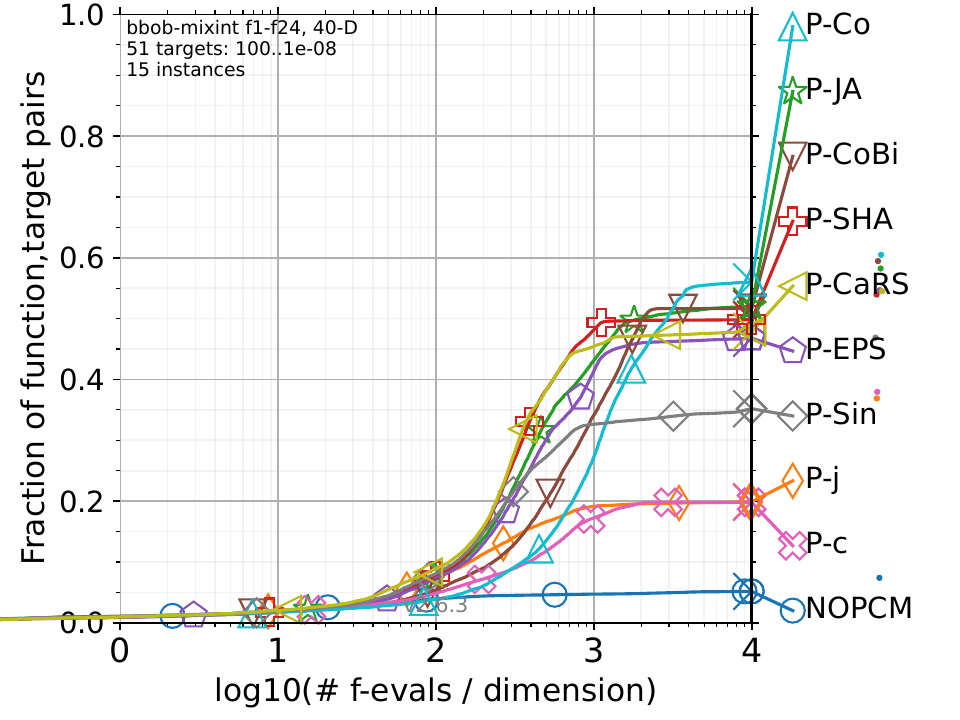}
}
\subfloat[$n=80$]{
\includegraphics[width=\width\textwidth]{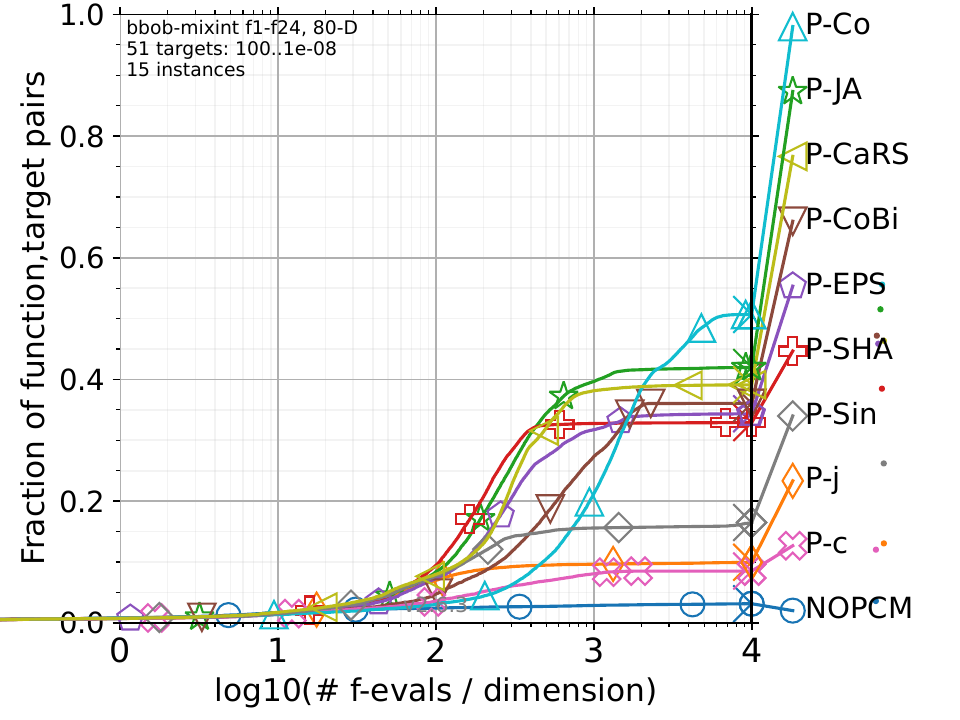}
}
\subfloat[$n=160$]{
\includegraphics[width=\width\textwidth]{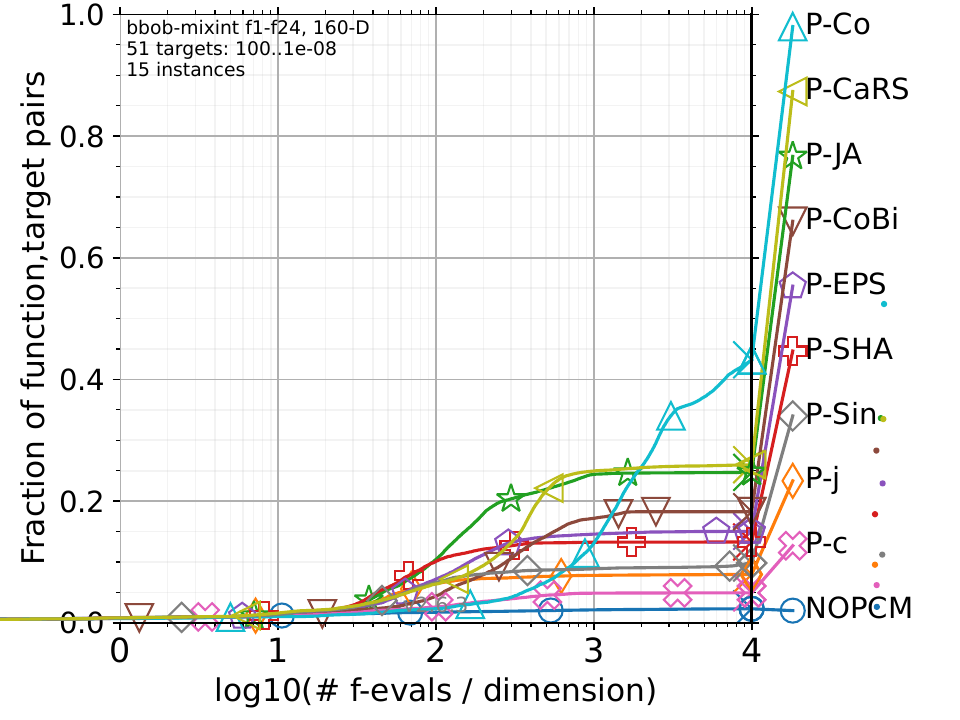}
}
\caption{Comparison of the nine PCMs and NOPCM  with the rand-to-$p$best/1 mutation strategy and the Lamarckian repair method on the 24 \texttt{bbob-mixint} functions for $n \in \{5, 10, 20, 40, 80, 160\}$.}
\label{supfig:vs_de_rand_to_pbest_1_Lamarckian}
\end{figure*}

\begin{figure*}[t]
\newcommand{\width}{0.31}
\centering
\subfloat[$n=5$]{
\includegraphics[width=\width\textwidth]{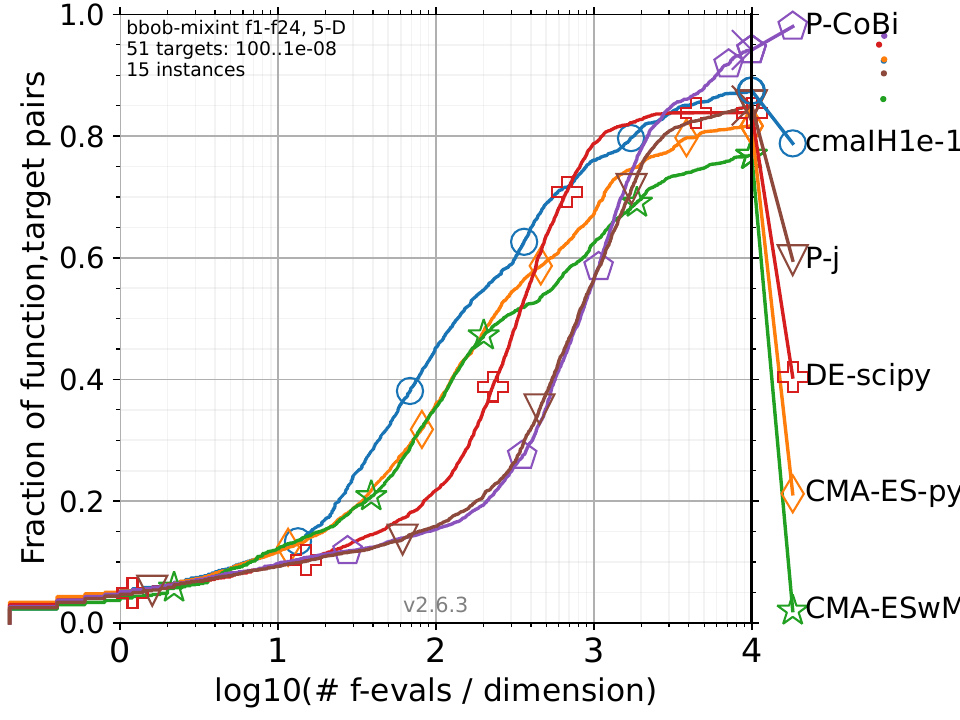}
}
\subfloat[$n=10$]{
\includegraphics[width=\width\textwidth]{./figs/vs_cma/cmaIH_CMA-E_CMA-E_DE-sc_P-CoB_P-j/pprldmany_10D_noiselessall.pdf}
}
\subfloat[$n=20$]{
\includegraphics[width=\width\textwidth]{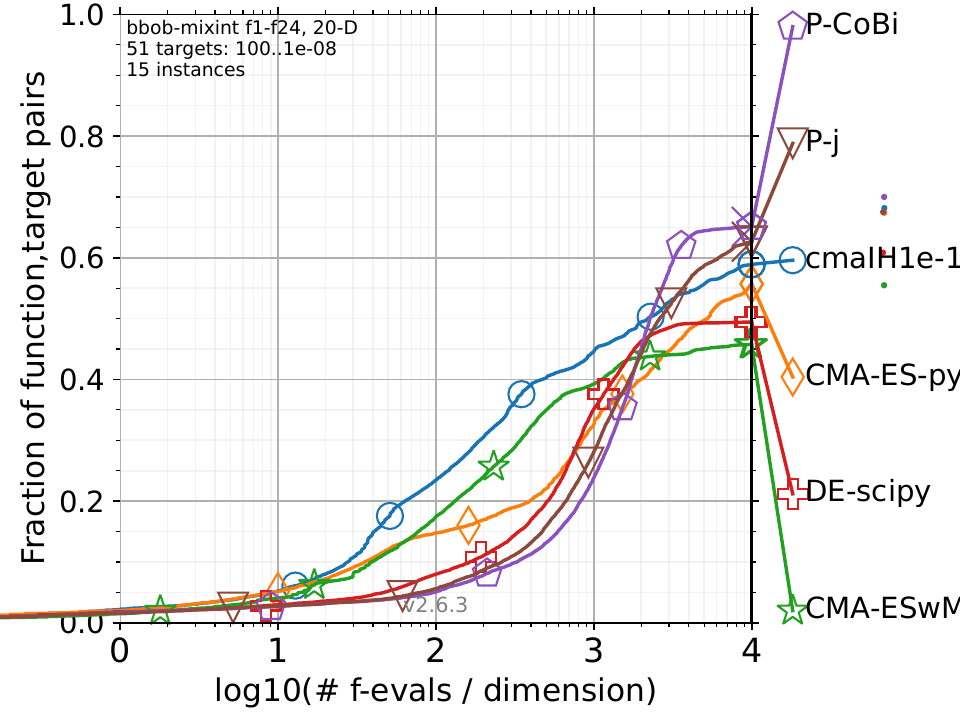}
}
\\
\subfloat[$n=40$]{
\includegraphics[width=\width\textwidth]{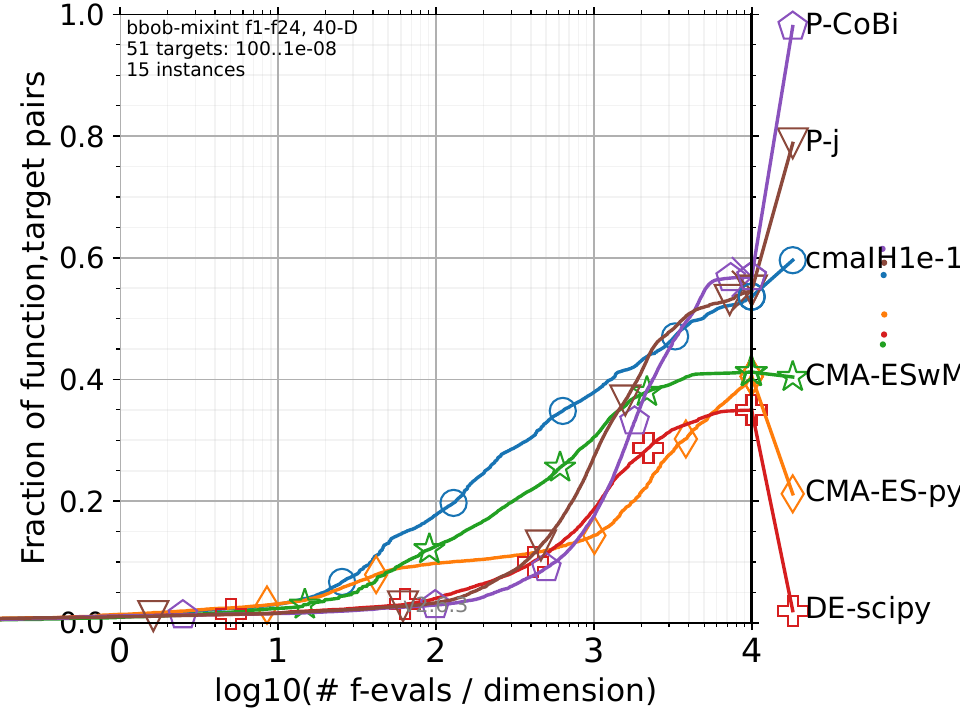}
}
\subfloat[$n=80$]{
\includegraphics[width=\width\textwidth]{./figs/vs_cma/cmaIH_CMA-E_CMA-E_DE-sc_P-CoB_P-j/pprldmany_80D_noiselessall.pdf}
}
\subfloat[$n=160$]{
\includegraphics[width=\width\textwidth]{./figs/vs_cma/cmaIH_CMA-E_CMA-E_DE-sc_P-CoB_P-j/pprldmany_160D_noiselessall.pdf}
}
\caption{Comparison of P-j and P-CoBi with the three CMA-ES variants on the 24 \texttt{bbob-mixint} functions with $n \in \{5, 10, 20, 40, 80, 160\}$.}
\label{supfig:vs_cma}
%
\centering
\subfloat[Separable functions ($f_1, ...,  f_5$)]{
\includegraphics[width=\width\textwidth]{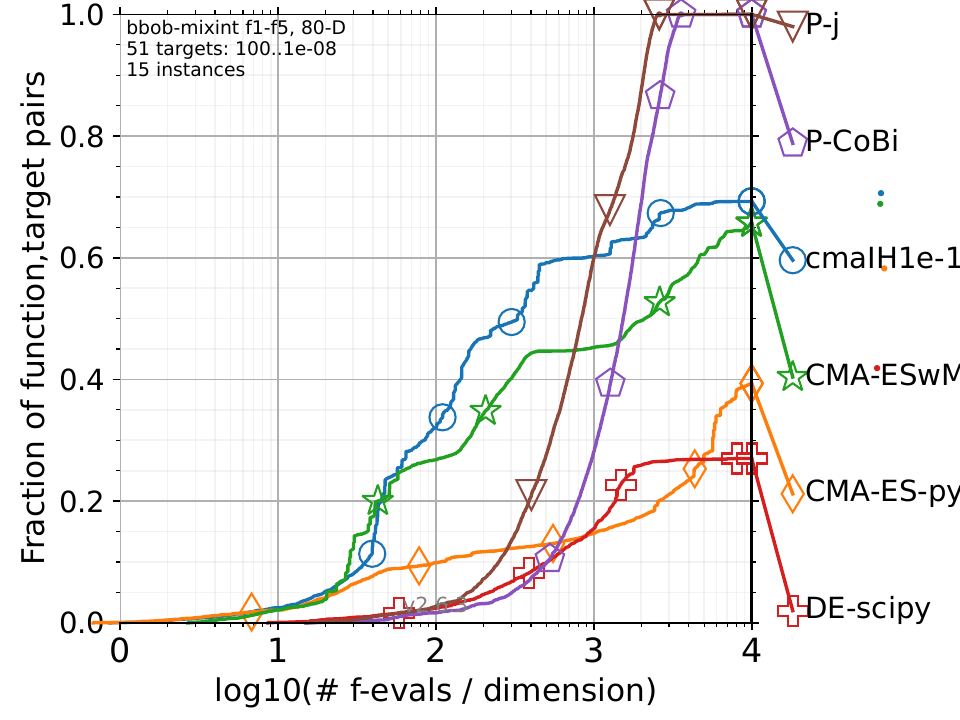}
}
\subfloat[Functions with low conditioning ($f_{6}, ..., f_{9}$)]{
\includegraphics[width=\width\textwidth]{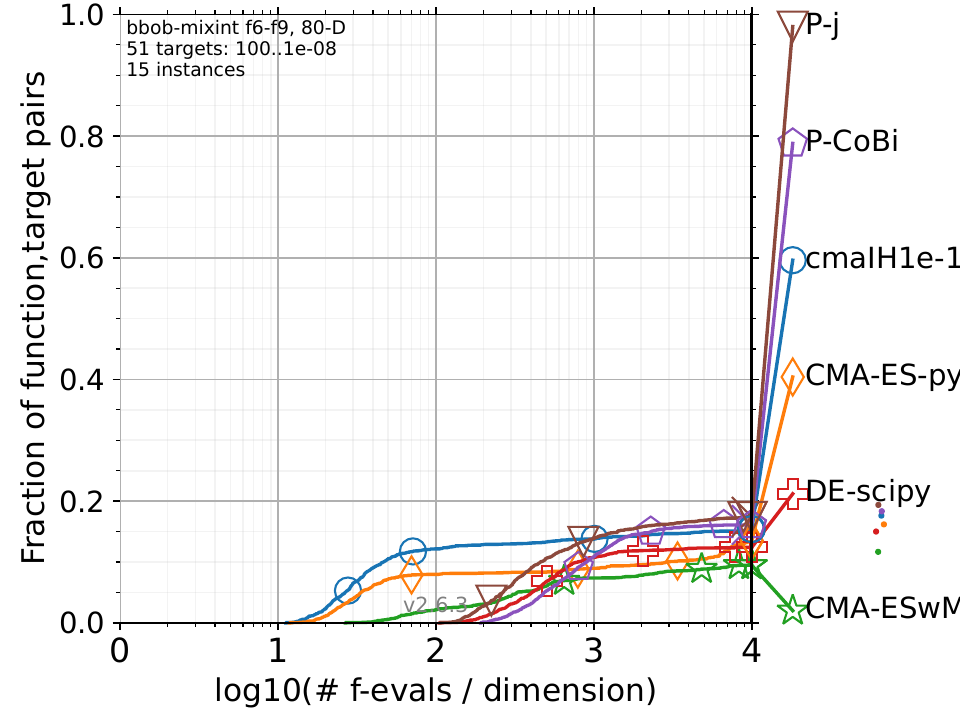}
}
\subfloat[Functions with high conditioning ($f_{10}, ..., f_{14}$)]{
\includegraphics[width=\width\textwidth]{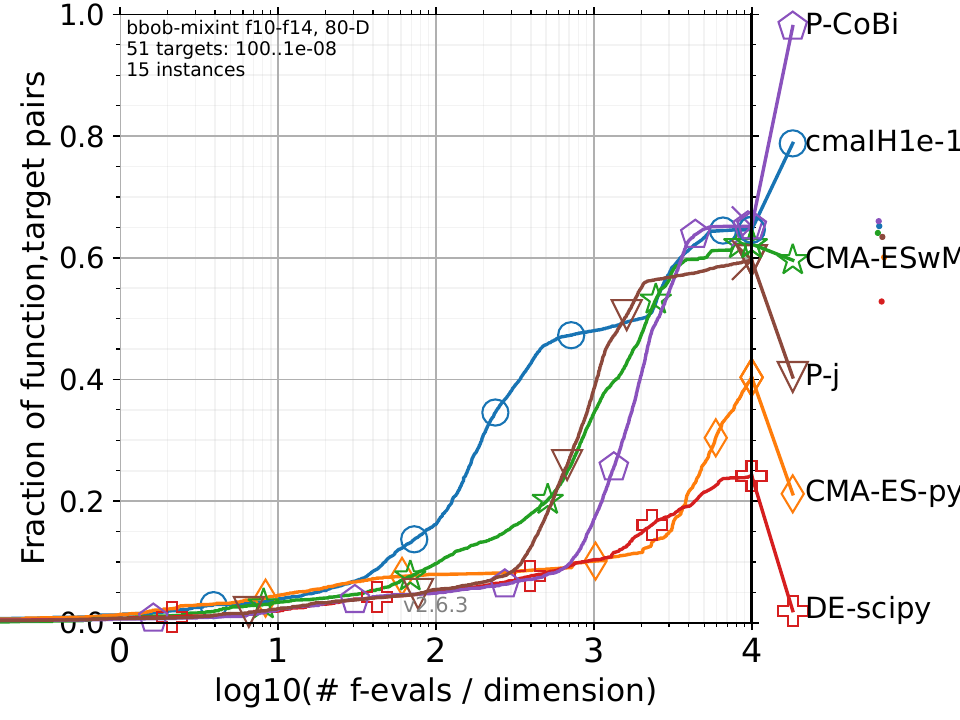}
}
\\
\subfloat[Multimodal functions with adequate global structure ($f_{15}, ..., f_{19}$)]{
  \includegraphics[width=\width\textwidth]{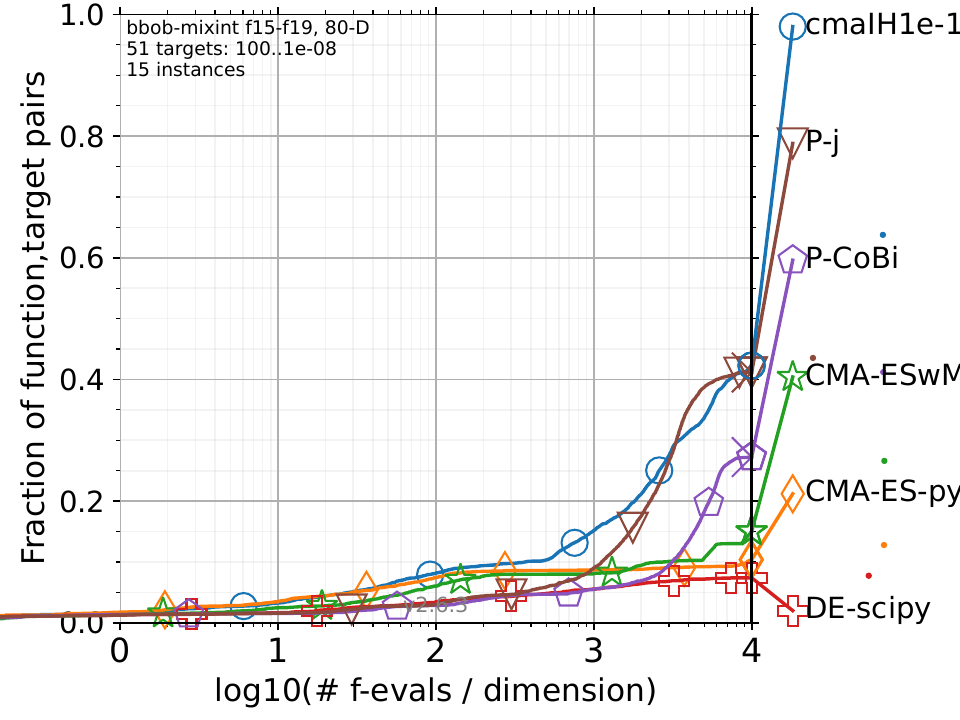}  
}
\subfloat[Multimodal functions with weak global structure ($f_{20}, ...,  f_{24}$)]{
  \includegraphics[width=\width\textwidth]{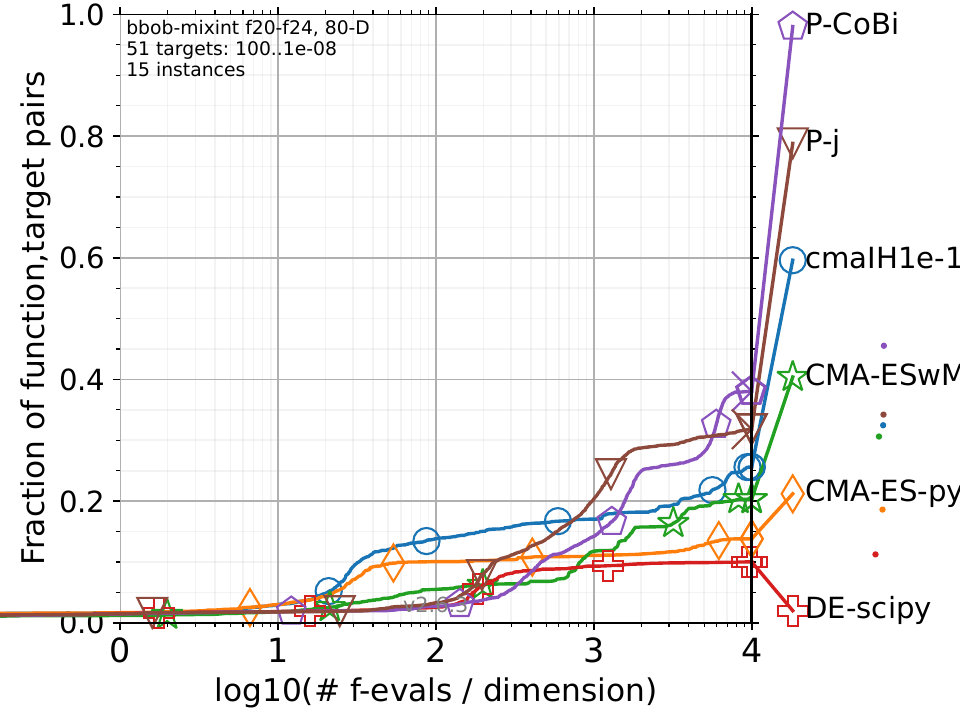}  
}
\caption{Comparison of P-j and P-CoBi with the three CMA-ES variants on each function group with  $n =80$.}
\label{supfig:vs_cma_each_group}
\end{figure*}

\begin{figure*}[htbp]  
\newcommand{\width}{0.235}
\centering
\subfloat[Error value]{
\includegraphics[width=\width\textwidth]{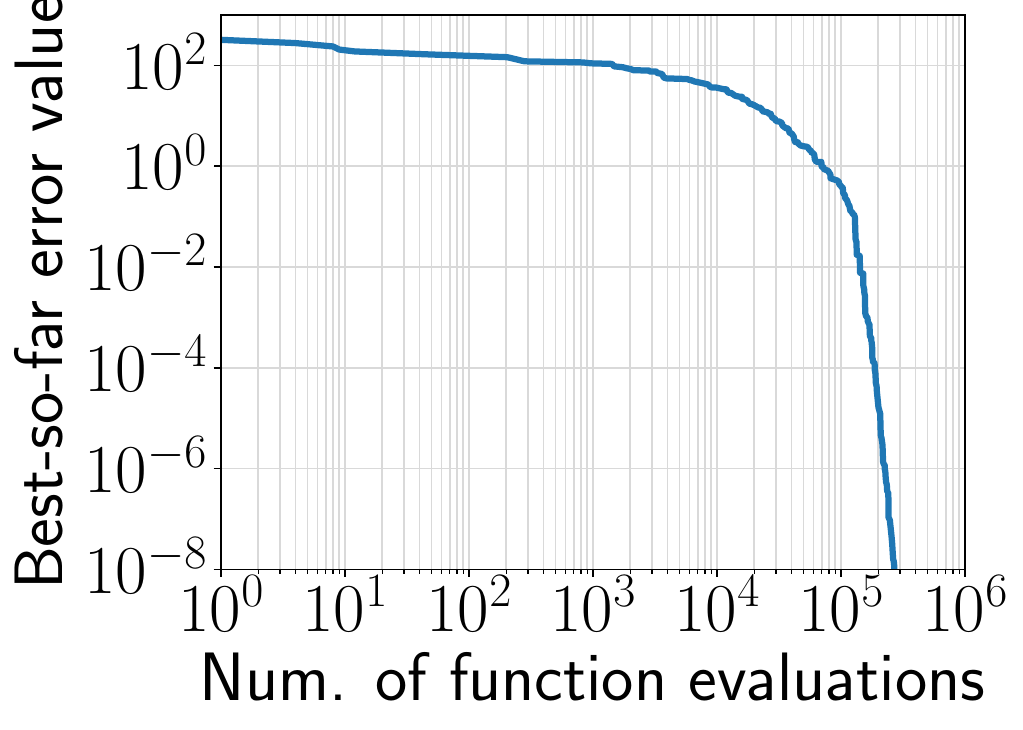}
}
\subfloat[\texttt{div} (left) and \texttt{nsame} (right) values]{
\includegraphics[width=0.26\textwidth]{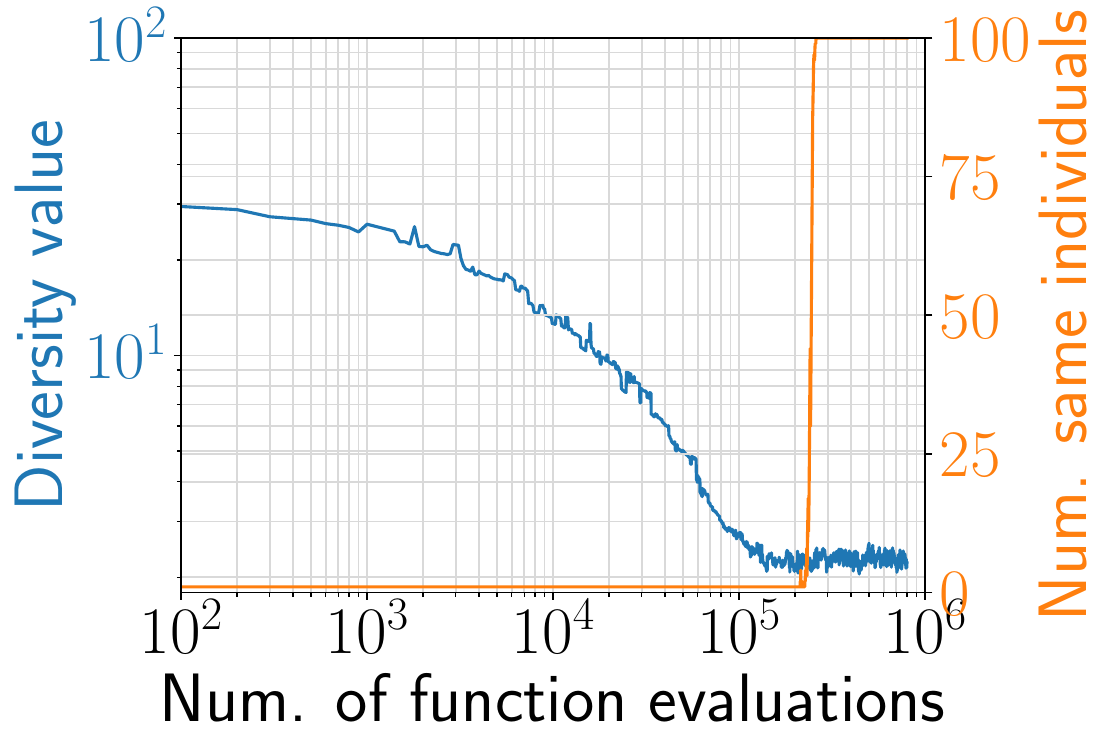}
}
\subfloat[$m_s$ and $m_c$]{
\includegraphics[width=\width\textwidth]{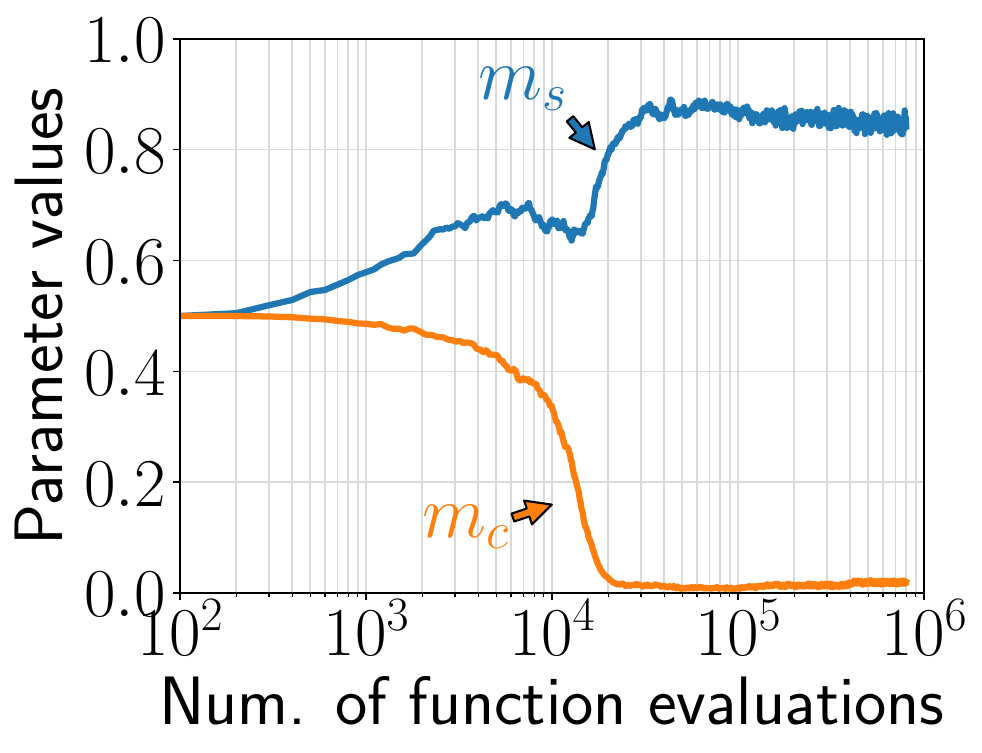}
}
\subfloat[Mean succ. parameters]{
\includegraphics[width=\width\textwidth]{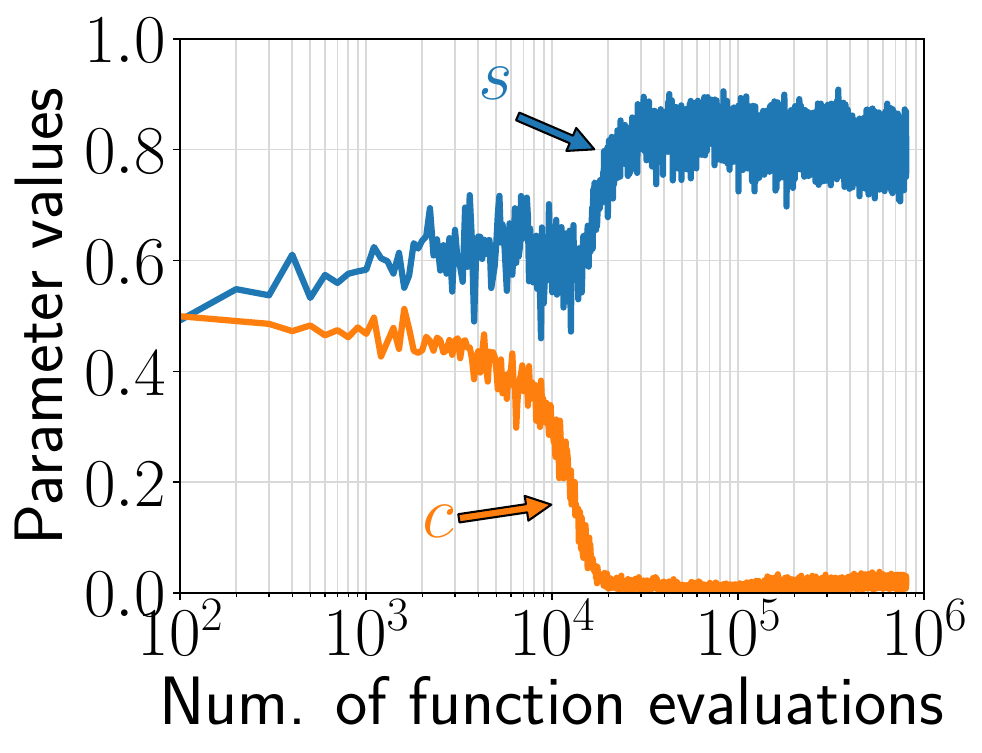}
}
\caption{Analysis results of a typical single run of P-JA on $f_3$ with $n=80$.}
\label{supfig:analysis_pja}
\end{figure*}









\end{document}